\documentclass[10pt, conference, a4paper, compsocconf]{IEEEtran}
\usepackage{cite}

\usepackage[nolist]{acronym}

\usepackage{graphicx}
\usepackage{color}
\usepackage{siunitx}
\DeclareSIUnit\pixel{P}
\usepackage[super]{nth}

\usepackage{comment}

\usepackage{amsmath,amssymb,amsfonts}
\usepackage[switch,columnwise]{lineno}
\usepackage{csquotes}
\usepackage{subcaption}
\usepackage{color,soul}
\usepackage[edges]{forest}
\usepackage{blindtext}
\usepackage{wrapfig}
\usepackage{mwe}
\usepackage{nicefrac}
\usepackage{bm}
\usepackage{mathtools}
\usepackage{latexsym}
\usepackage{tablefootnote}
\usepackage{threeparttable}
\usepackage{adjustbox}
\usepackage{enumitem} 
\usepackage{pifont}
\usepackage{xfrac}
\usepackage{hyperref}
\hypersetup{colorlinks=true,urlcolor=blue}
\usepackage{cleveref}
\Crefname{figure}{Fig.}{Figs.}
\Crefname{table}{Tab.}{Tabs.}
\Crefname{equation}{Eqn.}{Eqns.}
\Crefname{algorithm}{Algo.}{Algos.}
\Crefname{section}{Sec.}{Secs.}
\usepackage{tabularx}
\usepackage{multirow}
\usepackage{array}
\usepackage{booktabs}
\usepackage{import}
\usepackage{placeins}

\usepackage{layouts}
\usetikzlibrary{shadows.blur}
\usetikzlibrary{positioning}
\usetikzlibrary{arrows.meta}
\usetikzlibrary{positioning}
\usetikzlibrary{spy}
\usepackage[skins]{tcolorbox}
\usepackage{pgfplots}
\DeclareUnicodeCharacter{2212}{−}
\usepgfplotslibrary{groupplots,dateplot}
\usetikzlibrary{patterns,shapes.arrows}
\pgfplotsset{compat=newest}

\makeatletter
\DeclareRobustCommand\onedot{\futurelet\@let@token\@onedot}
\def\@onedot{\ifx\@let@token.\else.\null\fi\xspace}
\def\etal{et~al\onedot}
\def\eg{e.g\onedot, } 
\def\ie{i.e\onedot, }

\makeatother

\def\clap#1{\hbox to 0pt{\hss #1\hss}}%
\def\initials#1{\protect\clap{\smash{\raisebox{1.4ex}{\tiny{\textsf{\textit{#1}}}}}}}%
\makeatletter
\newcommand{\EDIT}[4][]{\protect\@ifundefined{hidecomments}{%
  \strut{\color{#3}{\hspace{0pt}\initials{#2}\protect\sout{#1}{~#4}}}%
  }{}}
\newcommand{\NOTEboxed}[3]{\protect\@ifundefined{hidecomments}{%
  {\begin{center}\fbox{\parbox{0.97\linewidth}{\protect\EDIT{#1}{#2}{#3}}}\end{center}}
  }{}}
\makeatletter
\newcommand{\DefAuthor}[2] 
{%
  \expandafter\newcommand\csname #1edit\endcsname[2][]{\protect\EDIT[##1]{#1}{#2}{##2}}
  \expandafter\newcommand\csname #1\endcsname[1]{\protect\csname #1edit\endcsname{[##1]}}
  \expandafter\newcommand\csname #1boxed\endcsname[1]{\NOTEboxed{#1}{#2}{##1}}
}
\makeatother

\def\numx#1e#2{{#1}\mathrm{e}{#2}}

\def\BibTeX{{\rm B\kern-.05em{\sc i\kern-.025em b}\kern-.08em
    T\kern-.1667em\lower.7ex\hbox{E}\kern-.125emX}}

\begin{acronym}
\acro{ANE}{Apple Neural Engine}
\acro{API}{Application Programming Interface}
\acro{CG}{Computer Graphics}
\acro{CNN}{Convolutional Neural Network}
\acro{CPU}{Central Processing Unit}
\acro{CV}{Computer Vision}
\acro{DL}{Deep Learning}
\acro{DNN}{Deep Neural Network}
\acro{EPE}{Endpoint Error}
\acro{ETF}{Edge Tangent Flow}
\acro{FPS}{Frames per second}
\acro{GPU}{Graphics Processing Unit}
\acro{GAN}{Generative Adversarial Network}
\acro{IB-AR}{Image-based Artistic Rendering}
\acro{K}{Kilo, thousand}
\acro{M}{Million}
\acro{MB}{Megabyte}
\acro{NST}{Neural Style Transfer}
\acro{NPR}{Non-photorealistic Rendering}
\acro{NPU}{Neural Processing Unit}
\acro{SNP}{Stylized Neural Painting}
\acro{ReLU}{Rectified Linear Unit}
\acro{TFLOPS}{$10^{12}$ Floating Point Operations Per Second}
\acro{PPN}{Parameter Prediction Network}
\acro{DoG}{Difference-of-Gaussians}
\acro{XDoG}{eXtended difference-of-Gaussians}
\acro{SSIM}{Structural Similarity Index Measure}
\acro{PSNR}{Peak Signal-to-Noise Ratio}
\acro{FID}{Fréchet Inception Distance}
\acro{LPIPS}{Learned Perceptual Image Patch Similarity}
\acro{BLADE}{Best Linear Adaptive Enhancement}
\acro{STROTSS}{Style Transfer by Relaxed Optimal Transport and Self-Similarity}
\end{acronym}

\def\isArxiv{}

\begin{document}

\title{Controlling Geometric Abstraction \\and Texture for Artistic Images}


\author{
\IEEEauthorblockN{Martin Büßemeyer*, Max Reimann*, Benito Buchheim}
\IEEEauthorblockA{Hasso Plattner Institute for Digital Engineering\\ 
University of Potsdam, Germany\\
* authors contributed equally
}
\and
\IEEEauthorblockN{Amir Semmo}
\IEEEauthorblockA{Digital Masterpieces GmbH\\
Potsdam, Germany
}
\and
\IEEEauthorblockN{J\"urgen D\"ollner, Matthias Trapp}
\IEEEauthorblockA{Hasso Plattner Institute for \\
Digital Engineering \\
University of Potsdam, Germany
}
}

\maketitle

\begin{figure*}
\centering
\hfill
\begin{subfigure}{0.175\textwidth}
    \includegraphics[trim={0cm 0.2cm 0cm 0cm},clip, height=7.1cm]{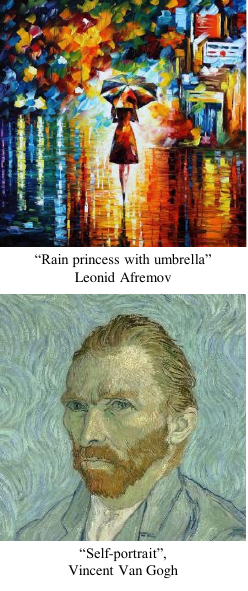}
    \caption*{Input}
\end{subfigure}\hfill
\begin{subfigure}{0.4\textwidth}
    \includegraphics[height=7.1cm]{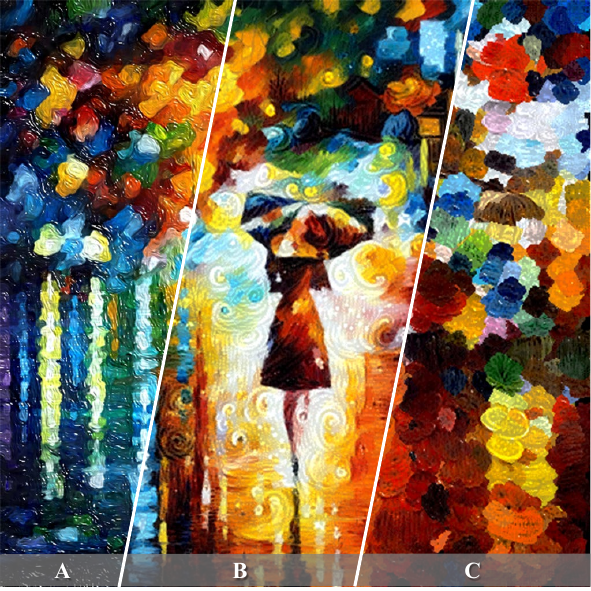}
    \caption*{Exemplary Texture Editing}
\end{subfigure}\hfill%
\begin{subfigure}{0.4\textwidth}
    \includegraphics[height=7.1cm]{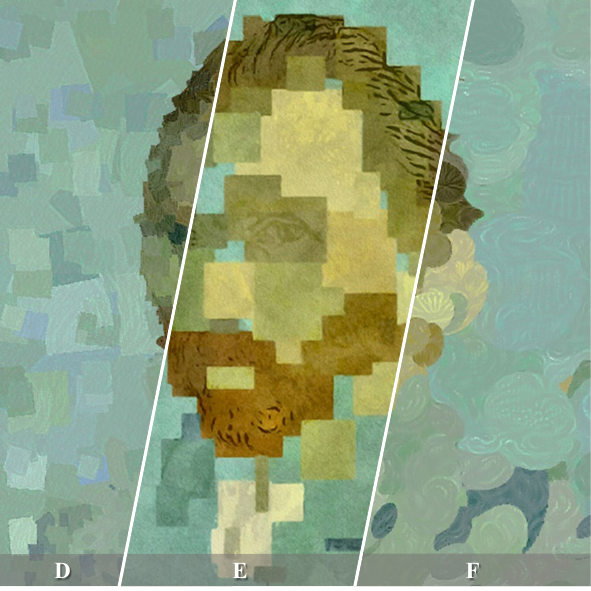}
    \caption*{Exemplary Geometric Abstraction Editing}
\end{subfigure}%
\hfill
  \caption{Our method enables texture and geometric abstraction editing while keeping the overall color and structure intact. In A-C, textures are adjusted by increasing the oiliness and stroke thickness (A) and using the text prompts \enquote{starry night} (B) and \enquote{impressionistic painting} (C). In D - F the geometric abstraction primitives are varied between rectangular strokes (D), blocks (E), and ellipsoids (F). These edits can be combined and interactively adjusted.}
  \label{fig:teaser}
  \vspace{-0.2cm}
\end{figure*}

\begin{abstract}
We present a novel method for the interactive control of geometric abstraction and texture in artistic images. Previous example-based stylization methods often entangle shape, texture, and color, while generative methods for image synthesis generally either make assumptions about the input image, such as only allowing faces or do not offer precise editing controls. By contrast, our holistic approach spatially decomposes the input into shapes and a parametric representation of high-frequency details comprising the image's texture, thus enabling independent control of color and texture. Each parameter in this representation controls painterly attributes of a pipeline of differentiable stylization filters. The proposed decoupling of shape and texture enables various options for stylistic editing, including interactive global and local adjustments of shape, stroke, and painterly attributes such as surface relief and contours. Additionally, we demonstrate optimization-based texture style-transfer in the parametric space using reference images and text prompts, as well as the training of single- and arbitrary style parameter prediction networks for real-time texture decomposition.
\end{abstract}


\section{Introduction}

%
\noindent While recent methods for image synthesis, such as diffusion models~\cite{rombach2022high}, and example-based stylization methods, such as \ac{NST}~\cite{gatys2016image,huang2017arbitrary}, achieve impressive results, their black-box character and lack of disentangled artistic control variables \cite{hertzmann2022toward} makes precise adjustment of geometric elements and texture challenging. Thus, given an artistic image obtained from such methods or given an artistic image without having control over its formation process, achieving a desired artistic look may necessitate further adjustments to geometric and textural artistic control variables. These variables include geometric elements such as brush shapes and sizes, and textural elements such as stroke patterns, tonal variations, and other intricate features in the image.
In this paper, we propose a novel method for geometric abstraction and texture control that allows example-based as well as parametric control over such artistic variables. Our method\footnote{Project and Code: \url{https://maxreimann.github.io/artistic-texture-editing/}} accepts an artistic image as input, without any restrictions on the content domain, and outputs an image with applied adjustments in the geometric and textural space which preserve the input color distribution.

To allow for disentangled editing of artistic control variables, we first decompose the input image into primitive shapes representing coarse structure and a parametric representation of high-frequency details that form the image's texture. The coarse structure decomposition is achieved using segmentation techniques, \eg superpixel segmentation \cite{achanta2012slic}, or layered approaches such as neural stroke-based rendering \cite{liu2021paint,zou2021stylized}, depending on the desired shape primitives. 
To decompose the high-frequency details into meaningful artistic control variables, a novel lightweight pipeline of  differentiable stylization filters is introduced. Filters in this pipeline are based on traditional \ac{IB-AR} filters \cite{kyprianidis2012state} implemented in an auto-grad enabled framework~\cite{lotzsch2022wise}, and are parameterized by explicitly defined stylistic or painterly attributes, such as contours, surface relief (\eg for oil-paint texture), and local contrast, among others. Detail decomposition is then achieved by optimizing filter parameters such that the output of the filter pipeline, conditioned on the coarse structure image, matches the input image.

The resulting spatial and value decomposition provides a wide range of options for editing textures and geometric abstraction (\Cref{fig:teaser}). 
The method of geometric abstraction is interchangeable and, depending on the method of choice, provides interactive control over the stroke shapes and the level-of-abstraction.
Furthermore, filter parameters can be adjusted interactively on both global and local levels using per-pixel parameter masks, either by manual editing or interpolating values based on extracted image attributes such as depth or saliency.
Overall, the decomposed representation allows for independent control of color and texture, enabling color adjustments without affecting the texture and vice versa, which is particularly useful for editing tasks such as correcting artifacts in images created by \ac{NST}.

 Our approach performs particularly well for example-based edits of fine-granular texture patterns by adapting to the underlying coarse shape primitives and achieving a spatially homogeneous appearance. At this, example-based losses can be used to optimize the texture in the parametric space to conform to a desired style image using \ac{NST} style-losses  \cite{gatys2016image, kolkin2019style} or text-based losses \cite{kwon2022clipstyler}. 
 For particular stylization tasks such as style transfer, the texture decomposition optimization can be further accelerated by training \acp{PPN} for real-time decomposition.

To summarize, we make the following contributions:
\begin{enumerate}
\item We present a holistic approach for geometric abstraction and texture editing that decomposes the image into coarse shapes and high-frequency detail texture.
\item We present a novel differentiable filter pipeline for texture editing. Compared to prior work it is lightweight and improves parameter editability.
\item We introduce \acp{PPN} for real-time  single and arbitrary style texture decomposition. 
\item We demonstrate that a texture can be adapted to new styles using example images or text prompts, and can be interactively adjusted on a global or local level.
\end{enumerate}

\section{Related Work}

\noindent Deep network-based methods, particularly \ac{NST} \cite{gatys2016image}, learn stylistic representations  in a black-box fashion, transferring texture style from a reference style image to a content image. 
In addition to the style-content trade-off control inherent to \ac{NST} \cite{gatys2016image}, several methods have been developed to enable control over aspects such as color \cite{GatysEBHS17} or strokes \cite{reimann2022controlling, Jing_2018_ECCV}.  
CLIPstyler \cite{kwon2022clipstyler} can perform style transfer from text prompts using CLIP-based losses \cite{radford2021learningCLIP}. 
However, these methods either cannot be interactively controlled (\ie optimization-based approaches) 
or are not easily composable with each other (\ie network-based approaches).

Traditional \ac{IB-AR} \cite{kyprianidis2012state}, on the other hand, represent styles as a chain of image filters that allow fine-granular control over multiple style aspects, but have to be specifically designed for a particular style, such as for watercolor \cite{bousseau2006interactive}, oil paint \cite{semmo2016image}, or cartoon \cite{winnemoller2006real}. Recently, Lötzsch \etal \cite{lotzsch2022wise} proposed an interactively controllable "whitebox" style representation by optimizing the parameters of such IB-AR filter pipelines to match a stylized reference image. 
Similar to this work, we also use a filter-based interpretable style representation.
However, Lötzsch \etal \cite{lotzsch2022wise} represent the entire input image in such parameters, \ie shape, color, and details (\eg local textures) are entangled in a large set of often redundant parameters and filters. This makes editing tedious and unintuitive, as adjusting parameters controlling painterly attributes may have unwanted side-effects on colors or shapes. 
Instead of matching the entire input image, we represent color and shape distribution in a geometric abstraction stage and use our subsequent filter pipeline to represent the high frequency components, \ie the local textures. In contrast to the intricate, style-specific filter pipelines \cite{bousseau2006interactive, semmo2016image} employed by  \cite{lotzsch2022wise}, we remove redundancy and improve parameter editability by proposing a light-weight filter pipeline capable of matching arbitrary styles while consisting of only four filters. 

While Lötzsch \etal \cite{lotzsch2022wise} demonstrate the viability of \acp{PPN} for a single img2img translation task in combination with an additional postprocessing CNN, we show that \acp{PPN} can be adapted for general single- and arbitrary style transfer decomposition tasks, and do not require postprocessing networks.

 Our geometric abstraction stage is inspired by approaches that use either segmentation-based representation of shapes \cite{song2008arty,ihde2022design} or stroke-based rendering \cite{hertzmann1998painterly, huang2019learning} to abstract the image into primitive shapes. Specifically, we make use of recent neural painting approaches using optimization-based \cite{zou2021stylized} or prediction-based \cite{liu2021paint} methods to represent the image as a set of layered primitives, or alternatively use SLIC\cite{achanta2012slic}-based segmentation for non-layered representation. 
 
\begin{figure}[t]
    \centering
    \includegraphics[width=\linewidth]{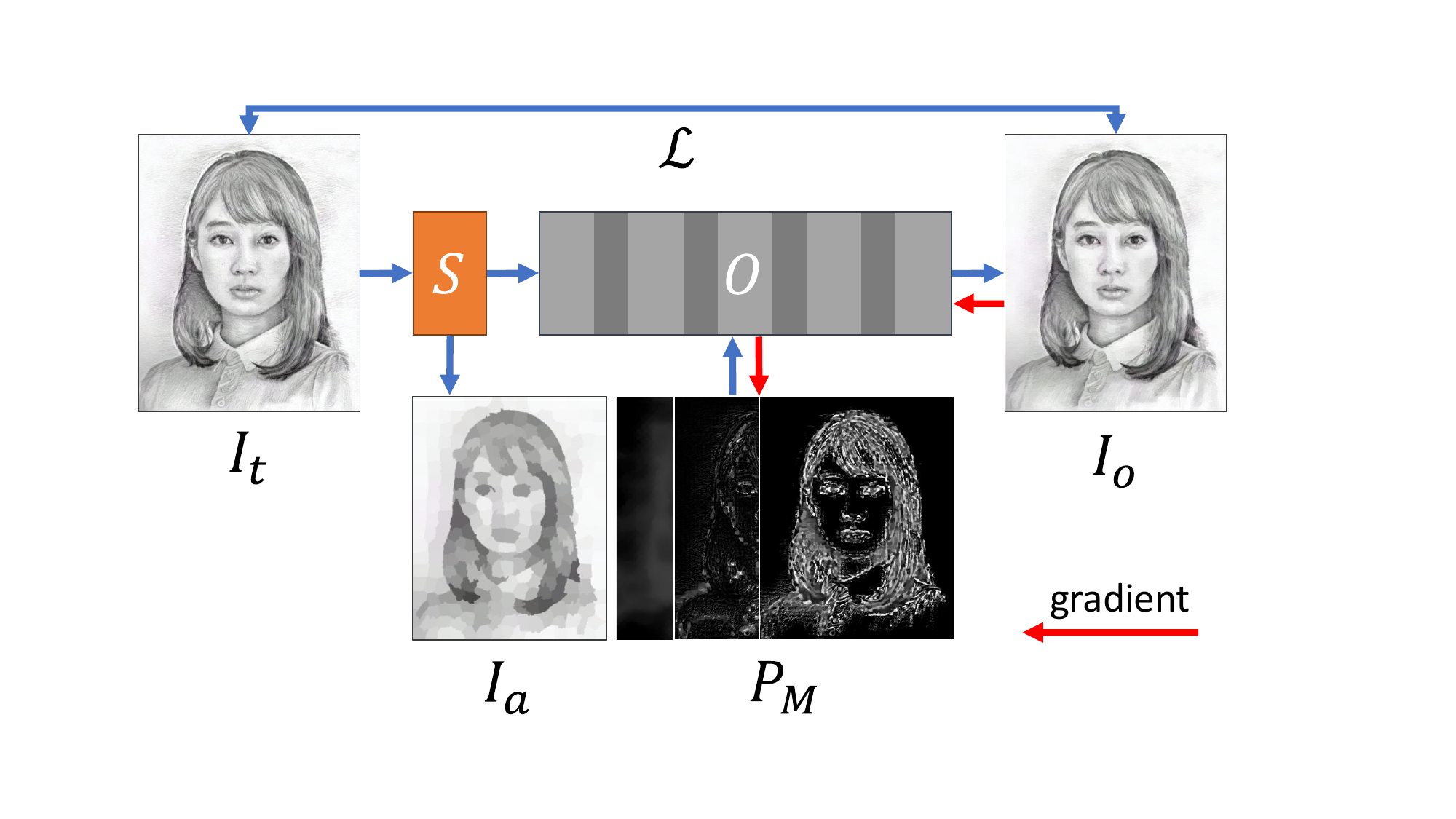}
    \vspace{-0.5cm}\caption{Texture decomposition. The parameter masks $P_M$ controlling individual painterly attributes in a filter pipeline $O$ are optimized through loss $\mathcal{L}$ to match output image $I_o$ and input image $I_t$. The pipeline expects an abstracted image $I_a$ as input, which can be obtained from different methods for the geometric abstraction of $I_t$ in the segmentation stage $S$.}
    \label{fig:pipeline}
\end{figure}

\section{Method}

\subsection{Framework Overview}
\noindent To obtain a decomposed image representation for downstream editing tasks, texture decomposition is initially executed. \Cref{fig:pipeline} shows the two involved stages: (1) a segmentation stage $S(\cdot)$ to control texture granularity and geometric abstraction, and (2) a pipeline of differentiable image filters $O(\cdot)$ to represent image details. The first stage renders an abstracted variant $I_a$ of the input image $I_t$ using shape primitives. The second stage represents the details, \ie the difference between $I_a$ and $I_t$, in the parameters of a filter pipeline $O$. Each of these filter parameters represents a specific artistic control variable, \eg contours, and local contrast, among others (\Cref{subsec:diffilterstage}), and is controllable on a per-pixel level using parameter masks $P_M$. Decomposition is achieved by first executing stage $S$ and then optimizing the decomposition loss over pipeline output $I_o$ to adapt $P_M$, which we detail in the following.

\subsection{Decomposition Loss}
\noindent Formally, $O$ is parametrized by a set of $M$ parameter masks, \ie $P_M = \{ P_i \in \mathbb{R}^{h \times w} | i \leq M \}$ and expects the output image of the segmentation stage $I_a=S(I_t)$ as input.
A stylized output image $I_o$ is thus obtained as:
\begin{equation}
    I_o = O\left(P_M, S(I_t)\right)
\end{equation}
To obtain the decomposed texture representation, the parameters $P_M$ are optimized using a loss function $\mathcal{L}$ that combines a target loss, denoted by $\mathcal{L}_{\mathit{target}}$, and the total variation loss, denoted by $\mathcal{L}_{\mathit{TV}}$ weighted by $\lambda_{\mathit{TV}}$:
\begin{equation}
\label{eq:overall}
    \mathcal{L} = \mathcal{L}_{\mathit{target}} + \lambda_{\mathit{TV}}\mathcal{L}_{\mathit{TV}}
\end{equation}
The target loss ensures closeness to a desired target style, while $\mathcal{L}_{\mathit{TV}}$ reduces noise in masks and enforces local consistency. 

Different types of loss functions can be used for the target loss as follows. When the objective is to allow for subsequent interactive editing, the $\ell_1$ loss is a suitable choice as it optimizes for reconstructing the details of the input image:
\begin{equation}
    \mathcal{L}_{\mathit{target}} = \left\| O(P_M, S(I_t)) - I_t \right\|_1
\end{equation}
When the objective is to control and change the details of a style, text or image-based style transfer losses for $\mathcal{L}_{\mathit{target}}$ can be used, as shown in \Cref{subsec:styledetailvariants}. Similarly, such losses can also be used to train a \ac{PPN} to reconstruct the details of a style-transferred input image in a single inference step. To demonstrate this, we introduce single-style and arbitrary-style decomposition networks in \Cref{subsec:styletransferprediction}.

\subsection{Segmentation Stage}
\noindent The segmentation stage $S$ divides the input image $I_t$ into distinct shape primitives, \eg brushstrokes or segments, of uniform color. As $S$ operates on pixels in both the input and output domains, a wide range of methods such as superpixel segmentation or stroke-based rendering can be utilized in this stage. We discuss several choices for the segmentation stage in \Cref{subsec:geomabstractioncontrol}. While changing abstraction settings in the segmentation stage is typically interactive and enables effect previews, the subsequent filter stage must be subsequently re-optimized to adapt to the newly introduced geometric structure.

\subsection{Differentiable Filter Stage}
\label{subsec:diffilterstage}

\noindent Existing heuristics-based stylization pipelines (\eg \cite{winnemoller2006real,bousseau2006interactive,semmo2016image}) are not well-suited for example-based stylization editing tasks, as they were not designed with these specific use cases in mind. They often have non-trainable parameters, such as color quantization \cite{winnemoller2006real} which require differentiable proxies \cite{lotzsch2022wise}, and furthermore have repetitive components in the pipeline, such as repeated smoothing, which can hinder the optimization and ease of editing local parameter masks. To this end, we ensure that all parameters in our proposed pipeline are differentiable and that the pipeline is kept simple with intuitive parameters that produce the expected effects, while at the same time having the capacity to match arbitrary textures. 
To preserve the input color distribution (\ie the segmentation stage output), filters should not be able to freely alter pixels on the color spectrum during optimization. Consequently, our pipeline's learnable parameters cannot modify color hue.

We propose a lightweight filter pipeline $O(\cdot)$ consisting of following differentiable filters: 

\begin{description}
\item[(1) Smoothing:] We use a Gaussian smoothing ($\sigma=1$) followed by a bilateral filter \cite{tomasi1998bilateral}, with learnable parameters $\sigma_d$ (distance kernel size) and $\sigma_r$ (range kernel size).  
 \item[(2) Edge enhancement:] We implement \ac{XDoG} \cite{winnemoller2012xdog} with learnable parameters contour amount and contour opacity. 
\item[(3) Painterly attributes:] We use bump mapping for surface relief control in our evaluation and implement Phong shading \cite{phong1975illumination} with learnable parameters bump-scale, Phong-specularity, and bump-opacity.
However, any differentiable painterly filter can be used, \eg  we also implement  wet-in-wet filters  \cite{wang2014towards} and wobbling \cite{bousseau2006interactive} for watercolorization control.
\item[(4) Contrast:] This learnable parameter controls the amount of local contrast enhancement applied.
\end{description}

\noindent Gradients are computed with respect to both, the parameter masks $P_M$ and the image input. To this end, such image filters are implemented in an auto-grad-enabled framework following \cite{lotzsch2022wise}.  
The ablation study in \Cref{subsec:quantativecomparisons} shows that all stages are necessary for arbitrary style representation. The proposed pipeline can be easily configured and optimized with further \ac{IB-AR} techniques \cite{kyprianidis2012state} for extended control of painterly attributes.

\subsection{Optimization of Parameter Masks}
\noindent The parameter masks $P_M$ are optimized using gradient descent minimization of $\mathcal{L}$. 
Optimizing only $\mathcal{L}_{\mathit{target}}$ as in \cite{lotzsch2022wise} results in strongly fragmented masks exhibiting large local variations of values (\Cref{fig:masks}, first row), which makes them hard to edit.
When optimizing with $\mathcal{L}_\mathit{TV}$, masks have reduced noise, increased sparsity, and greater smoothness compared to those optimized without constraints. 

Detail optimization broadly follows \cite{lotzsch2022wise}; local parameter masks are optimized with 100 iterations of Adam~\cite{kingma2014adam} and a learning rate of 0.01. We decrease the learning rate by a factor of 0.98 every 5 iterations starting from iteration 50. In contrast to~\cite{lotzsch2022wise}, smoothing of parameter masks is not required, as the employed filters do not exhibit artifacts. Throughout this paper, we use $\lambda_\mathit{TV}=0.2$ during optimization.

\newcommand{\imagewithspymaskfig}[3]{
    \begin{tikzpicture}[every node/.style={inner sep=0,outer sep=0}]
        \begin{scope}[ 
			inner sep = 0pt,outer sep=0,spy using outlines={rectangle, red, magnification=3}
                ]
        \node (n0)  { \includegraphics[width=\textwidth]{#1}};
        \spy [red,size=0.8cm] on (#2,#3) in node[anchor=south east,inner sep=3pt] at (n0.south east);
        \end{scope}
    \end{tikzpicture}
}
\begin{figure}[t]
    \centering
    \begin{subfigure}[b]{0.245\linewidth}%
        \imagewithspymaskfig{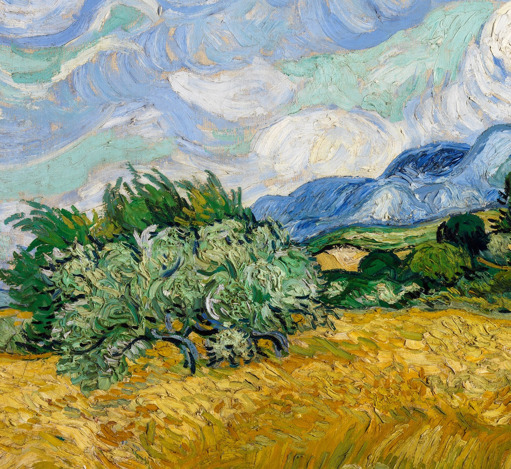}{-0.4cm}{-0.2cm}
        \label{fig:masks:input}%
    \end{subfigure}\hfill
    \begin{subfigure}[b]{0.245\linewidth}
        \includegraphics[width=\textwidth]{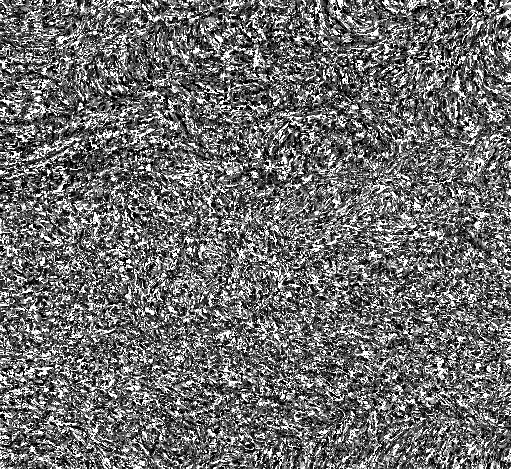}
        \label{fig:masks:bilateral_sigma_d_1:a}%
    \end{subfigure}\hfill
    \begin{subfigure}[b]{0.245\linewidth}
        \includegraphics[width=\textwidth]{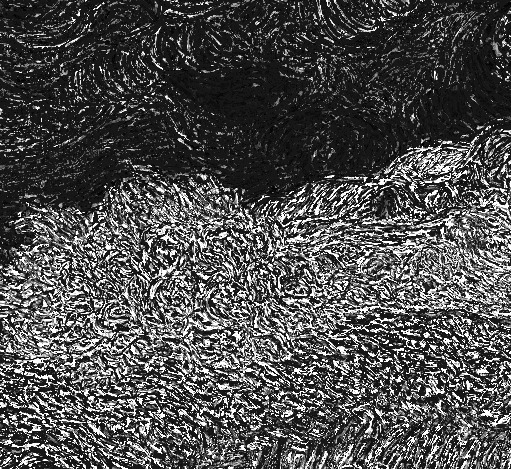}
        \label{fig:masks:bump_phong_specular:a}%
    \end{subfigure}\hfill
    \begin{subfigure}[b]{0.245\linewidth}%
        \includegraphics[width=\textwidth]{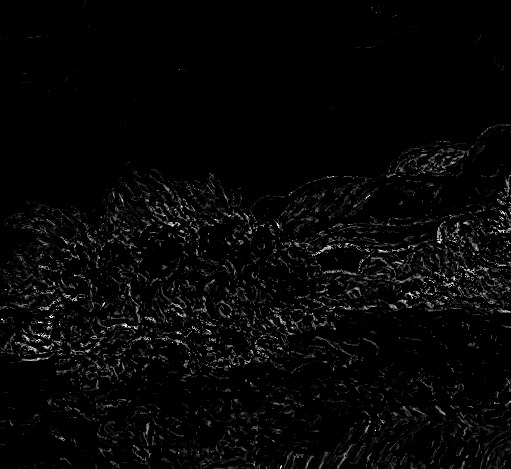}
        \label{fig:masks:contour_opacity:a}%
    \end{subfigure}\\[-2.6ex]
    \begin{subfigure}[t]{0.245\linewidth}%
        \imagewithspymaskfig{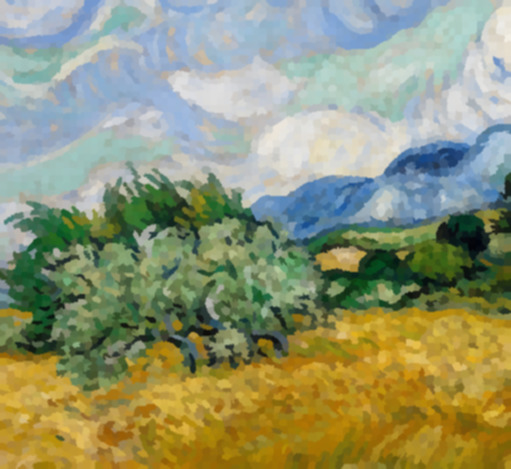}{-0.4cm}{-0.2cm}
        \vspace{-11pt}\caption{Inputs $I_t,I_a$}%
        \label{fig:masks:segmented}%
    \end{subfigure}\hfill
    \begin{subfigure}[t]{0.245\linewidth}%
        \includegraphics[width=\textwidth]{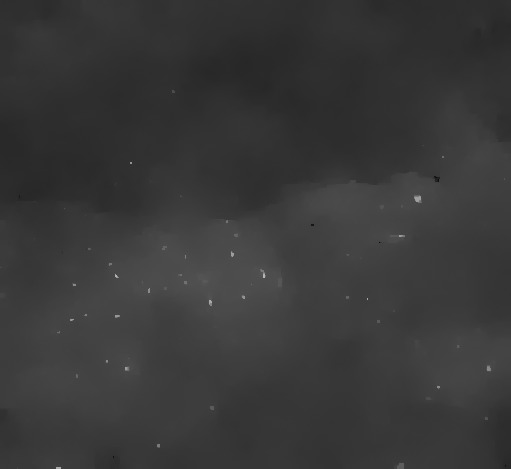}
        \caption{\footnotesize Bilateral $\sigma_d$}%
        \label{fig:masks:bilateral_sigma_d_1:b}%
    \end{subfigure}\hfill
    \begin{subfigure}[t]{0.245\linewidth}%
        \includegraphics[width=\textwidth]{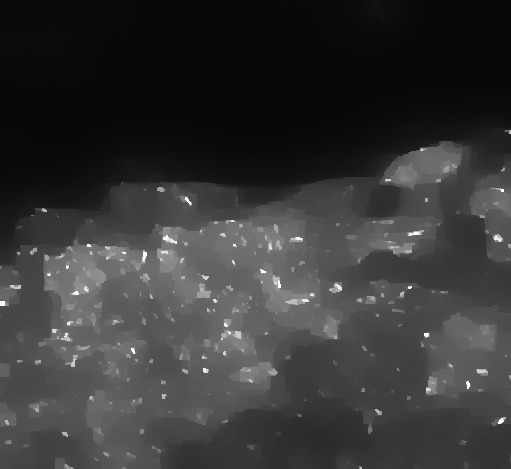}
        \caption{\footnotesize Phong specular}%
        \label{fig:masks:bump_phong_specular:b}%
    \end{subfigure}\hfill
    \begin{subfigure}[t]{0.245\linewidth}%
        \includegraphics[width=\textwidth]{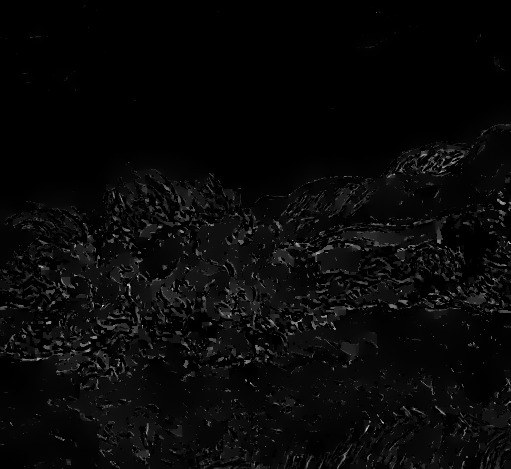}
        \caption{\footnotesize Contour $\alpha$}%
        \label{fig:masks:contour_opacity:b}%
    \end{subfigure}
    \caption{Parameter masks $P_M$ optimized without (top row) and with (bottom row) $\mathcal{L}_\mathit{TV}$ to fit the input $I_t$ (top row) to the given segmented image $I_a$.}
    \label{fig:masks}
\end{figure}

\subsection{Style Transfer Parameter Prediction}
\label{subsec:styletransferprediction}
\noindent While general-purpose, the optimization-based decomposition is compute-intensive, taking approx. \SI{3}{\minute} to optimize a 1MPix image on a Nvidia GTX 3090. For known tasks, such as \ac{NST}, the decomposition step can be sped-up to real-time inference by training a \ac{PPN} \cite{lotzsch2022wise} to directly predict $P_M$, similar to (pixel predicting) \ac{NST} networks~\cite{johnson2016perceptual,park2019arbitrary}. 
We propose single and arbitrary style transfer texture decomposition \acp{PPN} and train them as follows.

 \paragraph{Single Style Decomposition PPN}
We train a \ac{PPN}---$\text{PPN}_\mathit{sst}$---to decompose the texture of a single style image $I_s$. 
We use the \ac{NST} network of Johnson \etal\cite{johnson2016perceptual}, trained on $I_s$, to generate a stylized ground-truth image $I_t$ and segment it using $S$ as a preprocessing step to generate training inputs $I_a$ for our filter pipeline $O$.
The training loss for $\text{PPN}_\mathit{sst}$ is then:
\vspace{-0.2cm}\begin{align}
 \label{eq:nst_pred_io}
        I_o &= O\big(\text{PPN}_{sst}(I_a), I_a\big) \\
    \mathcal{L}_{\text{PPN}_{sst}} &= \mathcal{L}_{gram}(I_o, I_s) + \lambda\mathcal{L}_c(I_o, I_t)
    \label{eq:nst_pred_loss}
\end{align}
where $\mathcal{L}_{gram}$ is the Gram matrix style loss, and $\mathcal{L}_c$ the content loss over VGG~ \cite{simonyan2014very} features following Gatys~\etal\cite{gatys2016image}. The architecture for $\text{PPN}_\mathit{sst}$ is adapted from \cite{johnson2016perceptual} by setting the number of output channels in the last layer to the number of parameter masks ($\#P_M$).

\paragraph{Arbitrary Style Decomposition PPN} 
We train a \ac{PPN}---$\text{PPN}_\mathit{arb}$---on a large set of style images $I_s$ to decompose the textures of arbitrary styles, following the training methodology of previous work on arbitrary style transfer \cite{huang2017arbitrary,park2019arbitrary}.
The architecture for $\text{PPN}_\mathit{arb}$ is adapted from SANet~\cite{park2019arbitrary} by setting the number of output channels in the last layer of the decoder to $\#P_M$.
For training, we adapt the preprocessing steps from before by using SANet to generate $I_t$. The training loss for $\text{PPN}_\mathit{arb}$ is then:
\begin{align}
    I_o &= O\big(\text{PPN}_\mathit{arb}(I_a, I_s), I_a\big) \\
    \mathcal{L}_{\text{PPN}_\mathit{arb}} &= \lambda_s\mathcal{L}_{\text{AdaIN}}(I_o, I_s) + \lambda_c\mathcal{L}_c(I_o, I_t) 
\end{align}
Following SANet~\cite{park2019arbitrary}, we use the AdaIN~\cite{huang2017arbitrary} style loss. In contrast to SANet, we do not use a structure-retaining identity loss, as our filter pipeline $O$ is itself much more constrained in its ability to change content structures than SANet.

\paragraph{Training Details}
We trained both $\text{PPN}_\mathit{sst}$ and $\text{PPN}_{\mathit{arb}}$ using the MS-COCO dataset \cite{cocodataset} for content images and used the WikiArt dataset as style images for training $\text{PPN}_{\mathit{arb}}$. 
As \acp{PPN} learn to predict suitable parameters over a large set of examples, the predicted masks contain little noise compared to optimized masks, thus $\mathcal{L}_{\mathit{TV}}$ is not required for \ac{PPN} training.
Please see the supplementary for details on training hyperparameters.
\section{Controlling Aspects of Style} 
\noindent There are various methods for controlling the decomposed style representation after optimizing or predicting style parameters. 
We discuss methods for geometric abstraction control, and present techniques such as manual editing, reoptimizing using different style transfer losses, and interpolating predicted parameters. 

\begin{table}
    \centering
    \caption{Properties of the superpixel and neural painting methods such as PaintTransformer (PTf) and neural painter (NPtr), for application in the segmentation stage. PTf uses a fixed shape primitive during training, whereas NPtr uses a trainable generator network to synthesize brush shapes.  
    }    
    \label{tab:segmentationstageproperties}    
    \begin{tabular}{|l|l|l|l|l|}
    \hline    
        Method & Type & Primitive & Control & Runtime \\
    \hline
        SLIC \cite{achanta2012slic} & Superpixel & segment & \#segment & $<$ \SI{0.1}{\second} \\
        PTf~\cite{liu2021paint} & Neural painter & fixed & \#layer & $<$ \SI{1}{\second} \\
        NPtr~\cite{zou2021stylized} & Neural painter & trainable & \#stroke & $<$ \SI{60}{\second}\\
    \hline        
    \end{tabular}
\end{table}

\subsection{Geometric Abstraction Control}
\label{subsec:geomabstractioncontrol}
\noindent The geometric abstraction of the image can be  controlled through the choice of abstraction method (\Cref{tab:segmentationstageproperties}), primitive shape, number of strokes, segments, or layers.

\begin{figure}[t]
    \centering
    \begin{subfigure}{.325\linewidth}
      \centering
      \includegraphics[width=\linewidth]{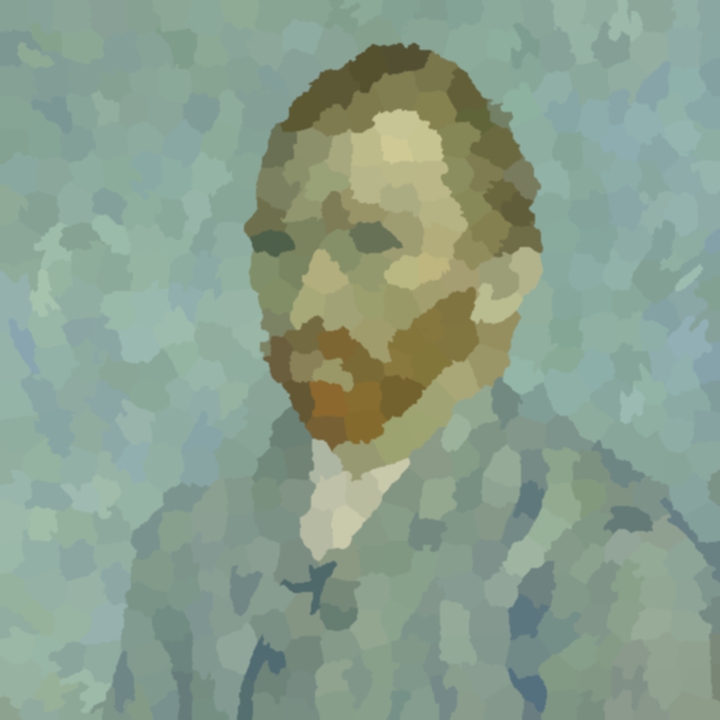}
      \caption{SLIC~\cite{achanta2012slic}}
      \label{fig:geometricstyles:SLIC}
    \end{subfigure}\hfill
    \begin{subfigure}{.325\linewidth}
        \centering
      \includegraphics[width=\linewidth]{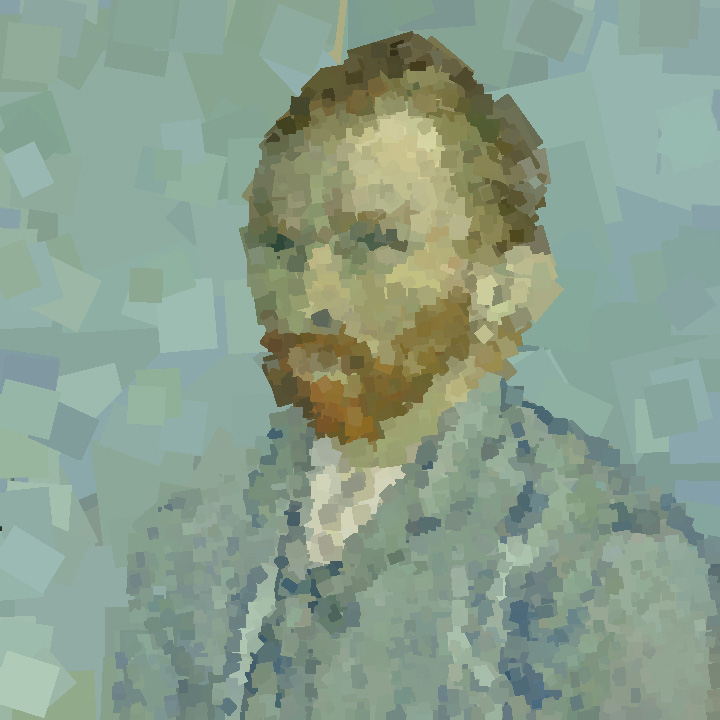}
      \caption{PaintTransf.~\cite{liu2021paint}}
      \label{fig:geometricstyles:adaptive}
    \end{subfigure}\hfill
    \begin{subfigure}{.325\linewidth}
      \centering
      \includegraphics[width=\linewidth]{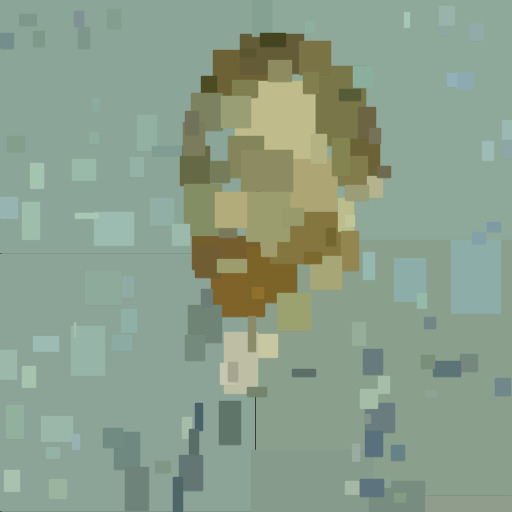}
      \caption{Neural Painter~\cite{zou2021stylized}}
      \label{fig:geometricstyles:8bit}
    \end{subfigure}
    \caption{Impact of various segmentation methods on the level of geometric abstraction in segmentation image $I_a$.  
    }
    \label{fig:geometricstyles}
\end{figure}

Generally, superpixel methods such as SLIC \cite{achanta2012slic} divide the image into segments that are clustered according to  color similarity and proximity, which align well with image regions and make them well-suited for use as an intermediate representation for interactive editing. These methods are primarily employed in our pipeline to represent uniform color regions of the image while abstracting away small-scale details (\Cref{fig:geometricstyles:SLIC}), which facilitates downstream editing of both color and fine-structure parameters separately (\Cref{subsec:manualediting}).

Stroke-based rendering methods on the other hand are better suited for stronger geometric abstraction. We integrate two neural painting methods (\Cref{tab:segmentationstageproperties}), which are trained to re-draw images using a particular shape primitive, such as a brush, rectangle, or circle. While the neural painter \cite{zou2021stylized} is able to create more abstract images, such as 8bit art (\Cref{fig:geometricstyles:8bit}) and optimize flexible shapes, the  PaintTransformer \cite{liu2021paint} uses fixed primitive-shapes, however, is fast enough to allow for immediate visual feedback and adjusting. 
To enable spatially varying levels of stroke details, we multiply the stroke decision confidence of the PaintTransformer with an optional level of detail parameter mask $P_S$, please refer to \cite{liu2021paint} for details on the method.
The mask $P_S$ can be acquired either manually or predicted, \eg using saliency or depth, as done in (\Cref{fig:geometricstyles:adaptive}) to threshold the foreground and background levels.

 Re-optimization of filter parameters adjusts the fine details to the shape and granularity of geometric elements and results in an integrated look, see  \Cref{fig:clipstyler1}, \Cref{fig:rw_compare_clip} and \Cref{fig:rw_compare_strotss}, as well as the supplemental material for examples of geometric shape and granularity variations.

\newcommand{\imagewithspy}[3]{
    \begin{tikzpicture}[every node/.style={inner sep=0,outer sep=0}]
        \begin{scope}[ 
			inner sep = 0pt,outer sep=0,spy using outlines={rectangle, red, magnification=4}
                ]
        \node (n0)  { \includegraphics[width=\linewidth]{#1}};
        \spy [red,size=1.1cm] on (#2,#3) in node[anchor=south east,inner sep=0pt] at (n0.south east);
        \end{scope}
    \end{tikzpicture}
}

\newcommand{\imagewithspytwo}[3]{
    \begin{tikzpicture}[every node/.style={inner sep=0,outer sep=0}]
        \begin{scope}[ 
			inner sep = 0pt,outer sep=0,spy using outlines={rectangle, red, magnification=3}
                ]
        \node (n0)  { \includegraphics[width=\linewidth]{#1}};
        \spy [red,size=1.0cm] on (#2,#3) in node[anchor=south east,inner sep=0pt] at (n0.south east);
        \end{scope}
    \end{tikzpicture}
}

 \begin{figure}[t]
    \centering
    \begin{subfigure}[b]{0.325\linewidth}%
        \imagewithspytwo{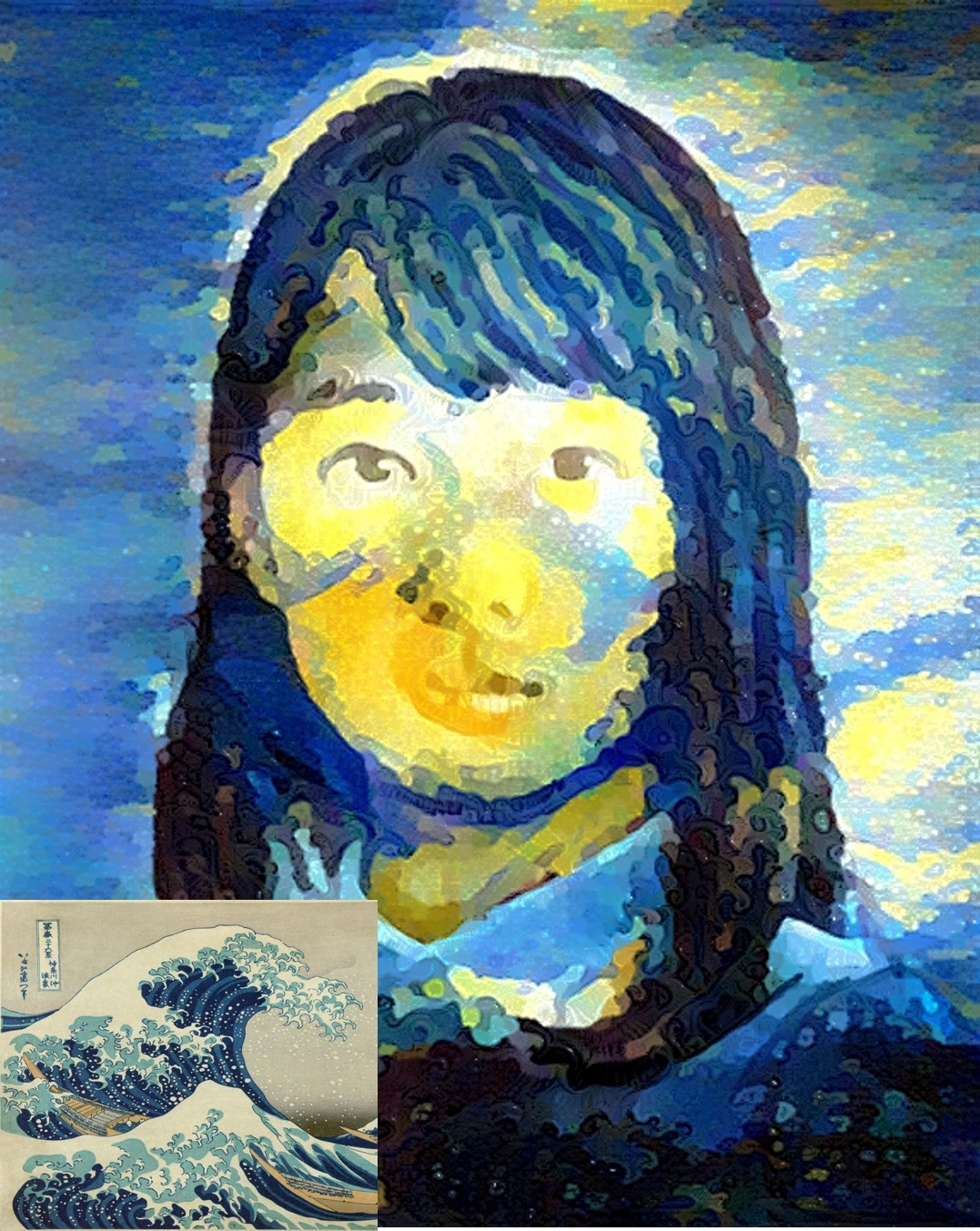}{-0.15cm}{0.25cm}%
        \caption{Great Wave}%
        \label{fig:reoptimnewstyle:greatwave}%
    \end{subfigure}\hfill
    \begin{subfigure}[b]{0.325\linewidth}
        \imagewithspytwo{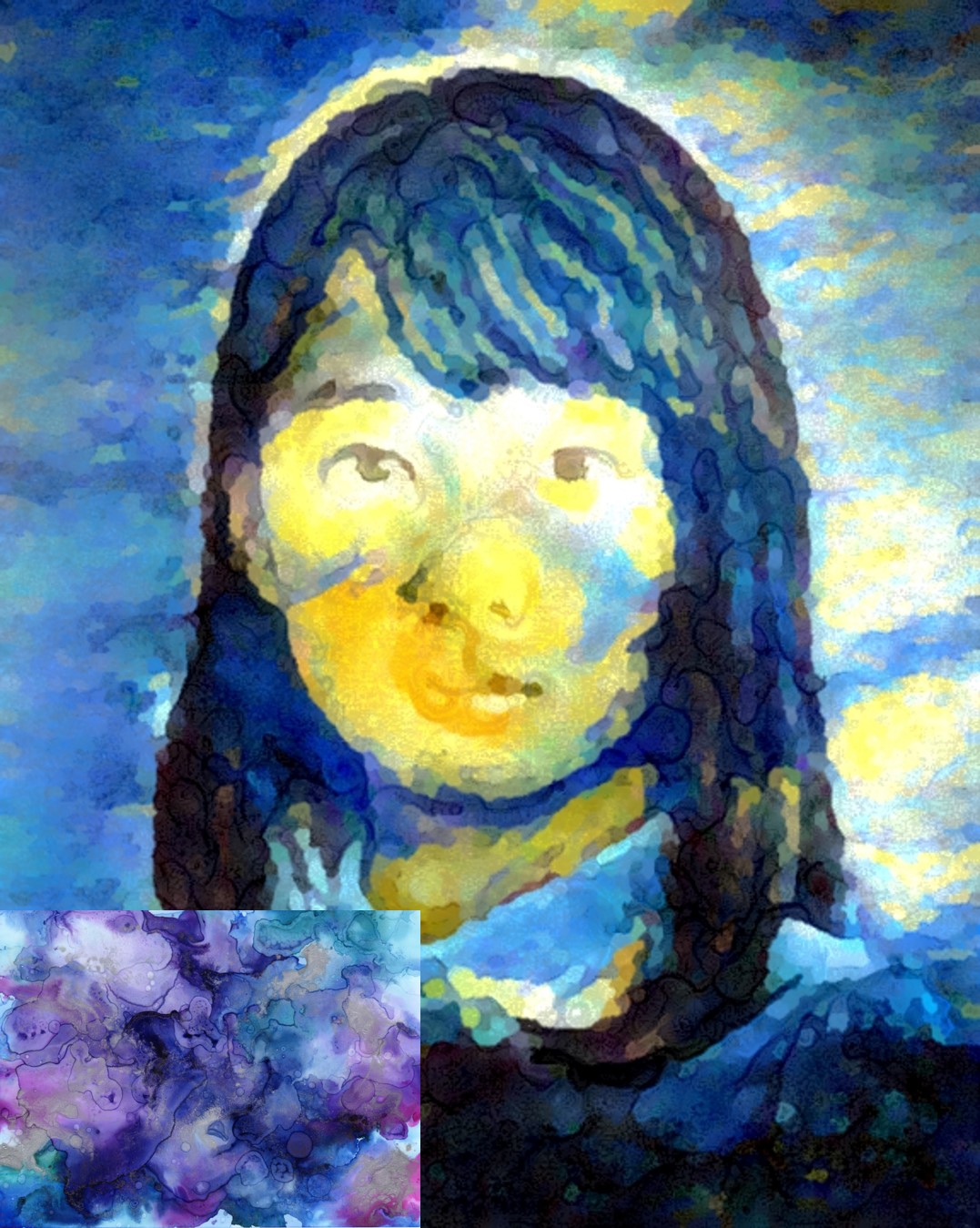}{-0.15cm}{0.25cm}%
        \caption{ink}
        \label{fig:reoptimnewstyle:ink}%
    \end{subfigure}\hfill
    \begin{subfigure}[b]{0.325\linewidth}%
        \imagewithspytwo{graphics/reoptim/bw\_wave.jpg}{-0.15cm}{0.25cm}%
        \caption{vertical waves}%
        \label{fig:reoptimnewstyle:vertwave}%
    \end{subfigure}
    \caption{Parametric texture style transfer using example images. The decomposed parameters of an image, created by \acs{NST} \cite{gatys2016image} using the \enquote{Starry Night} style, are retargeted to a new texture style using the shown style example images and are optimized with the STROTSS \cite{kolkin2019style} loss.}
    \label{fig:reoptim_newstyle}
\end{figure}

\begin{figure}[t]
    \centering
    \begin{subfigure}[t]{0.325\linewidth}%
        \imagewithspytwo{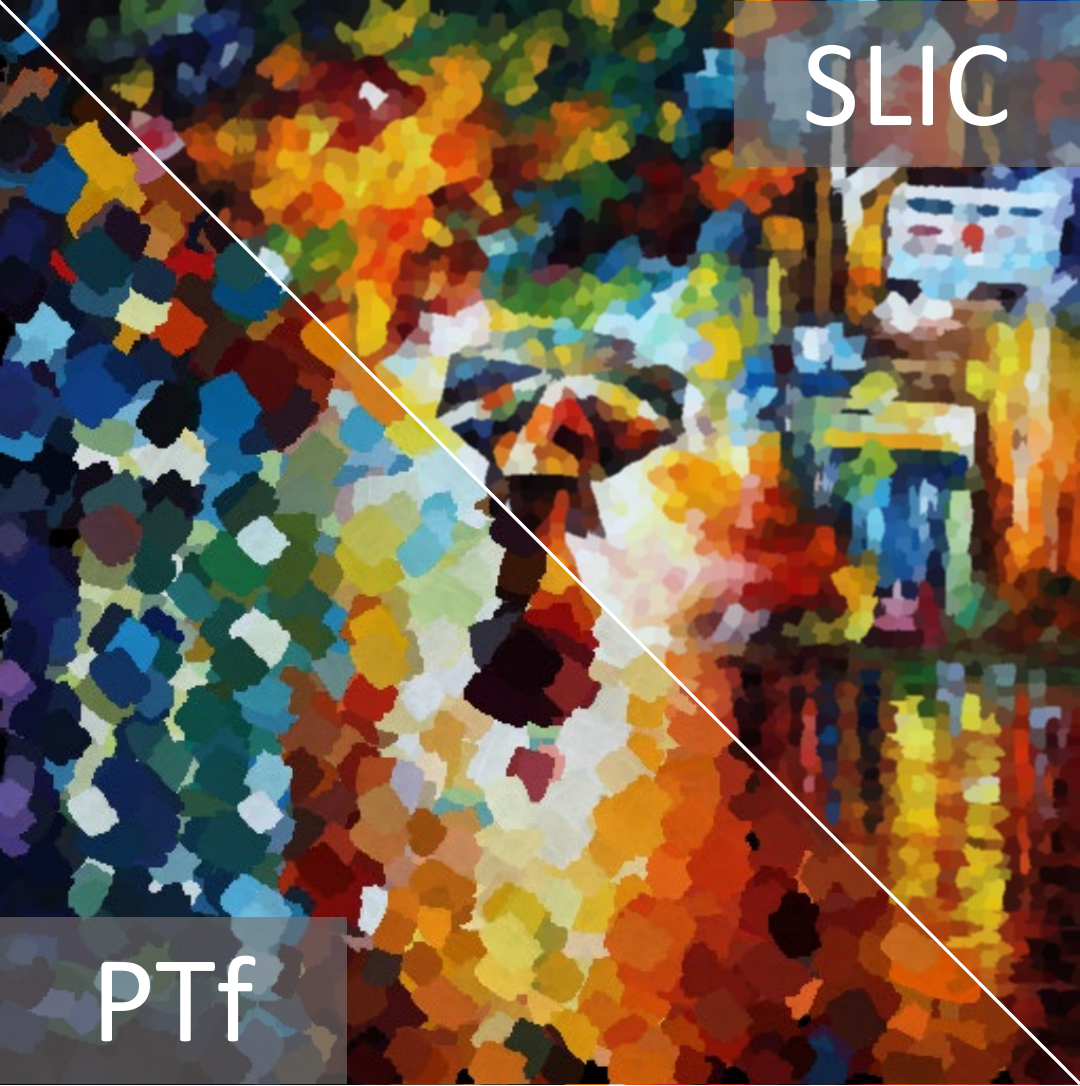}{-0.25cm}{0.25cm}%
        \caption{$S$ output}%
        \label{fig:clipstyler1:segmented}%
    \end{subfigure}\hfill
    \begin{subfigure}[t]{0.325\linewidth}
       \imagewithspytwo{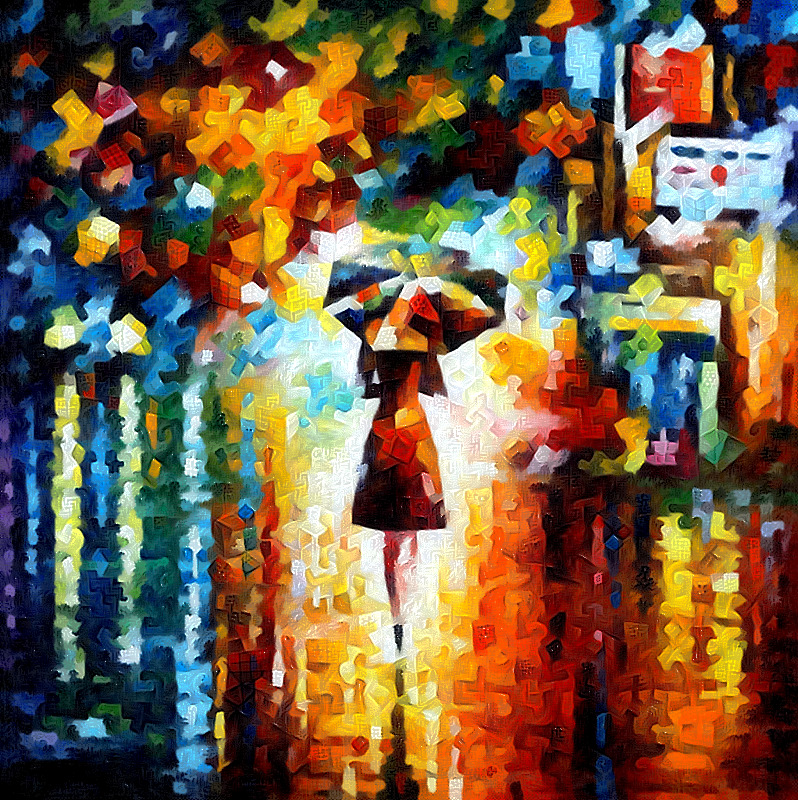}{-0.25cm}{0.25cm}%
        \vspace{-0.365cm}\caption{$S$ = 5000 segments}%
        \label{fig:clipstyler1:seg_cubistic}
    \end{subfigure}\hfill
    \begin{subfigure}[t]{0.325\linewidth}%
        \imagewithspytwo{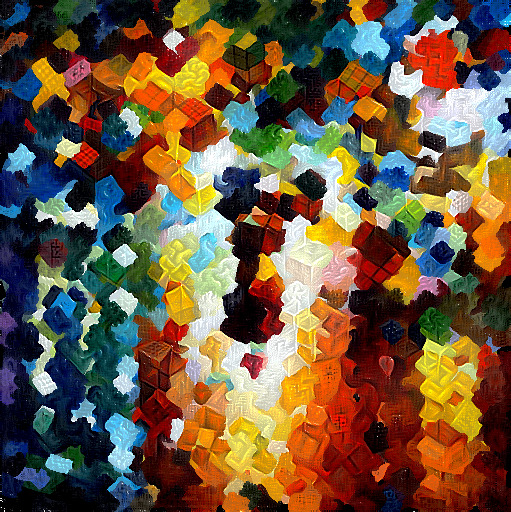}{-0.25cm}{0.25cm}
        \vspace{-0.365cm}\caption{$S$ = PTf-brush}%
        \label{fig:clipstyler1:ptt_cubistic}%
    \end{subfigure}\hfill
    \caption{Re-optimization using CLIPstyler \cite{kwon2022clipstyler} loss with different first stages~$S$. The input  painting "rain princess" (teaser) is optimized by the prompt "a cubistic painting". In (\subref{fig:clipstyler1:seg_cubistic})  we use a fine segmentation (5000 segments) and in (\subref{fig:clipstyler1:ptt_cubistic}) the PaintTransformer (PTf) \cite{liu2021paint} with brush-shapes as the first stage.}
    \label{fig:clipstyler1}
\end{figure}

\begin{figure}
    \centering
    \begin{subfigure}{0.495\linewidth}
    \setbox1=\hbox{\includegraphics[width=\linewidth,height=3cm]{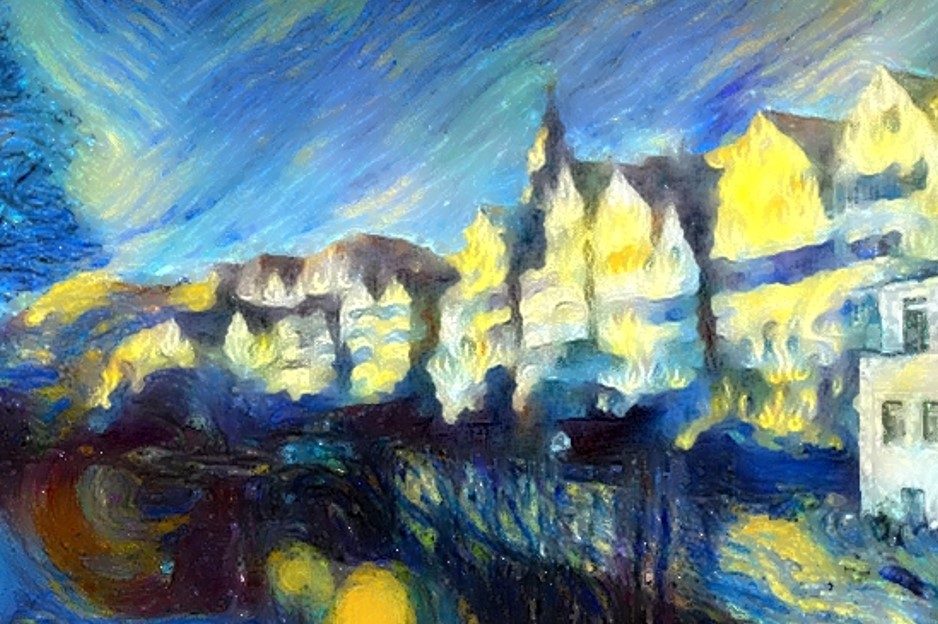}}
      \includegraphics[width=\linewidth,height=3cm]{graphics/saliency_interpolation/tub_figre_saliency_cropped.jpg}\llap{\makebox[\wd1][l]{\includegraphics[trim={1cm 1cm 0.2cm 0cm},clip,height=0.85cm]{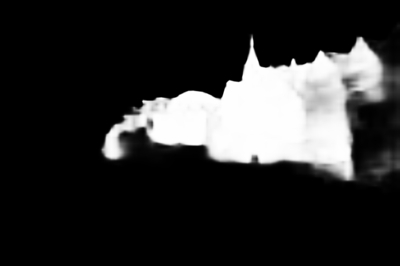}}}
    \end{subfigure}\hfill
    \begin{subfigure}{0.495\linewidth}
    \setbox2=\hbox{\includegraphics[width=\linewidth,height=3cm]{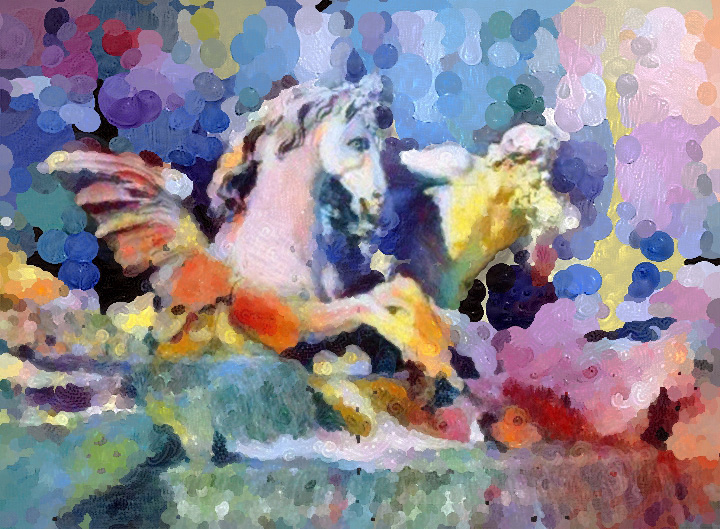}}
          \includegraphics[trim={0cm 2cm 0cm 0.6cm},clip,width=\linewidth,height=3cm]{graphics/interpolation/trevi_drops_depth_womanwithhat.jpg}\llap{\makebox[\wd2][l]{\includegraphics[trim={0cm 0.2cm 0cm 0.2cm},clip,height=0.85cm]{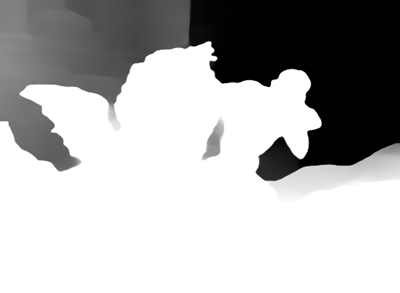}}}
    \end{subfigure}
    \caption{Using saliency (left) and depth (right) to interpolate parameters of the original \acs{NST} with the prompts \enquote{fire} (left) and \enquote{drops} (right).
    }
    \label{fig:interpolation}
\end{figure}

\subsection{Texture Style Transfer}
\label{subsec:styledetailvariants}
\noindent Parameters $P_M$ can be optimized using different losses to adapt the style details to new targets for varied artistic expression. As such, given a segmented image, we replace $\ell_1$ matching of $I_t$ with directly optimizing a style transfer loss in \Cref{eq:overall}. The optimization may either be initialized with a previous parameter decomposition to fine-tune the texture style or start from empty parameter masks.

\paragraph{Image-based Parameter Style Transfer}
\Cref{fig:reoptim_newstyle} exemplary shows results of optimizing the self-transport style loss (STROTTS \cite{kolkin2019style}) to adapt the parameter-decomposition of a stylized image to different texture styles. Notice how the overall structure and colors are maintained while the fine details and texture are adapted.

\paragraph{Text-based Parameter Style Transfer}
Text-based representation of style allows for more freedom of control as reference images are not required.
We adapt the losses from CLIPstyler \cite{kwon2022clipstyler} for text-based style transfer, however, in contrast to their method we do not train a \ac{CNN}, but directly optimize the effect parameters. In particular, we adapt their patch-based and directional loss---in most cases, the use of content losses is not necessary as the filter pipeline inherently retains the major content structures and colors.
\Cref{fig:clipstyler1} shows adapted style details of a real-world painting using prompts, with additional prompts results shown in the supplemental material. It can be observed that colors remain faithful to the input while details adapt to the prompt and are adjusted to the size of segments.

\begin{figure}[t]
    \centering
    \begin{subfigure}[b]{0.325\linewidth}%
        \imagewithspy{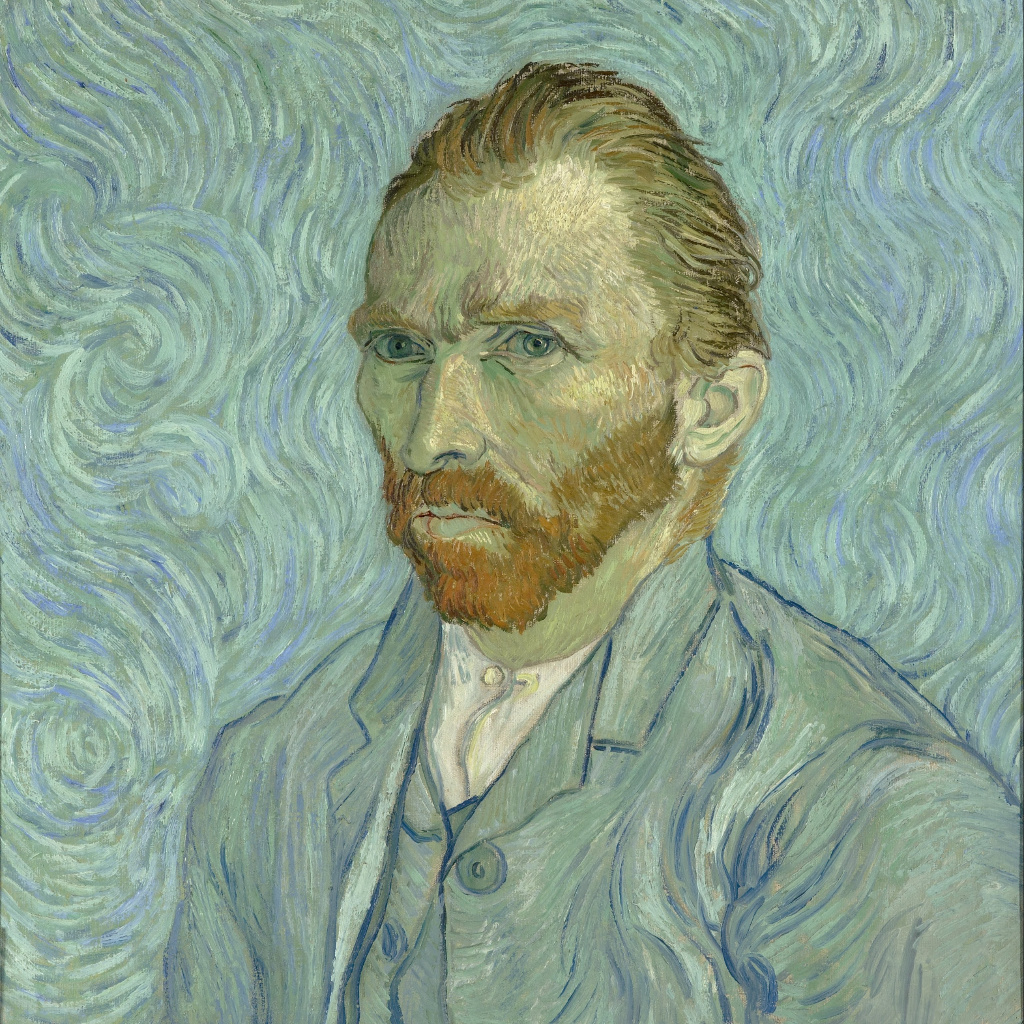}{-0.08cm}{0.45cm}%
        \caption{w/o detail edits}%
        \label{fig:globaledits:original_matched}%
    \end{subfigure}\hfill
    \begin{subfigure}[b]{0.325\linewidth}
        \imagewithspy{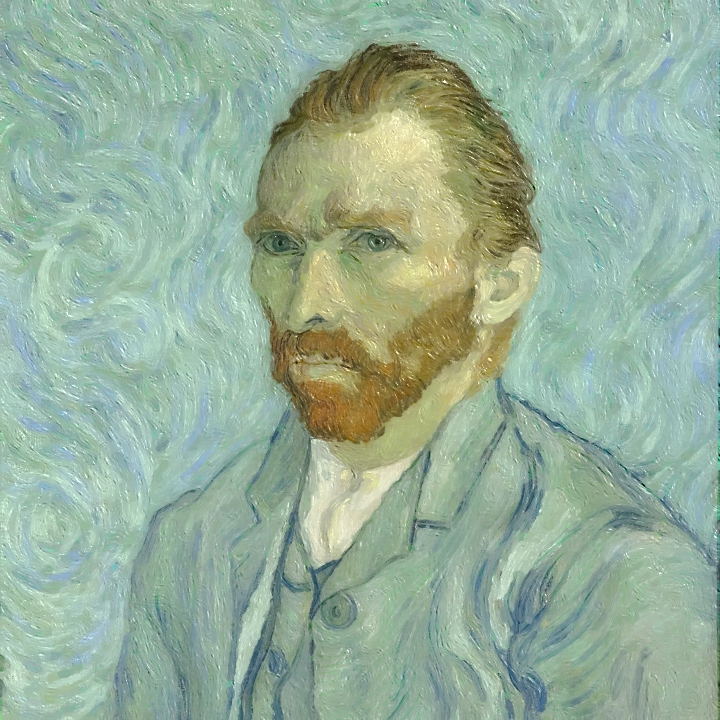}{-0.08cm}{0.45cm}%
        \caption{+bumpiness}%
        \label{fig:globaledits:original_increased_bumped}%
    \end{subfigure}\hfill
    \begin{subfigure}[b]{0.325\linewidth}%
        \imagewithspy{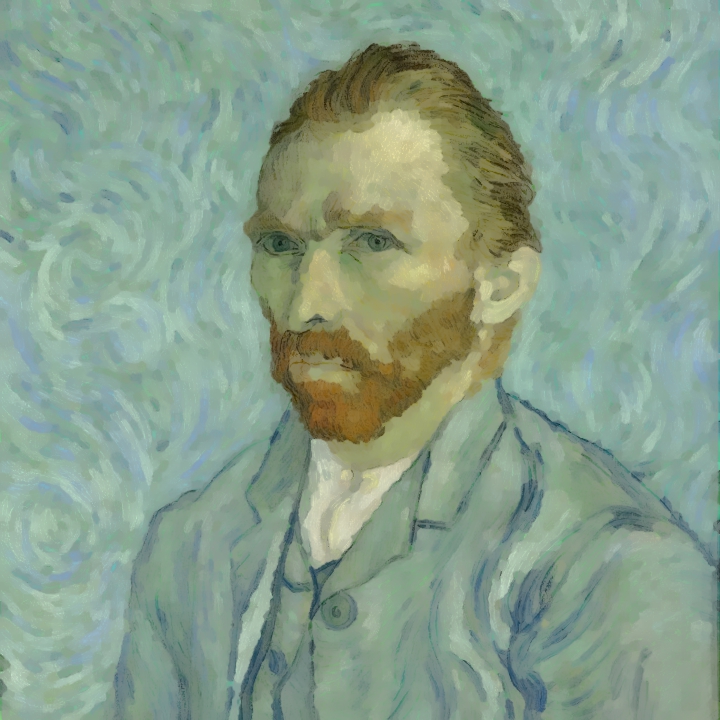}{-0.08cm}{0.45cm}%
        \caption{-bumpiness}%
        \label{fig:globaledits:original_decreased_bumped}%
    \end{subfigure}%
    \caption{Global parameter editing. Values of the bump-map scale $P_M$ are increased (\subref{fig:globaledits:original_increased_bumped}) and decreased (\subref{fig:globaledits:original_decreased_bumped}) uniformly.}
    \label{fig:global_edits}
\end{figure}

\begin{figure}
\centering
\begin{subfigure}{.325\linewidth}
  \centering
  \includegraphics[trim={0cm 1cm 0cm 0.5cm}, clip, width=\linewidth]{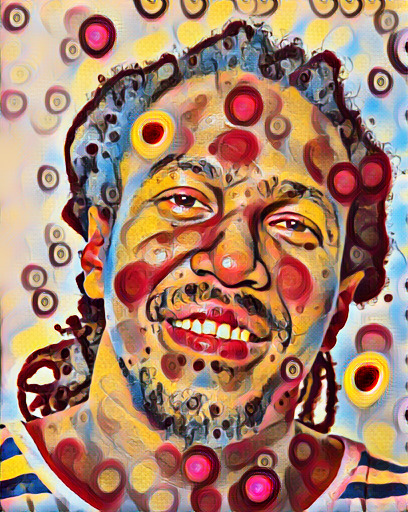}
  \caption{NST \cite{johnson2016perceptual} result}
  \label{fig:stylizationcorrection:sub1}
\end{subfigure}\hfill%
\begin{subfigure}{.325\linewidth}
  \centering
  \includegraphics[trim={0cm 1cm 0cm 0.5cm}, clip, width=\linewidth]{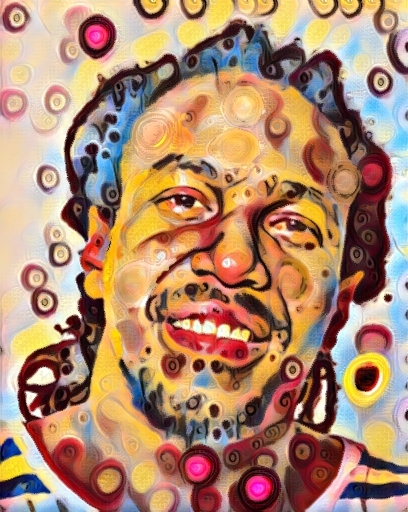}
  \caption{Color transfer}
  \label{fig:stylizationcorrection:sub2}
\end{subfigure}
\begin{subfigure}{.325\linewidth}
  \centering
  \includegraphics[trim={0cm 1cm 0cm 0.5cm}, clip, width=\linewidth]{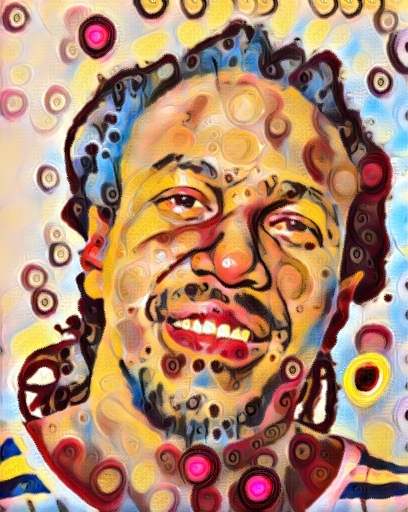}
  \caption{Repredicted}
  \label{fig:stylizationcorrection:sub3}
\end{subfigure}
\caption{Correcting artifacts in \ac{NST} images using a style-transfer \ac{PPN}. Red dots in the face are removed using histogram matching of segments to other colors in the face. The $\text{PPN}_{sst}$ re-prediction (candy style) reintegrates the manual edits seamlessly into the overall style.}
\label{fig:stylizationcorrection}
\end{figure}

\subsection{Parameter Editing}
\label{subsec:manualediting}
\noindent One major advantage of our filter parameter-based approach is that various stylistic aspects can be edited interactively, using both global and local parameter tuning. 

\begin{description}
\item[Global Parameter Editing:]
Parameters masks of the pipeline presented in \Cref{subsec:diffilterstage} can be easily modified after optimization or prediction by uniformly adding or subtracting values on the parameter masks. \Cref{fig:global_edits} shows the result of globally increasing and decreasing the bump mapping parameter by 50\%.
\item[Local Parameter Editing:]
Parameters can be  edited locally thro\-ugh adjustment of the parameter masks using brush meta\-phors. For example, contours can be locally increased in the eye and mouth region of a portrait, to make them stand out (see supplementary video).
\item[Parameter Interpolation:]
Parameters can be interpolated locally using binary or continuous-valued blending masks. For automated stylization, we test saliency prediction (using LPF \cite{wei2020label}) and depth prediction (using DPT \cite{ranftl2021vision}) to interpolate between two parameter sets (\Cref{fig:interpolation}). 
\end{description}

\subsection{Correction of Style Transfer Artifacts}
\noindent The superpixels can be used for efficient selection and editing of regions in the stylization generated  by  methods such as \ac{NST}, particularly for correcting localized artifacts and tailoring local style elements to personal aesthetic preferences. The uniformly colored segments can be easily selected and their colors adjusted while the fine-structural texture is retained.
For example, a selected region of superpixels can be color-matched to a different region using histogram matching, which can be used to remove unwanted stylization patterns or colors. In \Cref{fig:stylizationcorrection}, a single-style \ac{PPN} re-predicts the style's texture during the editing of segment colors to adapt the textural patterns to the new segment colors and structures in real time, enabling an integrated and interactive editing workflow (see supplementary material).
\section{Results and Discussion}

\subsection{Qualitative Comparisons}
\begin{figure*}[tb]
    \begin{subfigure}{0.165\textwidth}%
          \includegraphics[width=\textwidth]{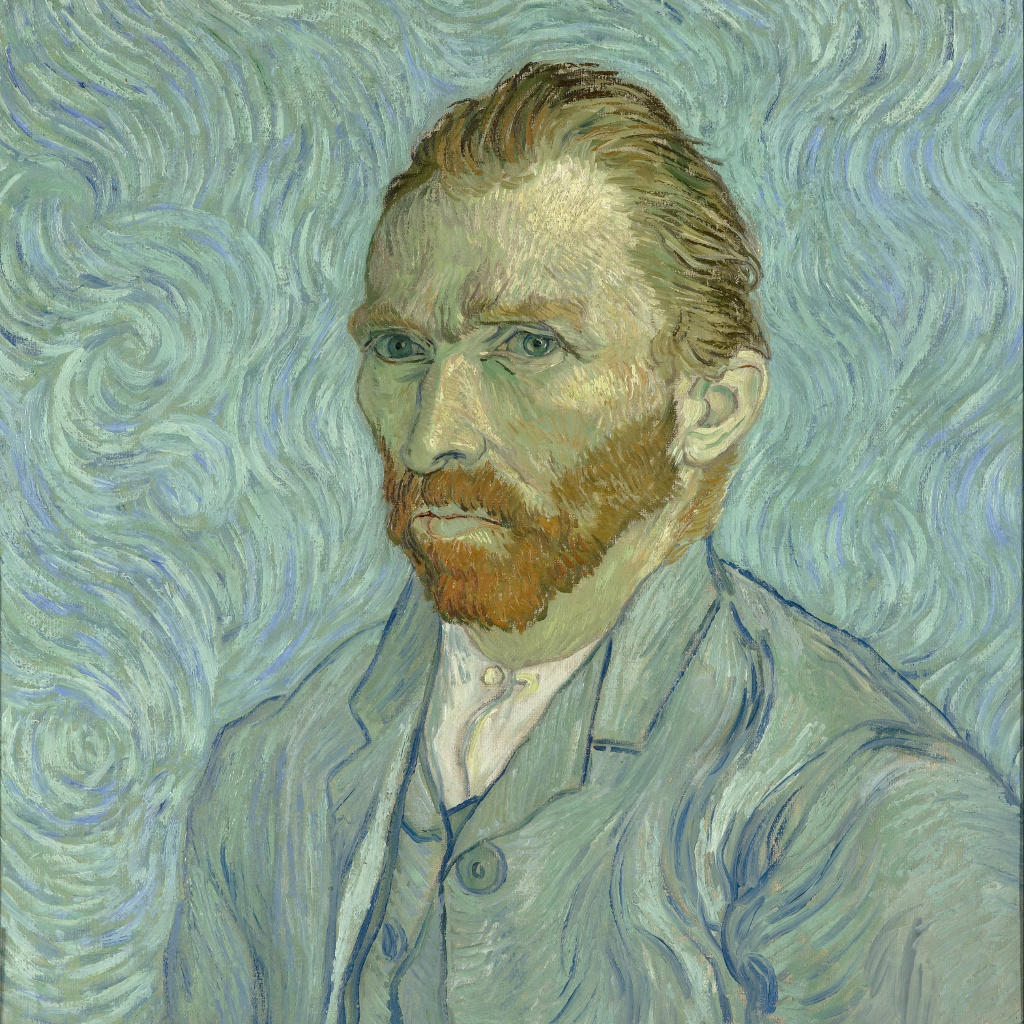}%
      \end{subfigure}\hfill
      \begin{subfigure}{0.165\textwidth}%
          \includegraphics[width=\textwidth]{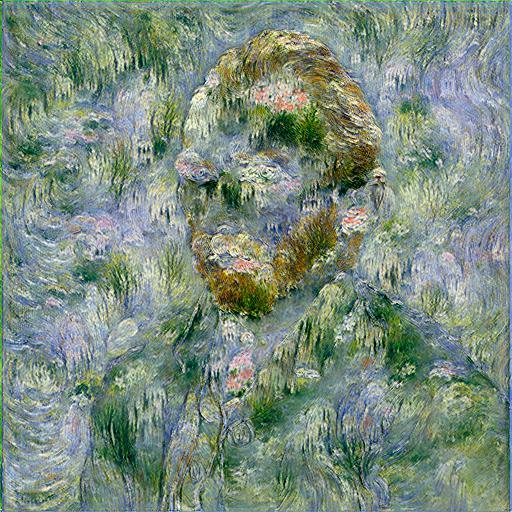}%
      \end{subfigure}\hfill
    \begin{subfigure}{0.165\textwidth}%
      \includegraphics[width=\textwidth]{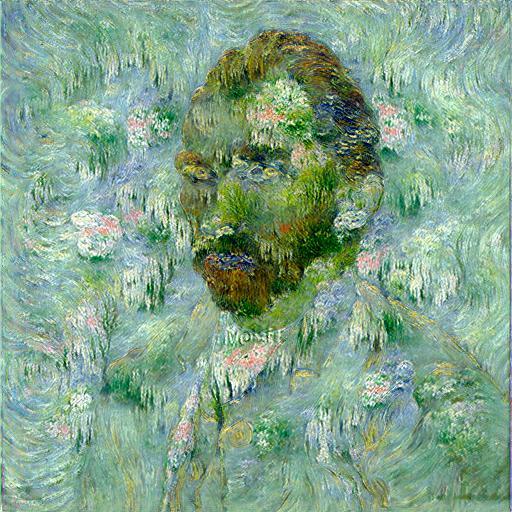}%
    \end{subfigure}\hfill
    \begin{subfigure}{0.165\textwidth}%
          \includegraphics[width=\textwidth]{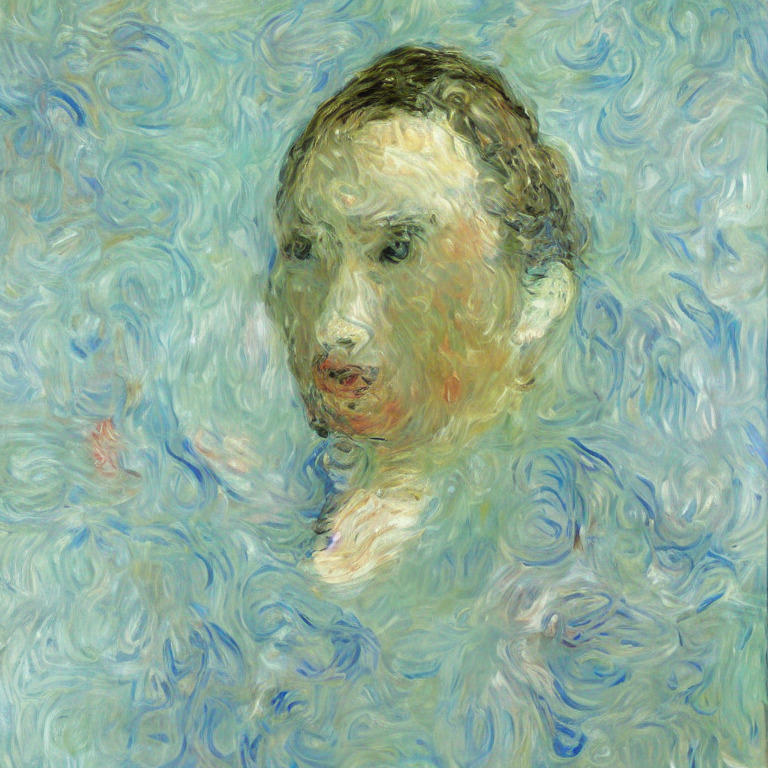}%
      \end{subfigure}\hfill
    \begin{subfigure}{0.165\textwidth}%
      \includegraphics[width=\textwidth]{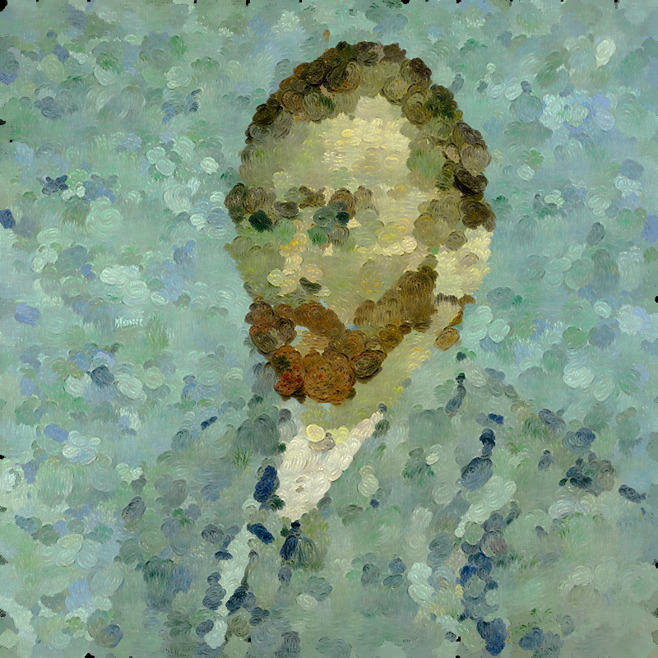}%
    \end{subfigure}\hfill 
    \begin{subfigure}{0.165\textwidth}%
      \includegraphics[width=\textwidth]{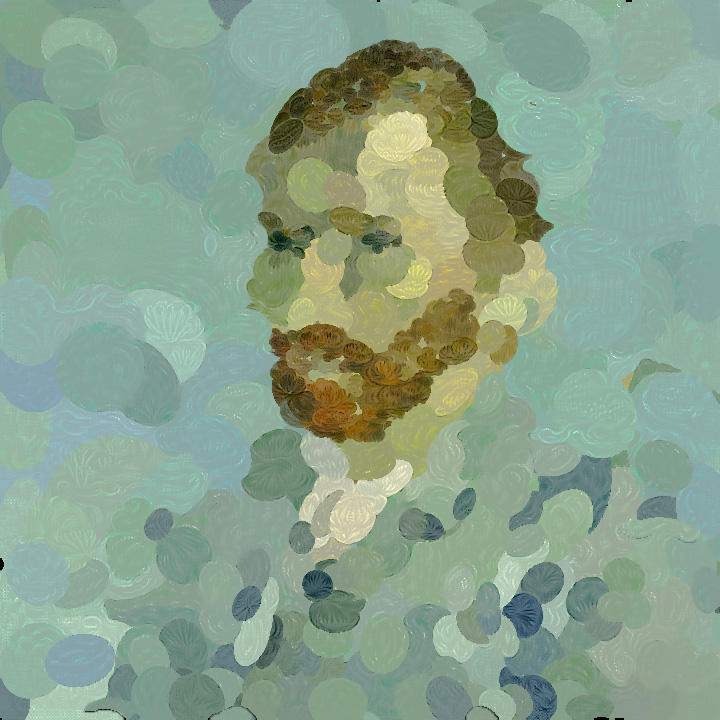}%
    \end{subfigure}\\
    \begin{subfigure}{0.165\textwidth}%
        \includegraphics[width=\textwidth]{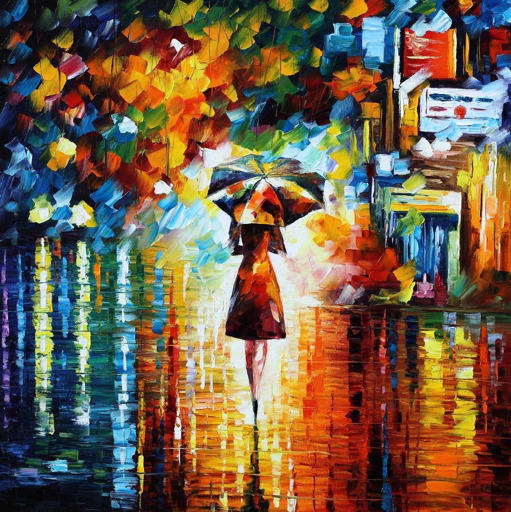}%
        \caption{Original}%
       \label{fig:rw_compare:orig}%
    \end{subfigure}\hfill
    \begin{subfigure}{0.165\textwidth}%
          \includegraphics[width=\textwidth]{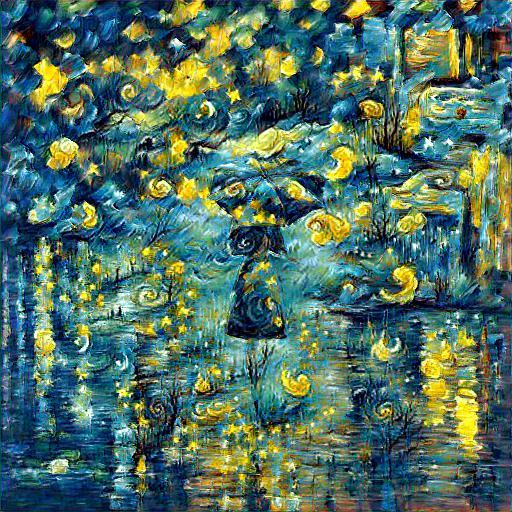}%
          \caption{CLIPstyler \cite{kwon2022clipstyler}}%
        \label{fig:rw_compare:clipstyler}%
      \end{subfigure}\hfill
    \begin{subfigure}{0.165\textwidth}%
      \includegraphics[width=\textwidth]{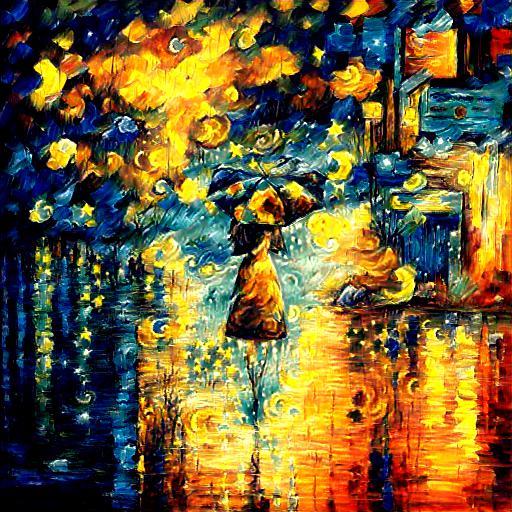}%
      \label{fig:rw_compare:clipstyler-hist}%
  \caption{CLIPstyler-Hist}%
    \end{subfigure}\hfill
    \begin{subfigure}{0.165\textwidth}%
          \includegraphics[width=\textwidth]{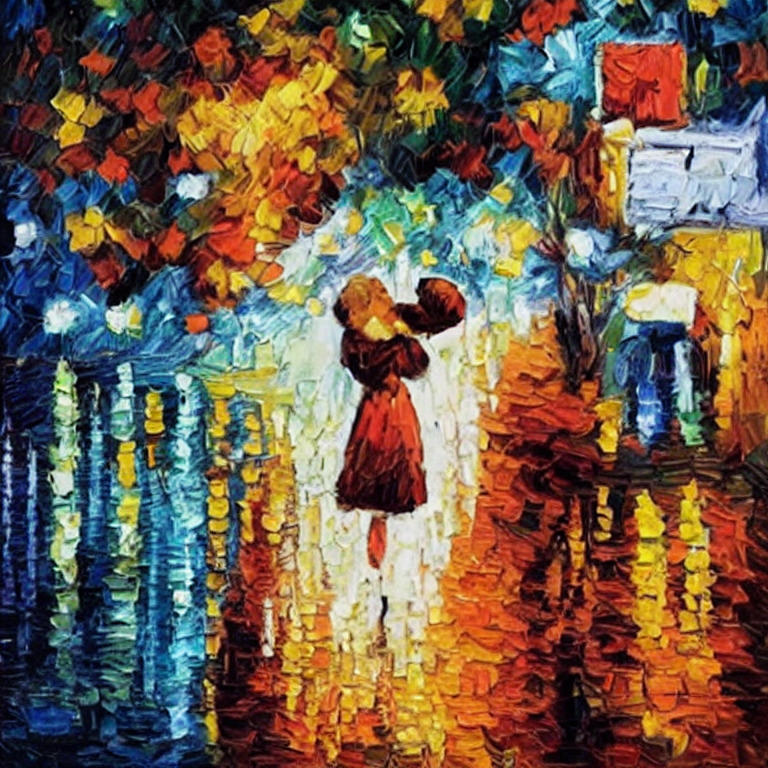}%
          \label{fig:rw_compare:stablediffusion}%
      \caption{Stable Diffusion \cite{rombach2022high}}%
      \end{subfigure}\hfill
    \begin{subfigure}{0.165\textwidth}%
      \includegraphics[width=\textwidth]{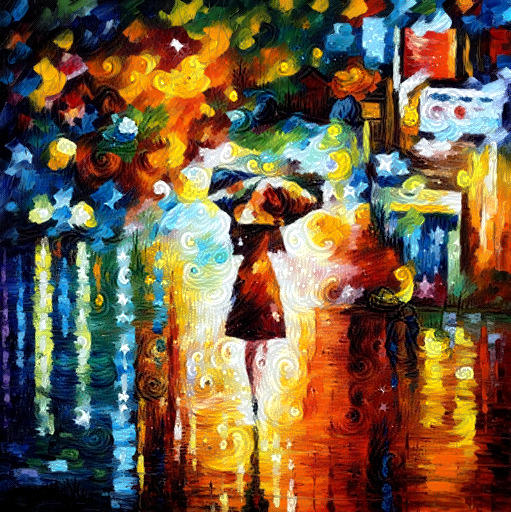}%
      \label{fig:rw_compare:ours_fine}
  \caption{Ours (fine)}%
    \end{subfigure}\hfill 
    \begin{subfigure}{0.165\textwidth}%
        \includegraphics[width=\textwidth]{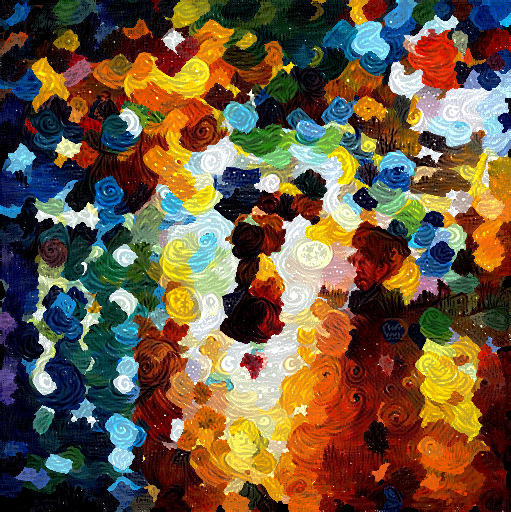}%
        \label{fig:rw_compare:ours_coarse}%
    \caption{Ours (coarse)}%
    \end{subfigure}%
    \caption{Comparisons to related methods for text-based generation. Text prompts, per row, are 1) "round brushstrokes in the style of monet", 2) "starry night".  Please zoom in to compare details. }
    \label{fig:rw_compare_clip}
  \end{figure*}

\begin{figure*}[tb]
    \begin{subfigure}{0.165\textwidth}%
          \includegraphics[width=\textwidth]{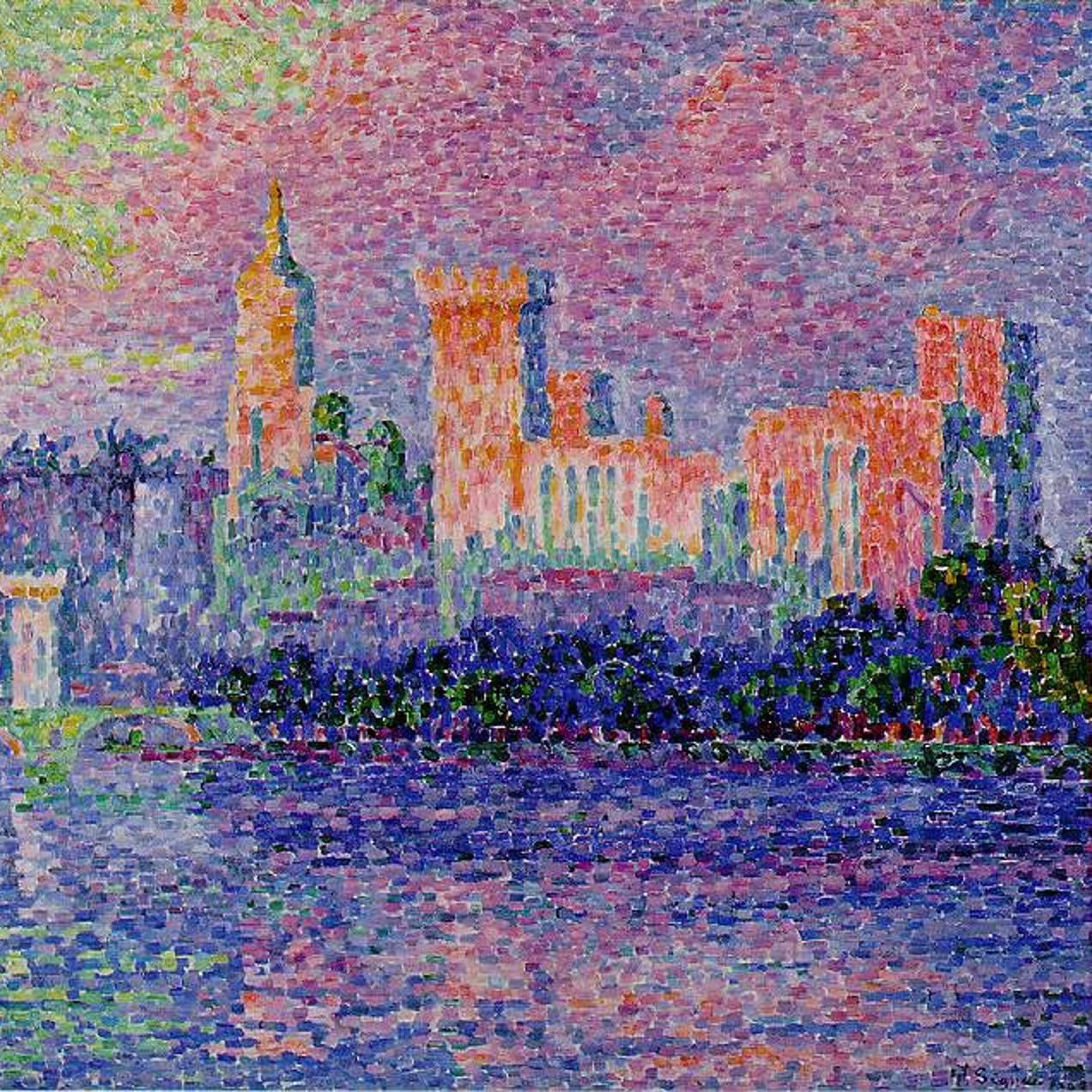}
      \end{subfigure}\hfill
      \begin{subfigure}{0.165\textwidth}%
          \includegraphics[width=\textwidth]{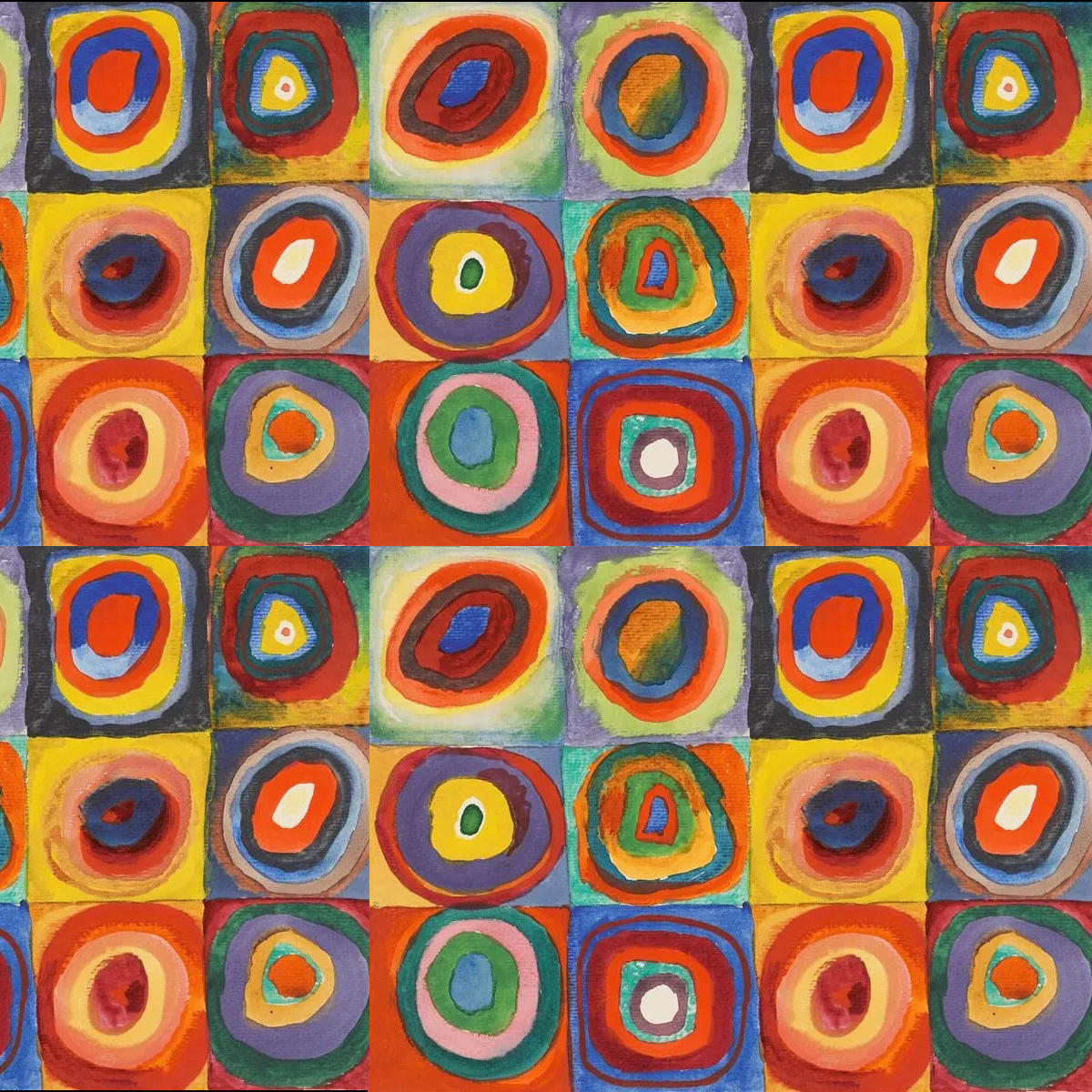}
      \end{subfigure}\hfill
    \begin{subfigure}{0.165\textwidth}%
      \includegraphics[width=\textwidth]{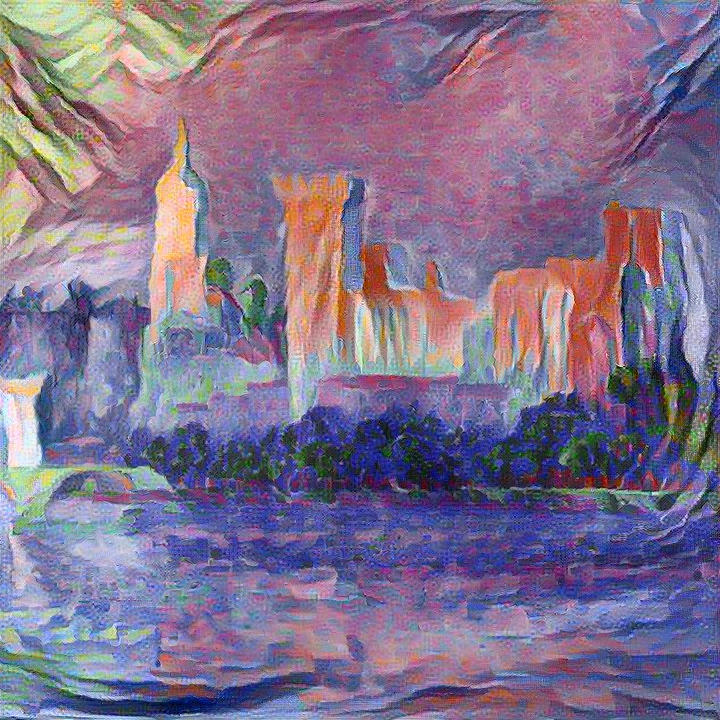}
    \end{subfigure}\hfill
    \begin{subfigure}{0.165\textwidth}%
          \includegraphics[width=\textwidth]{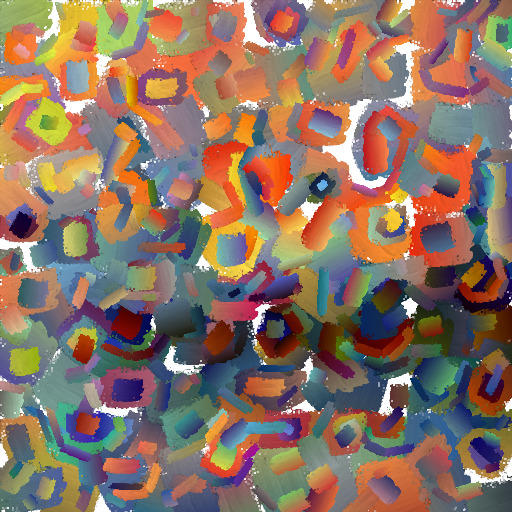}
      \end{subfigure}\hfill
    \begin{subfigure}{0.165\textwidth}%
      \includegraphics[width=\textwidth]{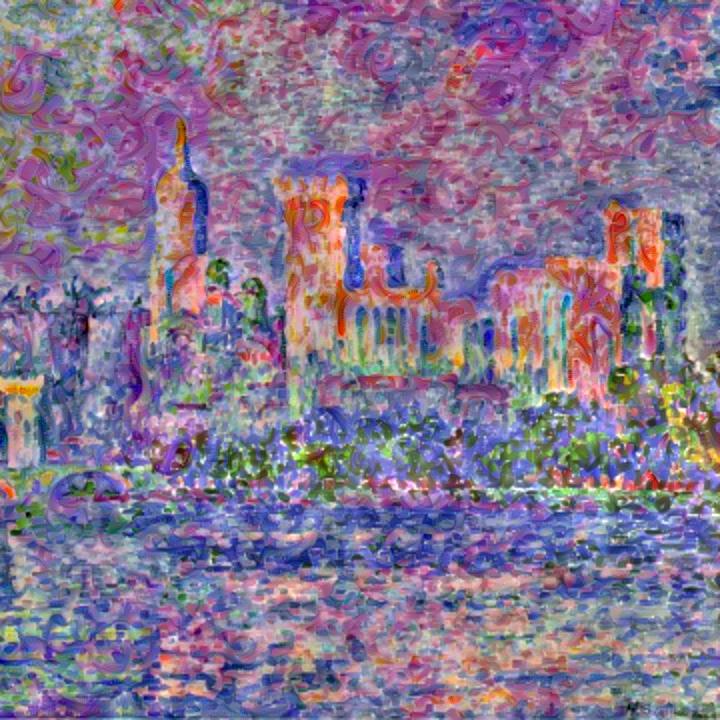}
    \end{subfigure}\hfill 
    \begin{subfigure}{0.165\textwidth}%
      \includegraphics[width=\textwidth]{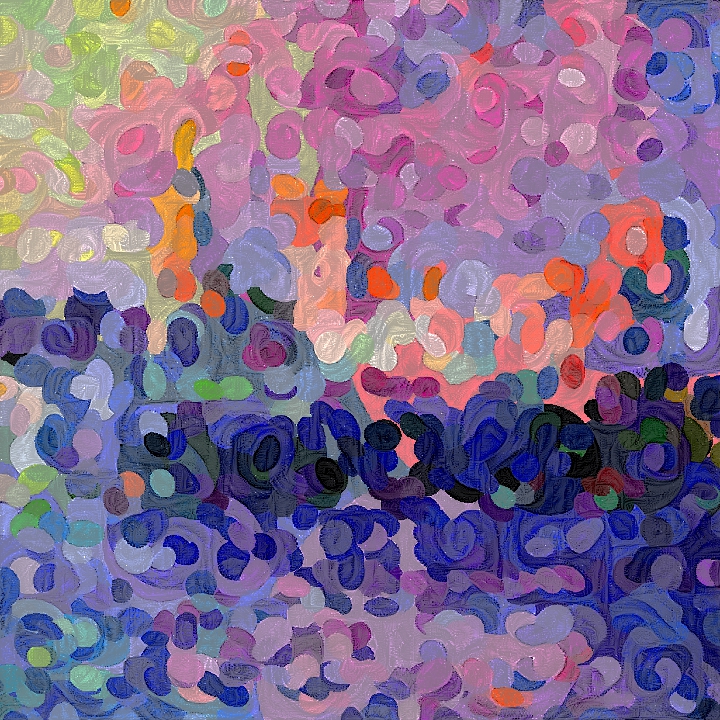}
    \end{subfigure}\\
    \begin{subfigure}{0.165\textwidth}%
      \includegraphics[width=\textwidth]{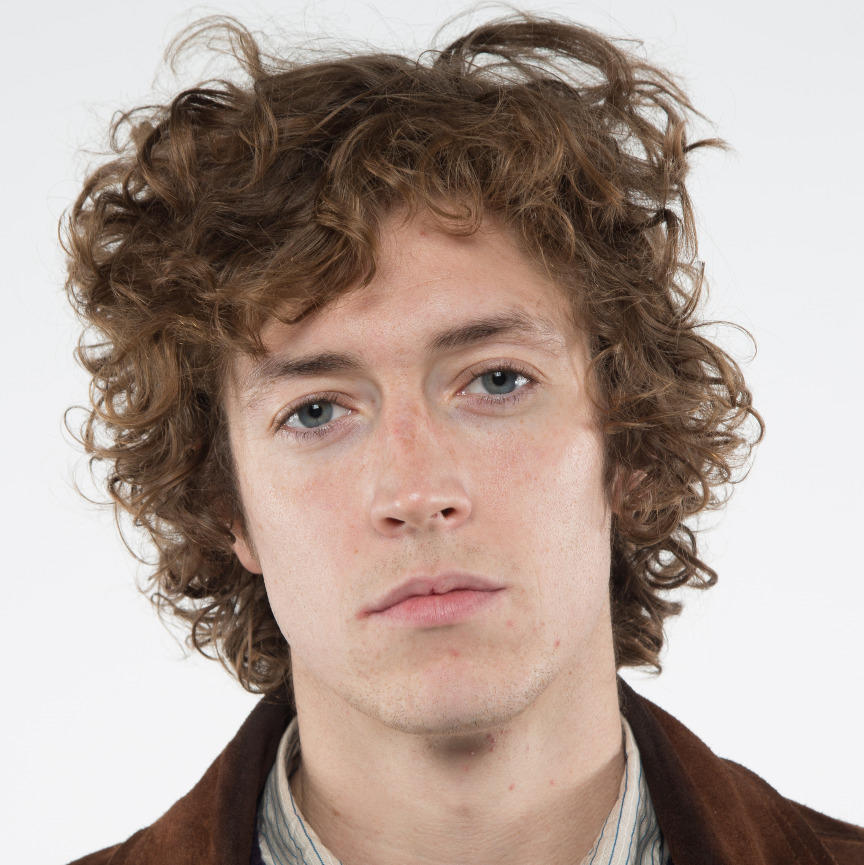}
      \caption{Input}
    \end{subfigure}\hfill
      \begin{subfigure}{0.165\textwidth}%
          \includegraphics[width=\textwidth]{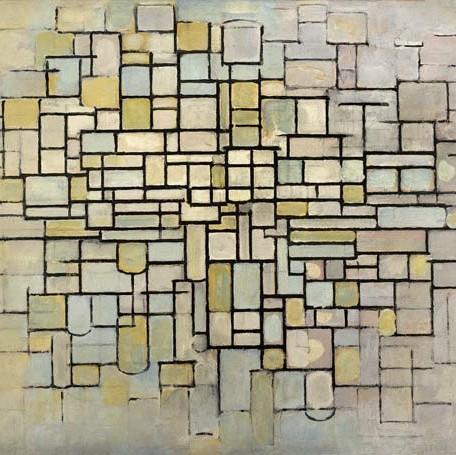}
          \caption{Style}
      \end{subfigure}\hfill
    \begin{subfigure}{0.165\textwidth}%
      \includegraphics[width=\textwidth]{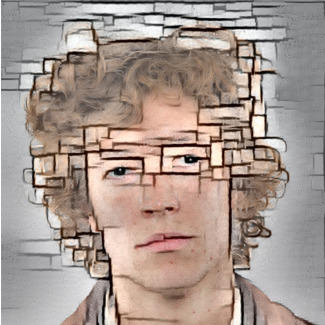}
      \caption{\footnotesize Color-pres. NST \cite{GatysEBHS17}}
    \end{subfigure}\hfill
      \begin{subfigure}{0.165\textwidth}%
      \includegraphics[width=\textwidth]{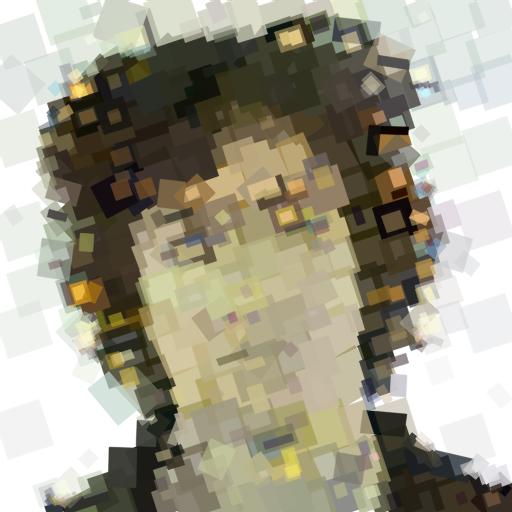}
          \caption{SNP \cite{zou2021stylized}}
          
      \end{subfigure}\hfill
    \begin{subfigure}{0.165\textwidth}%
          \includegraphics[width=\textwidth]{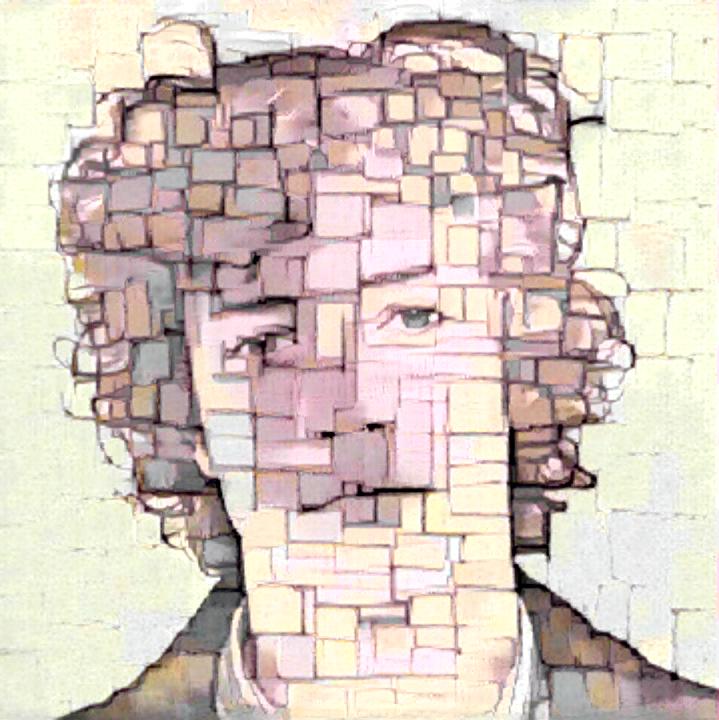}
          \caption{STROTSS~\cite{kolkin2019style}-Hist}
      \end{subfigure}\hfill
    \begin{subfigure}{0.165\textwidth}%
      \includegraphics[width=\textwidth]{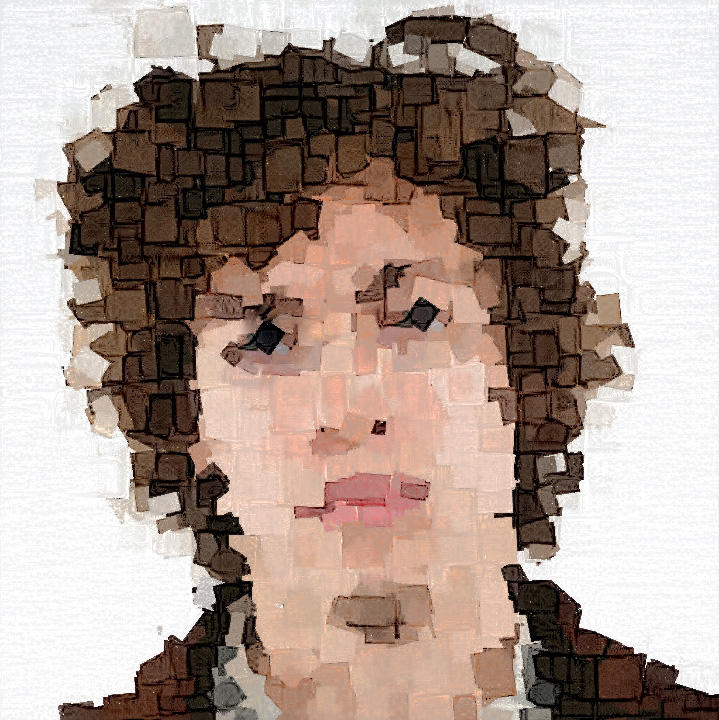}
          \caption{Ours}
    \end{subfigure}
    \caption{Comparisons to related methods for image-based style transfer. Please zoom in to compare details.}
    \label{fig:rw_compare_strotss}
  \end{figure*}

\noindent We compare our texture style transfer against other methods for example-based texture control on their ability to adapt texture style while retaining the overall composition and colors of the input image.
\Cref{fig:rw_compare_clip} compares our text-based filter-optimization against CLIPstyler \cite{kwon2022clipstyler} and img2img Stable Diffusion \cite{rombach2022high}. As CLIPstyler also adapts the colors, we add a histogram-matching loss \cite{jonschkowski2016end} (CLIPstyler-Hist) for a fair comparison. It is visible that CLIPstyler introduces new content structures and the strength of stylization is strongly dependent on the used text prompt, furthermore, colors deviate from the original even with histogram losses. Stable Diffusion, on the other hand, introduces strong variations from the original image content.
\Cref{fig:rw_compare_strotss} compares our image-based filter optimization against STROTTS~\cite{kolkin2019style} with histogram-matching loss \cite{jonschkowski2016end}, Gatys \ac{NST} \cite{gatys2016image} with color-preserving loss \cite{GatysEBHS17}, and the \ac{SNP} style transfer \cite{liu2021paint}. Similarly to the CLIP-optimized parameters, our method is able to preserve more colors and is spatially more homogeneous than the preceding methods.

\subsection{Quantitative Comparisons}
\label{subsec:quantativecomparisons}

\paragraph{Style-matching Capabilities}
\Cref{tab:quantiveresults} compares style-matching capabilities of our parameter prediction and optimization to their respective baselines. While the single-style \ac{NST}~\cite{johnson2016perceptual} is better in terms of content preservation compared to $\text{PPN}_\mathit{sst}$, which is expected as the \ac{PPN} operates on the segmented stylized image, the style preservation is close. For the arbitrary style \ac{PPN}, the content preservation is almost on par with its baseline while even improving on the style loss score. For optimization, we measure the $\ell_1$ to its target (generated by STROTTS \cite{kolkin2019style}), and achieve very close matching. Compared to the stylization pipelines employed by \cite{lotzsch2022wise}, we achieve similar values, even though their method does not use loss constraints on the parameter masks. 

\begin{table}
\centering
\caption{Comparison to baselines. To assess decomposition quality, we measure content loss $\mathcal{L}_c$ and style loss $\mathcal{L}_s$ \cite{gatys2016image} on 10 \ac{NST} styles and 20 content images (see suppl. material).  $\text{PPN}_\mathit{sst}$ and  $\text{PPN}_\mathit{arb}$ are compared to their pixel-predicting baselines. Optimization is futhermore measured in $\ell_1$ distance to its target and compared with the pipelines of Loetzsch \etal \cite{lotzsch2022wise}. We use $\mathcal{L}_\mathit{TV}$ with  $ \lambda_{tv}=0.2$. The segmentation stage $S$ uses SLIC \cite{achanta2012slic} with $s=1000$ and $s=5000$ segments.}
\label{tab:quantiveresults}
\vspace{-0.2cm}\begin{threeparttable}
\begin{tabular}{|l | r r r |}
 \hline
NST Method & $\mathcal{L}_c$ & $\mathcal{L}_s$ &  $\ell_1$ \tnote{1} \\ [0.5ex]
\hline
 $\text{PPN}_\mathit{sst}$ (s=1K$|$5K) & 0.092$|$0.082 & 0.068$|$0.051 &  -\\
Johnson NST \cite{johnson2016perceptual} & \textbf{0.063} & \textbf{0.044} & -\\
\hline
 $\text{PPN}_\mathit{arb}$ (s=1K$|$5K) & 0.084$|$0.088 & 1.434$|$\textbf{1.294} &  -\\
SANet NST \cite{park2019arbitrary} & \textbf{0.081} & 1.434 & -\\
\hline
Optim. (s=1K$|$5K) & \textbf{0.046}$|$0.047 & 0.442$|$0.443 & 0.031|\textbf{0.026}\\
Watercolor \cite{lotzsch2022wise} & 0.056 & 0.704 & 0.036\\
Oilpaint \cite{lotzsch2022wise} & 0.048 & \textbf{0.426} & 0.029\\

\hline
\end{tabular}
       \begin{tablenotes} 
      \footnotesize
      \item[1] The $\ell_1$ distance is computed to the output of STROTTS  \cite{kolkin2019style}
    \end{tablenotes}
      \end{threeparttable}
\end{table}

\paragraph{Mask Editability}
\Cref{tab:mask_editability} compares the effect of the Total variation loss $\mathcal{L}_\mathit{TV}$ on mask noisiness. Masks with less noise are typically more editable.
Compared to our method without $\mathcal{L}_\mathit{TV}$, the noise in the masks is reduced by more than two orders of magnitude when optimized with  $\mathcal{L}_\mathit{TV}$, as measured by noise $\sigma$. The masks produced by the oilpaint pipeline of Loetzsch \etal\cite{lotzsch2022wise} are even more noisy.

\begin{table}
\centering
\caption{Comparison of noise levels in parameter masks. Our pipeline optimized with and without $\mathcal{L}_\mathit{TV}$ is compared against the oilpaint pipeline of Loetzsch~\etal\cite{lotzsch2022wise}. The experimental setup is the same as used for~ \Cref{tab:quantiveresults}. 
}
\label{tab:mask_editability}
\begin{threeparttable}
\begin{tabularx}{\linewidth}{|l @{\extracolsep{\fill}}| r r |}
 \hline
Pipeline & $\sigma$ of estimated noise\tnote{1}  & $\mathcal{L}_\mathit{TV}$ \\ [0.5ex]
\hline
Ours & \textbf{0.0039} & \textbf{0.0041}\\
Ours w/o $\mathcal{L}_\mathit{TV}$ & 0.0858 & 0.0514\\
Oilpaint \cite{lotzsch2022wise} & 0.1015 & 0.0614\\

\hline
\end{tabularx}
       \begin{tablenotes} 
      \footnotesize
      \item[1] We use skimage \href{https://scikit-image.org/docs/stable/api/skimage.restoration.html\#skimage.restoration.estimate\_sigma}{estimate\_sigma} to estimate Gaussian noise $\sigma$
    \end{tablenotes}
      \end{threeparttable}
      \vspace{-0.3cm}
\end{table}

\paragraph{Ablation Study}
\Cref{tab:full_ablation_study} provides an ablation study using the experimental setup also used for \Cref{tab:quantiveresults}, to determine if all filters are necessary in our pipeline by removing filters and computing the $\ell_1$ to its target $I_t$ after optimization. We observe that removing any of the filters has a significant impact on the ability to match arbitrary styles. Please refer the supplementary material for the full ablation study and qualitative examples.

\begin{table}[tb]
\centering
\vspace{-0.1cm}\caption{Filter ablation study. The effect of removing filters on the $\ell_1$ distance to the target is measured. 
Refer to suppl. material for study setup.
}
\label{tab:full_ablation_study}
\vspace{-0.2cm}\begin{tabular}{| r | r r r r |}
\hline
full & w/o Bilat. & w/o Bump & w/o XDoG & w/o Contrast \\
\hline
\textbf{0.0255} & 0.0305 & 0.0372 & 0.0358 & 0.0342\\
\hline
\end{tabular}
\vspace{-0.3cm}
\end{table}

\subsection{Limitations}
\noindent While filter optimization can accurately match a target image, the quality of decomposition into painterly attributes is contingent on several factors. 
First, the parameters controlling a filter must produce distinct effects on the output, \ie be non-overlapping with other filters as the optimization is guided towards generating plausible outputs and editable masks, but does not take the semantic meaning of parameters into account.
Second, the options for explicit painterly control are limited to the pipeline's parameters, adding other differentiable filters requires manual configuration. Further, while our method allows global adaption of textures using geometric abstraction techniques and example-based controls which are visually distributed uniformly, it does not guarantee to maintain statistical textural properties such as spatial homogeneity. 

\section{Conclusions}
\noindent We presented a lightweight, differentiable pipeline for texture editing using text prompts and image examples, and demonstrated its integration with controllable geometric abstraction techniques. Our approach demonstrates the benefits of a decomposed representation of texture for interactive and exploratory editing. This technology and its future enhancements enable the bridging of established image editors and tools with modern image generation techniques. Joint optimization of parameters and shape primitive placement holds the potential for even better decomposition of texture and style and is an avenue for future research.

\section*{Acknowledgments}
\noindent This work was funded by the German Federal Ministry of Education and Research (BMBF) (through grants 01IS15041 -- \enquote{mdViPro}and 01IS19006 -- \enquote{KI-Labor ITSE}), and the German Federal Ministry for Economic Affairs and Climate Action (BMWK) (through grant 16KN086401 -- \enquote{PO-NST}).

\FloatBarrier%

\bibliographystyle{IEEEtran}
\bibliography{IEEEabrv,references}

\ifdefined\isArxiv%
    \newcommand{\mytitle}[1]{%
        \twocolumn[%
        \centering
        {\Huge\sffamily #1\vspace{1em}} 
        ]%
    }

    \mytitle{Controlling Geometric Abstraction \\and Texture for Artistic Images -\\ Supplementary Material}
    \setcounter{section}{0}
    
\vspace{-0.2cm}\section{PPN training} 

\paragraph{Training and implementation}
We trained both $\text{PPN}_\mathit{sst}$  and $\text{PPN}_{\text{arb}}$ for 24 epochs using the MS-COCO dataset \cite{cocodataset} for content images and the WikiArt dataset~\cite{Painterbynumbers2016} for style images for $\text{PPN}_{\text{arb}}$. 
 Both datasets contain approximately 80,000 training images. 
During training, we resize the shorter edge of the input images to 256 pixels and then randomly crop a region of size $256 \times 256$ pixels. The style images are also resized to a resolution of 256 pixels. The parameters of the effect pipeline have specified ranges that vary for each parameter, such as the contour opacity range of $[0,1]$. In order to improve training stability, we normalize the predicted parameter ranges to be between $[-0.5,0.5]$ utilizing the sigmoid function following the final convolutional layer of the \ac{PPN}. The parameter ranges are subsequently scaled back to their original values prior to being input in the effect pipeline. To ensure performance on diverse settings of the segmentation stage $S$, we uniformly vary the global smoothing sigma and kernel sizes within the ranges of [1.0, 3.0] and [1, 7], respectively, as well as the number of segments within the range of [470, 5475].
We used the Adam optimizer \cite{kingma2014adam} with a learning rate of 0.0005 for $\text{PPN}_\mathit{sst}$ and 0.00001 for $\text{PPN}_{\text{arb}}$. A style weight between $[1\times10^{10}, 1\times10^{11}]$ is utilized for $\text{PPN}_\mathit{sst}$, depending on the style image  with a content weight of $1x10^{5}$. 
We employ the same loss weights for $\text{PPN}_{\text{arb}}$ as Park \etal\cite{park2019arbitrary} use for SANet. Using the identity retaining loss of SANet~\cite{park2019arbitrary} is not necessary for training our $\text{PPN}_{\text{arb}}$, as our lightweight filter pipeline is not capable of altering the semantic content structures given by $I_a$ in a major way. 

\section{Runtime Performance}
\vspace{-0.05cm}
In \Cref{tab:runtime_benchmark}, we present a comparison of the runtime performance of our proposed methods for various image sizes. Both PPNs are about equally fast as they only require a forward pass of the network. Furthermore the SLIC segmentation takes around 75\% of the execution time,  and does not have to be continously re-executed in interactive editing scenarios. On the other hand, the optimization method requires several seconds even for small images and is roughly compareable in runtime to other optimization-based NSTs. 

\begin{table}[t]
\centering
\caption{Runtime comparison of our parameter prediction and optimization methods, using a NVIDIA RTX 3090. The optimization is performed for 100 iterations. Note that computation time may not increase linearly with image size due to GPU under-utilization for small images. For the PPNs, we show runtime of the complete method, including target generation using feedforward NST \cite{johnson2016perceptual}, and segmentation using SLIC \cite{achanta2012slic}, and in parentheses the runtime for the PPN and effect only.  Results are presented in seconds. }
\label{tab:runtime_benchmark}
\begin{tabular}{| l | c c c |}
\hline
Image Size & Optimization & $\text{PPN}_\mathit{sst}$ & $\text{PPN}_{\text{arb}}$\\
\hline
256x256 & 11.45 & 0.20 (0.08)  & 0.20 (0.08)\\
512x512 & 17.63 & 0.54 (0.12) & 0.55 (0.13)\\

\hline
\end{tabular}
\end{table}

\begin{table}[t]
    \centering
    \caption{Full ablation study results. In the first two rows, we show results for the full pipeline, all subsequent rows show the results for pipeline configurations with one filter removed. All rows except the first use $\mathcal{L}_\mathit{TV}$ during optimization. All configurations were executed using 1000 and 5000 segments of SLIC segmentation, denoted as (1K$|$5K). The experimental setup is described in \Cref{sec:ablationstudyfull}. }
    \label{tab:sup:full_ablation_study}
    \begin{tabular}{| l | r r r |}
    \hline
    Pipeline Config & $\ell_1$ to ST \cite{kolkin2019style} & $\mathcal{L}_c$ & $\mathcal{L}_s$ \\
    \hline
    w/o $\mathcal{L}_\mathit{TV}$ & 0.028$|$0.023 & 0.047$|$0.048 & 0.409$|$0.421 \\
    w. $\mathcal{L}_\mathit{TV}$ & 0.031$|$0.026 & 0.046$|$0.047 & 0.442$|$0.443 \\
    \hline
    No Bilateral & 0.031$|$0.031 & 0.046$|$0.047 & 0.442$|$0.448 \\
    No Bump Mapping & 0.037$|$0.037 & 0.046$|$0.047 & 0.465$|$0.466 \\
    No XDoG & 0.036$|$0.036 & 0.046$|$0.047 & 0.503$|$0.505 \\
    No Contrast & 0.030$|$0.034 & 0.044$|$0.044 & 0.510$|$0.499 \\
    
    \hline
    \end{tabular}
    \vspace{-0.2cm}
\end{table}


\newcommand{\imagewithspymaskfigone}[3]{
    \begin{tikzpicture}[every node/.style={inner sep=0,outer sep=0}]
        \begin{scope}[ 
			inner sep = 0pt,outer sep=0,spy using outlines={rectangle, red, magnification=2}
                ]
        \node (n0)  { \includegraphics[trim={0cm 128px 0cm 127px},clip,width=\textwidth,height=0.75\textwidth]{#1}};
        \spy [red,size=1.2cm] on (#2,#3) in node[anchor=south east,inner sep=3pt] at (n0.south east);
        \end{scope}
    \end{tikzpicture}
}

\newcommand{\imagewithspymaskfigsecond}[3]{
    \begin{tikzpicture}[every node/.style={inner sep=0,outer sep=0}]
        \begin{scope}[ 
			inner sep = 0pt,outer sep=0,spy using outlines={rectangle, red, magnification=2}
                ]
        \node (n0)  { \includegraphics[trim={0cm 122px 0cm 122px},clip,width=\textwidth,height=0.75\textwidth]{#1}};
        \spy [red,size=1.2cm] on (#2,#3) in node[anchor=south east,inner sep=3pt] at (n0.south east);
        \end{scope}
    \end{tikzpicture}
}

\newcommand{\imagewithspymaskfigthird}[3]{
    \begin{tikzpicture}[every node/.style={inner sep=0,outer sep=0}]
        \begin{scope}[ 
			inner sep = 0pt,outer sep=0,spy using outlines={rectangle, red, magnification=2}
                ]
        \node (n0)  { \includegraphics[trim={0cm 148px 0cm 148px},clip,width=\textwidth,height=0.75\textwidth]{#1}};
        \spy [red,size=1.2cm] on (#2,#3) in node[anchor=south east,inner sep=3pt] at (n0.south east);
        \end{scope}
    \end{tikzpicture}
}

\newcommand{\imagewithspymaskfigfourth}[3]{
    \begin{tikzpicture}[every node/.style={inner sep=0,outer sep=0}]
        \begin{scope}[ 
			inner sep = 0pt,outer sep=0,spy using outlines={rectangle, red, magnification=2}
                ]
        \node (n0)  { \includegraphics[trim={0cm 188px 0cm 68px},clip,width=\textwidth,height=0.75\textwidth]{#1}};
        \spy [red,size=1.2cm] on (#2,#3) in node[anchor=south east,inner sep=3pt] at (n0.south east);
        \end{scope}
    \end{tikzpicture}
}

\begin{figure*}[p]
\centering
\begin{tabular}{m{0.1\textwidth}m{0.2\textwidth}m{0.2\textwidth}m{0.2\textwidth}m{0.2\textwidth} }
Style & \begin{subfigure}{0.08\textwidth}%
		\includegraphics[width=\textwidth,height=\textwidth]{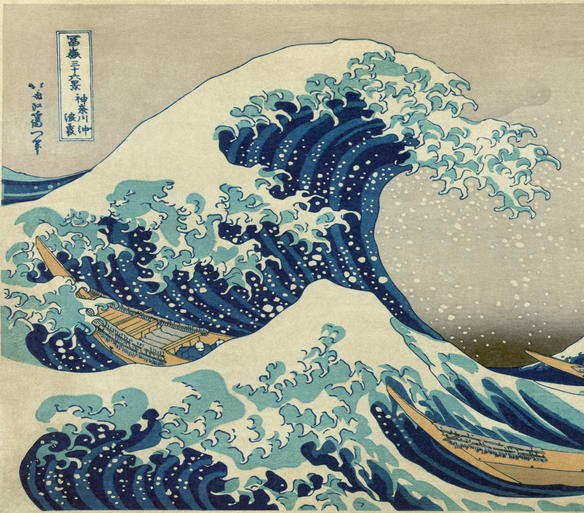}%
	\end{subfigure} & \begin{subfigure}{0.08\textwidth}%
		\includegraphics[width=\textwidth,height=\textwidth]{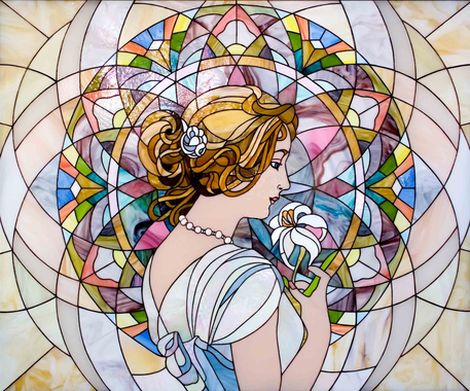}%
	\end{subfigure} & \begin{subfigure}{0.08\textwidth}%
		\includegraphics[width=\textwidth,height=\textwidth]{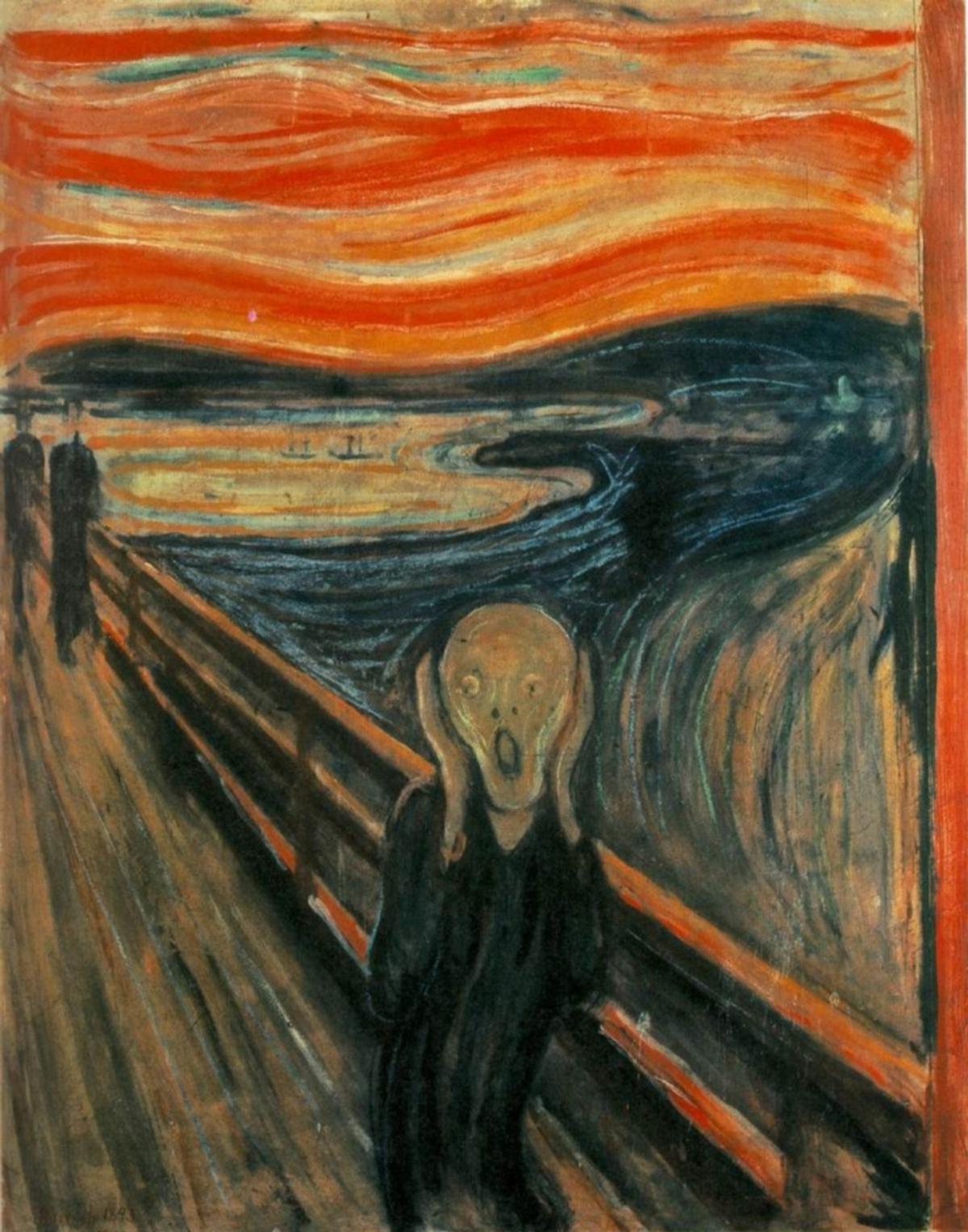}%
	\end{subfigure} & \begin{subfigure}{0.08\textwidth}%
		\includegraphics[width=\textwidth,height=\textwidth]{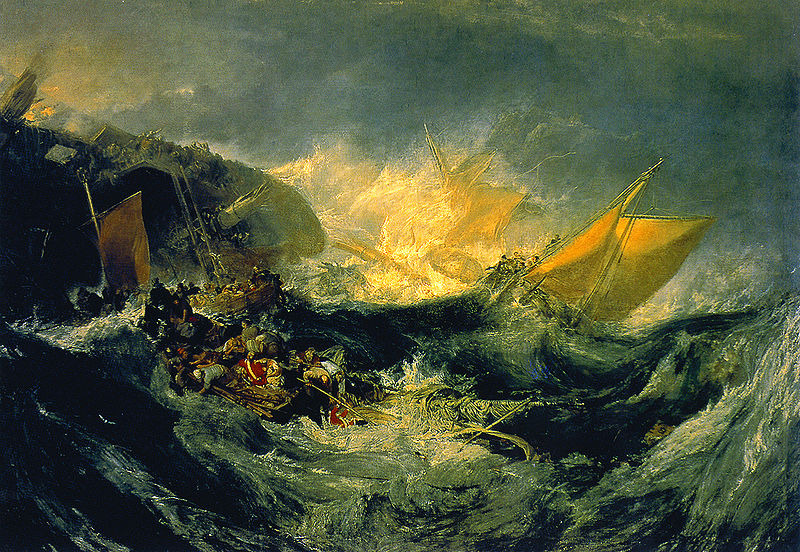}%
	\end{subfigure}\\
Content & \begin{subfigure}{0.2\textwidth}%
		\includegraphics[trim={0cm 128px 0cm 127px},clip,width=\textwidth,height=0.75\textwidth]{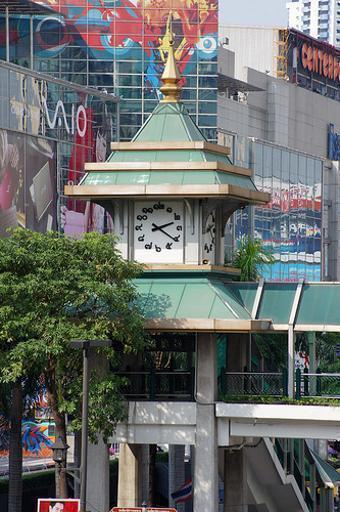}%
	\end{subfigure} & \begin{subfigure}{0.2\textwidth}%
		\includegraphics[trim={0cm 122px 0cm 122px},clip,width=\textwidth,height=0.75\textwidth]{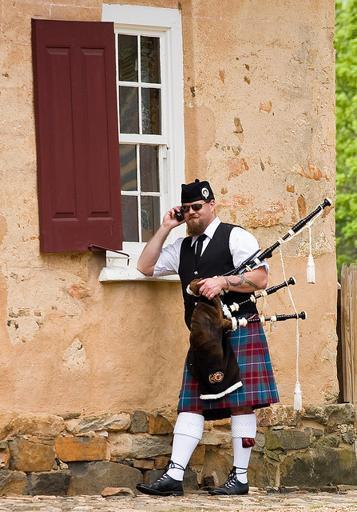}%
	\end{subfigure} & \begin{subfigure}{0.2\textwidth}%
		\includegraphics[trim={0cm 148px 0cm 148px},clip,width=\textwidth,height=0.75\textwidth]{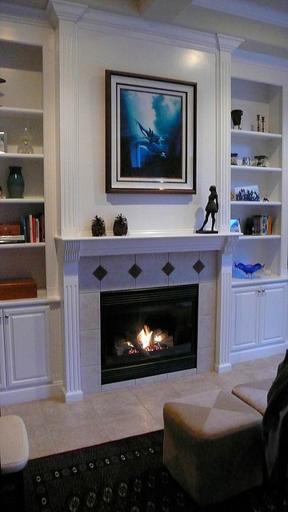}%
	\end{subfigure} & \begin{subfigure}{0.2\textwidth}%
		\includegraphics[trim={0cm 188px 0cm 68px},clip,width=\textwidth,height=0.75\textwidth]{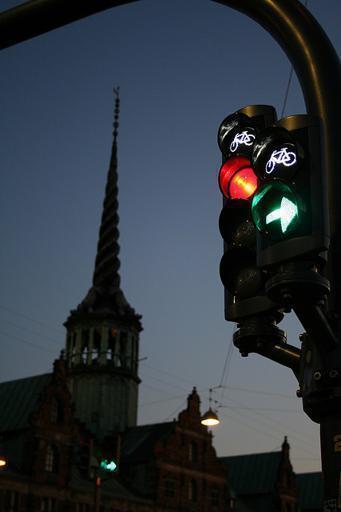}%
	\end{subfigure}\\
STROTTS \cite{kolkin2019style} ground truth & \begin{subfigure}{0.2\textwidth}%
		\imagewithspymaskfigone{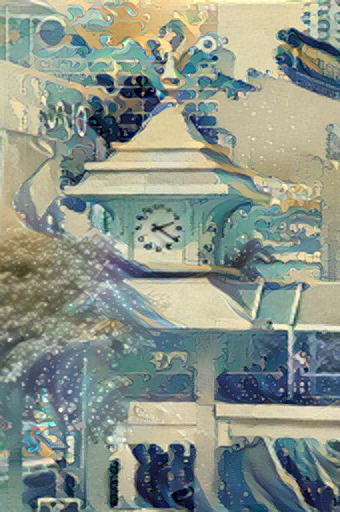}{-0.13cm}{0.29cm}%
	\end{subfigure} & \begin{subfigure}{0.2\textwidth}%
		\imagewithspymaskfigsecond{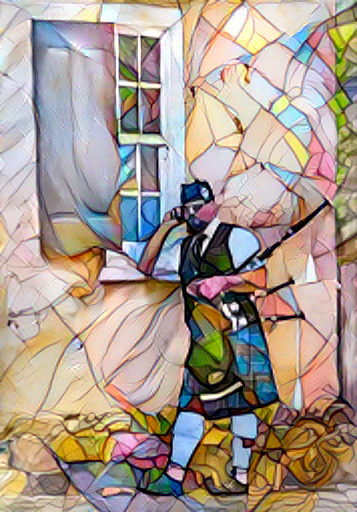}{0.2cm}{0.45cm}
	\end{subfigure} & \begin{subfigure}{0.2\textwidth}%
		\imagewithspymaskfigthird{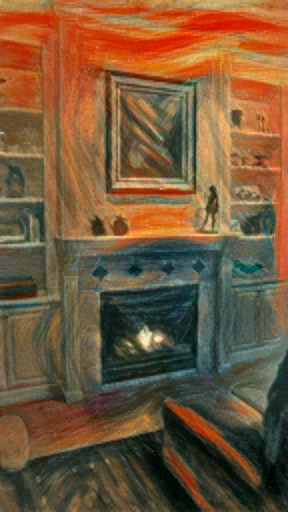}{0.9cm}{0.6cm}
	\end{subfigure} & \begin{subfigure}{0.2\textwidth}%
		\imagewithspymaskfigfourth{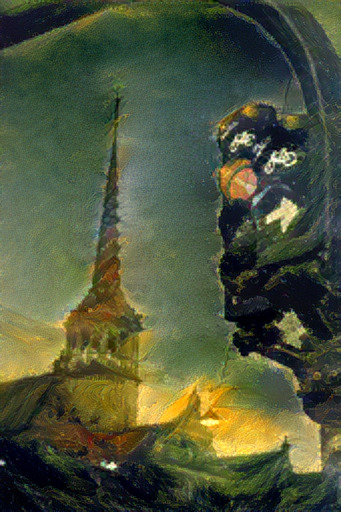}{0.76cm}{0.72cm}%
	\end{subfigure}\\
Full Pipeline & \begin{subfigure}{0.2\textwidth}%
		\imagewithspymaskfigone{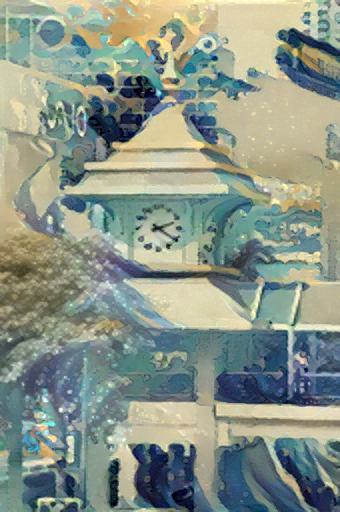}{-0.13cm}{0.29cm}%
	\end{subfigure} & \begin{subfigure}{0.2\textwidth}%
		\imagewithspymaskfigsecond{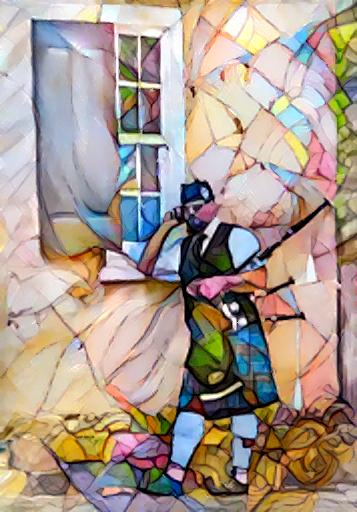}{0.2cm}{0.45cm}%
	\end{subfigure} & \begin{subfigure}{0.2\textwidth}%
		\imagewithspymaskfigthird{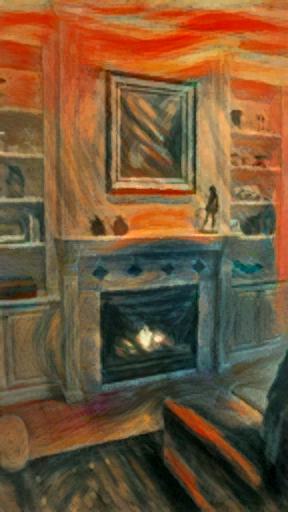}{0.9cm}{0.6cm}%
	\end{subfigure} & \begin{subfigure}{0.2\textwidth}%
		\imagewithspymaskfigfourth{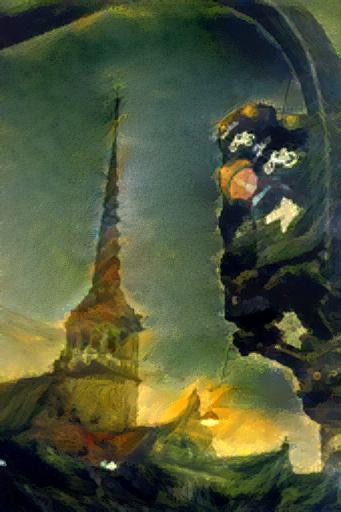}{0.76cm}{0.72cm}%
	\end{subfigure}\\
No Bilateral & \begin{subfigure}{0.2\textwidth}%
		\imagewithspymaskfigone{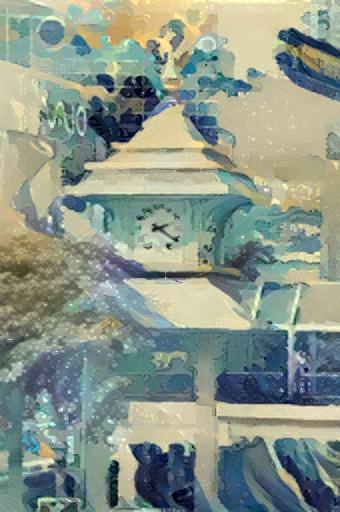}{-0.13cm}{0.29cm}%
	\end{subfigure} & \begin{subfigure}{0.2\textwidth}%
		\imagewithspymaskfigsecond{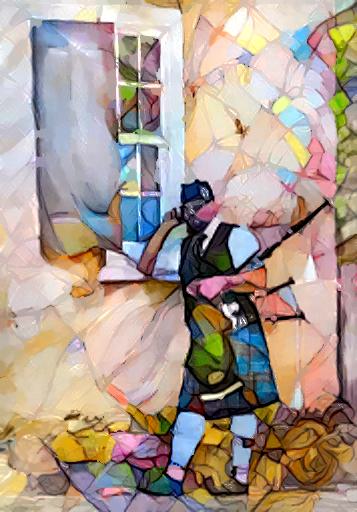}{0.2cm}{0.45cm}%
	\end{subfigure} & \begin{subfigure}{0.2\textwidth}%
		\imagewithspymaskfigthird{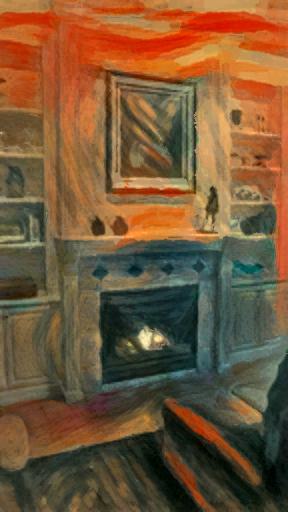}{0.9cm}{0.6cm}%
	\end{subfigure} & \begin{subfigure}{0.2\textwidth}%
		\imagewithspymaskfigfourth{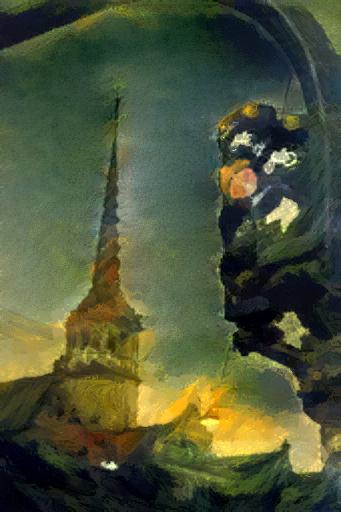}{0.76cm}{0.72cm}%
	\end{subfigure}\\
No XDoG & \begin{subfigure}{0.2\textwidth}%
		\imagewithspymaskfigone{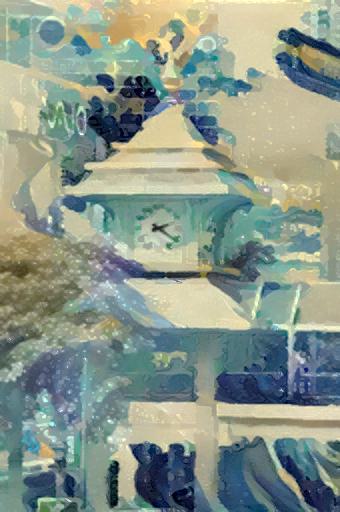}{-0.13cm}{0.29cm}%
	\end{subfigure} & \begin{subfigure}{0.2\textwidth}%
		\imagewithspymaskfigsecond{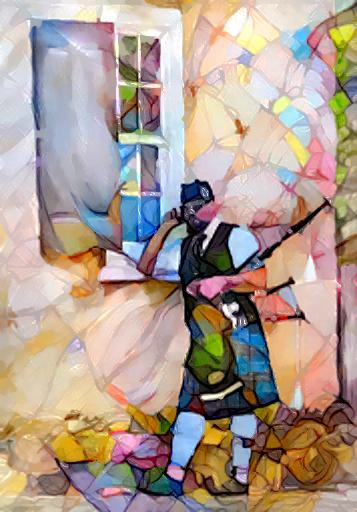}{0.2cm}{0.45cm}%
	\end{subfigure} & \begin{subfigure}{0.2\textwidth}%
		\imagewithspymaskfigthird{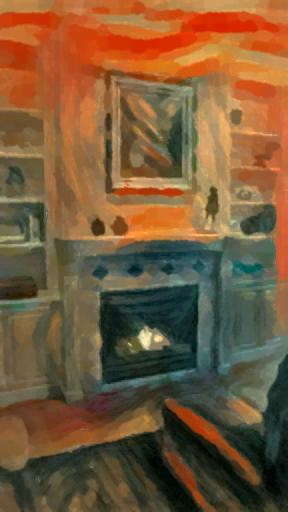}{0.9cm}{0.6cm}%
	\end{subfigure} & \begin{subfigure}{0.2\textwidth}%
		\imagewithspymaskfigfourth{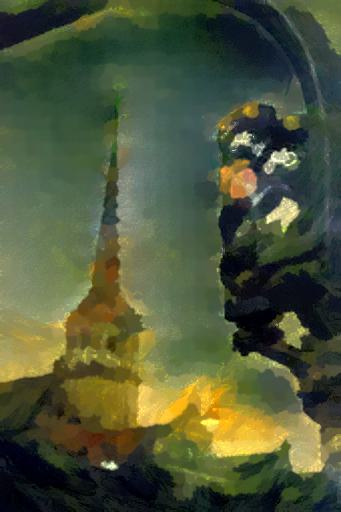}{0.76cm}{0.72cm}%
	\end{subfigure}\\
No Bump Mapping & \begin{subfigure}{0.2\textwidth}%
		\imagewithspymaskfigone{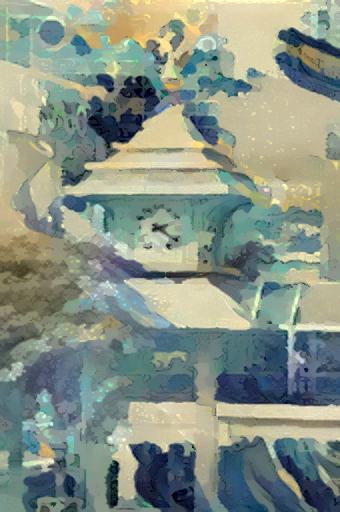}{-0.13cm}{0.29cm}%
	\end{subfigure} & \begin{subfigure}{0.2\textwidth}%
		\imagewithspymaskfigsecond{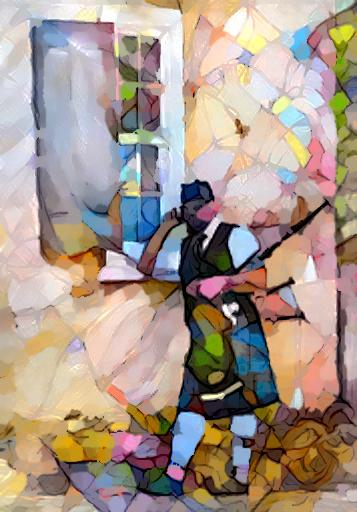}{0.2cm}{0.45cm}%
	\end{subfigure} & \begin{subfigure}{0.2\textwidth}%
		\imagewithspymaskfigthird{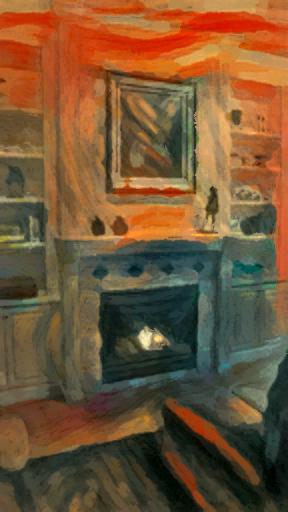}{0.9cm}{0.6cm}%
	\end{subfigure} & \begin{subfigure}{0.2\textwidth}%
		\imagewithspymaskfigfourth{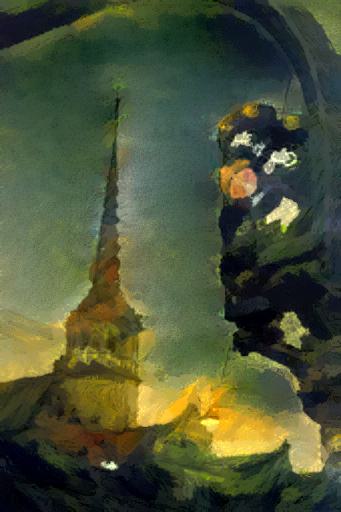}{0.76cm}{0.72cm}%
	\end{subfigure}\\
No Contrast & \begin{subfigure}{0.2\textwidth}%
		\imagewithspymaskfigone{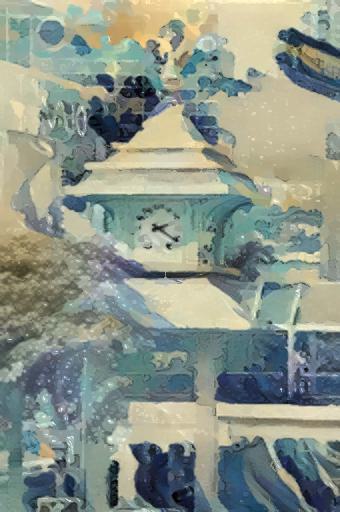}{-0.13cm}{0.29cm}%
	\end{subfigure} & \begin{subfigure}{0.2\textwidth}%
		\imagewithspymaskfigsecond{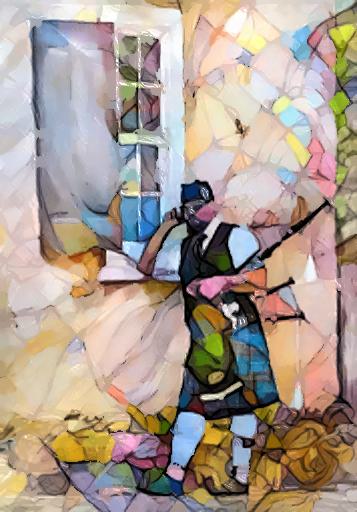}{0.2cm}{0.45cm}%
	\end{subfigure} & \begin{subfigure}{0.2\textwidth}%
		\imagewithspymaskfigthird{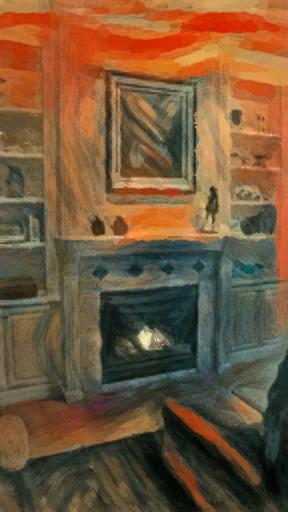}{0.9cm}{0.6cm}%
	\end{subfigure} & \begin{subfigure}{0.2\textwidth}%
		\imagewithspymaskfigfourth{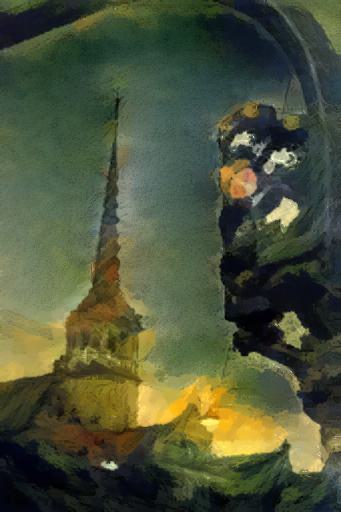}{0.76cm}{0.72cm}%
	\end{subfigure}\\
\end{tabular}
\caption{Example results of the ablation study. Each column contains one stylization target with the results of different pipeline configurations. }
\label{fig:sup:ablation_study_result_image_matrix}
\end{figure*}

\section{Ablation Study - Filter Pipeline}
\label{sec:ablationstudyfull}
In our proposed lightweight pipeline, a set of filters were selected based on their ability to reconstruct texture after geometric abstraction and their capacity to offer meaningful artistic control parameters. The effectiveness of the pipeline is verified by its ability to match a wide range of styles, as shown in Table 2 of the main paper. To determine if all the chosen filters in the pipeline are indeed necessary for representing these styles, an ablation study was conducted, as reported in Table 4 of the main paper. The following sections provide the full details and results of the ablation study.

\paragraph{Study setup.} We selected 20 random content images from MS COCO~\cite{cocodataset} and 10 popular style images with fine and coarse grained texture - we show these style images and a selection of the content images in \Cref{fig:sup:ablation_study_result_image_matrix,fig:sup:nst_comparison_grid}. We used the STROTTS~\cite{kolkin2019style} style transfer algorithm to generate a NST result and then optimized parameters to match this result using the $\ell_1$  loss for $\mathcal{L}_{target}$. We tested different pipeline configurations, where each configuration except our full pipeline has one filter deactivated.
For evaluation, we used the following metrics: VGG\cite{simonyan2014very}-based content and style loss~\cite{gatys2016image}, and the $\ell_1$ difference to the ground truth generated by STROTTS. We used the following hyperparameters throughout the study: content image size (long edge) of 512, style image size of 256, and content and style weights of 0.01 and 5000 respectively, 0.2 for $\lambda_\mathit{TV}$, and  Adam with a learning rate of 0.01 for 500 iterations. We executed the study with 1000 and 5000 segments using SLIC~\cite{achanta2012slic} in the segmentation stage.

\paragraph{Results.} In \Cref{tab:sup:full_ablation_study}, we show full results of our ablation study. Using $\mathcal{L}_\mathit{TV}$ on the full pipeline only slightly increases the  $\ell_1$  loss to the STROTTS~\cite{kolkin2019style} ground truth, whereas removing one filter increases the loss value considerably. 
Example images from the ablation study are shown in Figure~\ref{fig:sup:ablation_study_result_image_matrix}.  Our proposed full pipeline, outlined in Section 3.3 of the main paper, and its optimization target, as generated by STROTTS \cite{kolkin2019style}, demonstrate a close match. The subsequent rows highlight that the removal of a single filter impairs the pipeline's capability to reconstruct details. All ablated configurations exhibit difficulties in reconstructing the clock face in the first column. The second column illustrates that the absence of bump mapping leads to a failure in reconstructing areas of high luminance, such as on the forehead and hat. The removal of any of the other filters results in difficulties in painting over segment boundaries, resulting in artifacts as seen in the forehead area. The third column demonstrates that small-scale contours are not properly retained. The last column illustrates the pipeline's inability to locally enhance contrast when the contrast filter is omitted.

\paragraph{Other pipelines.} We briefly test other pipelines for arbitrary style representation. The oilpaint and watercolor pipelines of Semmo~\cite{semmo2016image} and Bousseau~\cite{bousseau2006interactive}, as implemented in the differentiable framework of Loetzsch \etal\cite{lotzsch2022wise}  are not well-suited for editing. Next to the noisiness of their parameter masks (Table 3., main paper), they often contain redundant parameters, which do not contain any information after optimization.  For example,~\Cref{fig:sup:watercolor_disadvantage} depicts that parameters such as edge-darkening, wetness-scale or color-amount do not contain locally adjusted values - even though one might for example expect colour amount to be increased in the saturated areas of the stylized image. 

\begin{figure}[]
    \centering
    \begin{subfigure}{.325\linewidth}
      \centering
      \includegraphics[width=\linewidth]{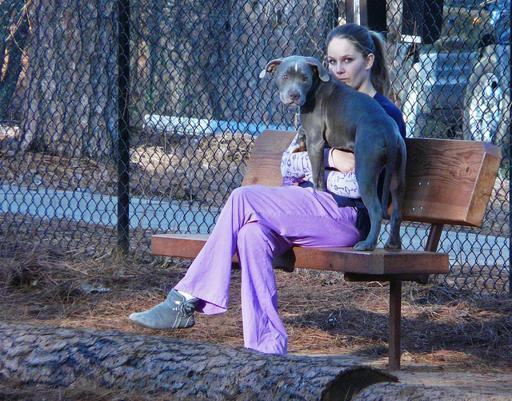}
      \caption{Content}
    \end{subfigure}\hfill
    \begin{subfigure}{.325\linewidth}
        \centering
              \includegraphics[width=\linewidth]{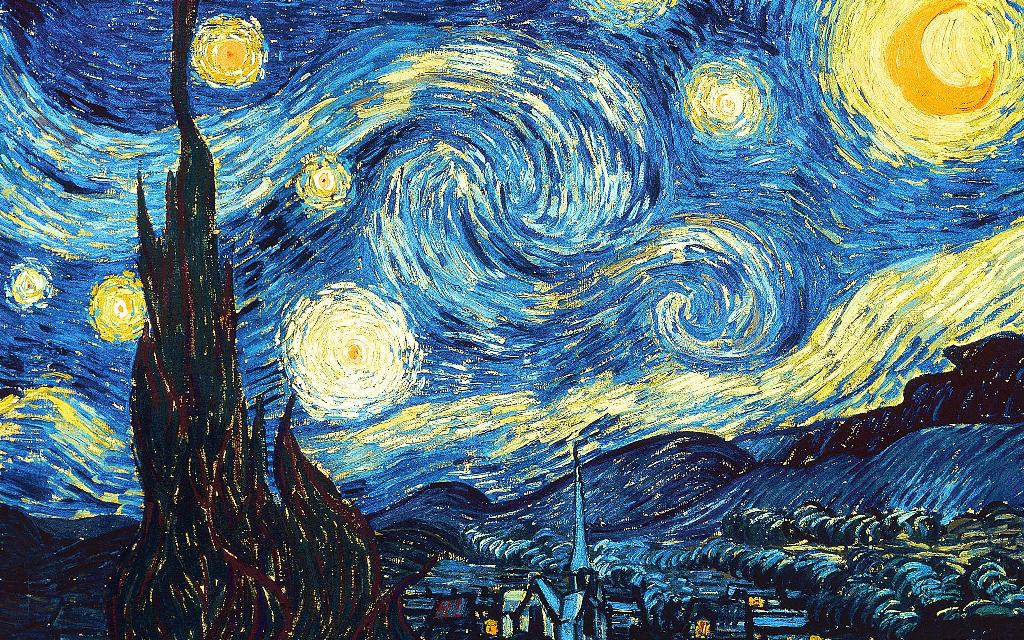}
      \caption{Style}
    \end{subfigure}\hfill
    \begin{subfigure}{.325\linewidth}
      \centering
      \includegraphics[width=\linewidth]{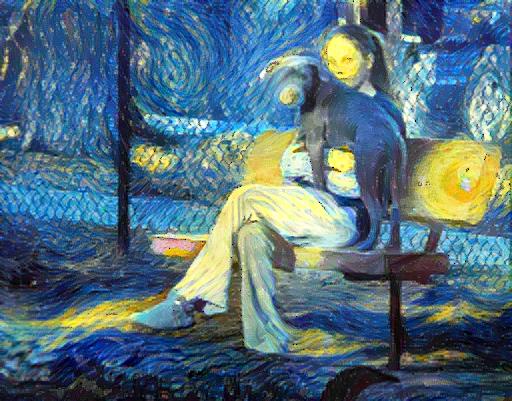}
      \caption{Stylized}
      \label{fig:sup:watercolor_disadvantage:stylized}
    \end{subfigure}\\
    \begin{subfigure}{.325\linewidth}
      \centering
      \includegraphics[width=\linewidth]{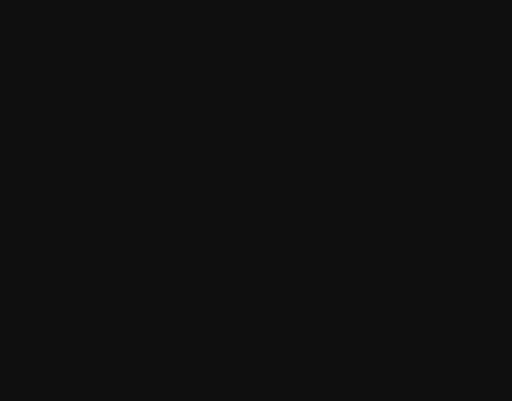}
      \caption{Color Amount}
    \end{subfigure}\hfill
    \begin{subfigure}{.325\linewidth}
        \centering
      \includegraphics[width=\linewidth]{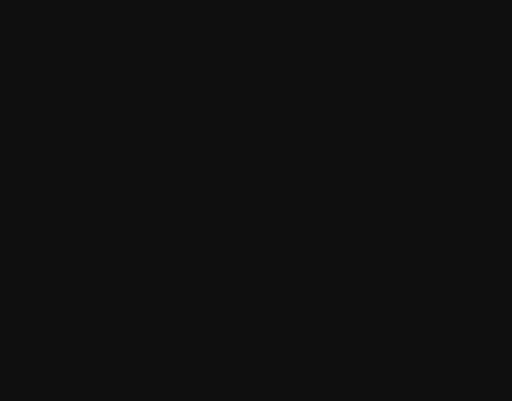}
      \caption{Edge Darkening}
    \end{subfigure}\hfill
    \begin{subfigure}{.325\linewidth}
      \centering
      \includegraphics[width=\linewidth]{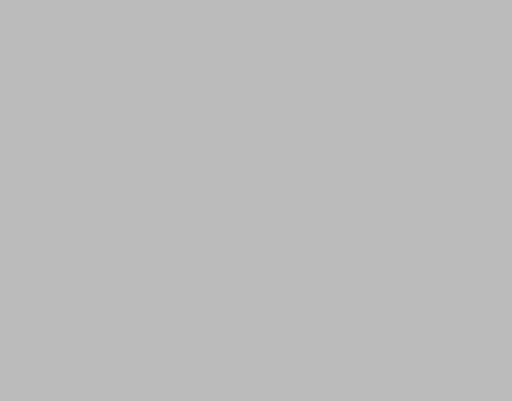}
      \caption{Wetness Scale}
    \end{subfigure}\\
    \caption{Selected parameter masks of the watercolor \cite{bousseau2006interactive} differentiable filter pipeline \cite{lotzsch2022wise}. We show the masks after optimization to match a NST image (\subref{fig:sup:watercolor_disadvantage:stylized}). The parameters color Amount, Edge Darkening, and Wetness Scale are not used or do not display variation in values.}
    \label{fig:sup:watercolor_disadvantage}
\end{figure}

\section{Style-matching Capabilities}
In Table 2 of the main paper, we quantitatively evaluated the style-matching capabilities of our proposed parameter prediction and optimization methods against their respective pixel-predicting and optimizing NST baselines. The results indicate that our methods are capable of achieving good style-matching performance. We provide additional visual evidence in \Cref{fig:sup:nst_comparison_grid} by showing the output of our parameter optimization and single and arbitrary style parameter prediction networks. The results are visually very similar to their baselines, e.g., refer to \Cref{fig:sup:pipeline_comparison} for a comparison between the STROTTS \cite{kolkin2019style} baseline and the parameter optimization. 
However, it is noted that the arbitrary style parameter prediction network ($\text{PPN}_\mathit{arb}$) sometimes overuses the bump-mapping parameter, resulting in an excessive prediction of specular oilpaint texture where the input NST would not have placed such elements. Moreover, we compare our filter pipeline with the style-specific differentiable pipelines implemented by L\"otzsch \etal\cite{lotzsch2022wise} in \Cref{fig:sup:pipeline_comparison}. We compare example results obtained from L\"otzsch \etal\cite{lotzsch2022wise} for the oil-paint \cite{semmo2016image} and watercolor \cite{bousseau2006interactive} filter pipelines with our method. Both style-specific pipelines tend to have problems in matching the color in small regions, even though a dedicated hue adaption step is employed by these pipelines.

\section{Editing Application}
We implemented an prototypical application that makes use of our decomposed shape and texture representation to edit results of a NST. We provide editing metaphors for the structural abstracted output  $I_a$ of the segmentation stage $S$ (using SLIC~\cite{achanta2012slic}). This enables coarse granular edits of color and structure. The application further uses a $\text{PPN}_\mathit{sst}$ to predict parameter masks $P_M$ for the filter pipeline. After performing structural edits on $I_a$, the $P_M$ can be re-predicted by $\text{PPN}_\mathit{sst}$ to adapt fine details to these edits. 

\newcommand{\editingexamplespy}[3]{
    \begin{tikzpicture}[every node/.style={inner sep=0,outer sep=0}]
        \begin{scope}[
			inner sep = 0pt,outer sep=0,spy using outlines={rectangle, red, magnification=2}
                ]
        \node (n0)  { \includegraphics[trim=0cm 1cm 0cm 1.5cm, clip,width=\textwidth]{#1}};
        \spy [red,size=1.2cm] on (#2,#3) in node[anchor=south east,inner sep=0pt] at (n0.south east);
        \end{scope}
    \end{tikzpicture}
    \vspace{-0.5cm}
}

\begin{figure*}[p]
\captionsetup[subfigure]{labelformat=empty}
\addtolength{\tabcolsep}{-0.4em} 
\centering
\begin{tabular}{m{0.045\textwidth}m{0.85\textwidth}}
Un-edited & \begin{subfigure}[t]{0.26\textwidth}%
		\includegraphics[trim=0cm 1cm 0cm 1.5cm, clip,width=\textwidth]{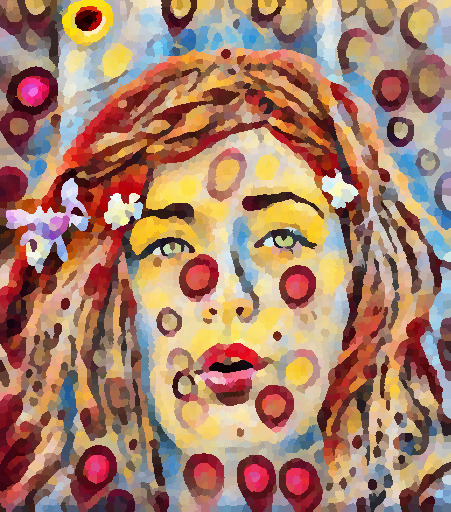}%
        \caption{Segmentation Start}
	\end{subfigure} \begin{subfigure}[t]{0.26\textwidth}%
		\includegraphics[trim=0cm 1cm 0cm 1.5cm, clip,width=\textwidth]{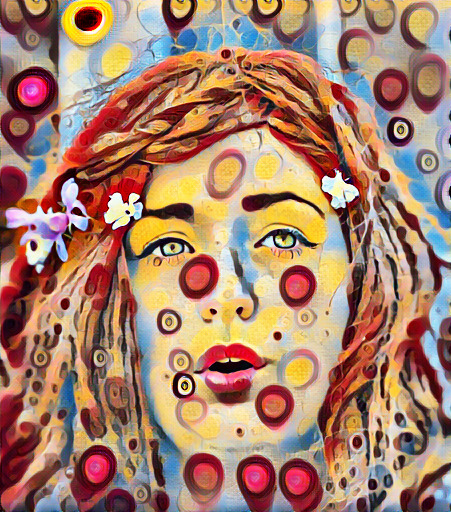}%
    \caption{Johnson NST Result}
	\end{subfigure} \begin{subfigure}[t]{0.26\textwidth}%
		\editingexamplespy{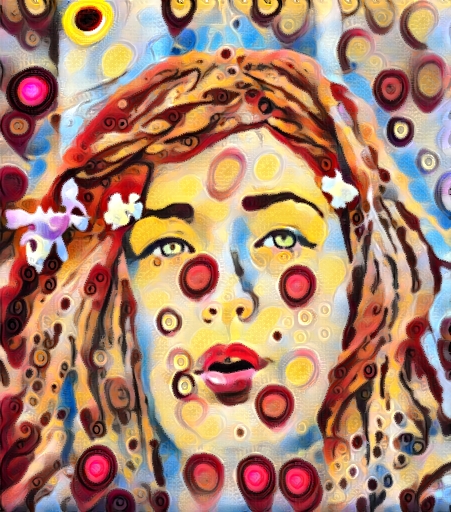}{0.05cm}{-1.1cm}
    \caption{Unedited Output}
	\end{subfigure}\\
 $1^{st}$ Edit & \begin{subfigure}[B]{0.26\textwidth}%
		\editingexamplespy{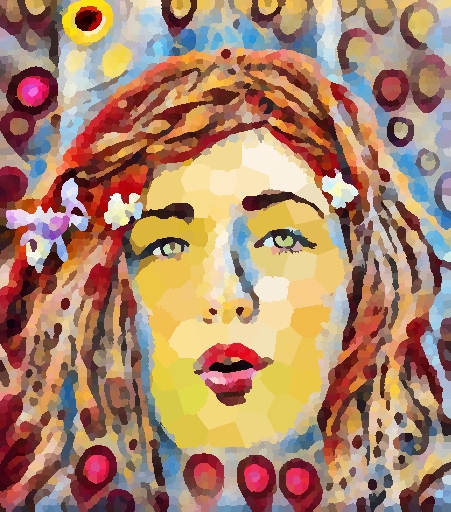}{0.05cm}{-1.1cm}
    \caption{Segmentation}
	\end{subfigure} \begin{subfigure}[B]{0.26\textwidth}%
		\editingexamplespy{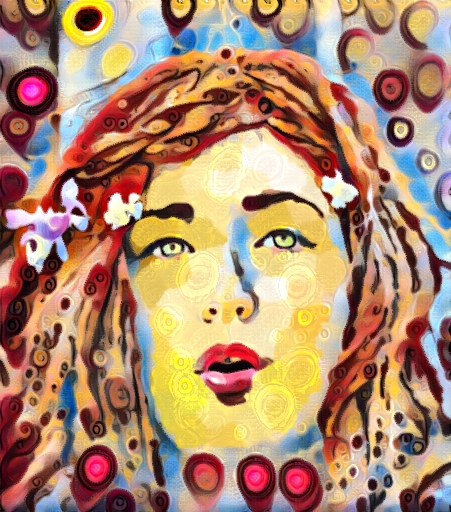}{0.8cm}{-0.6cm}
    \caption{Output}
	\end{subfigure} \begin{subfigure}[B]{0.26\textwidth}%
		\editingexamplespy{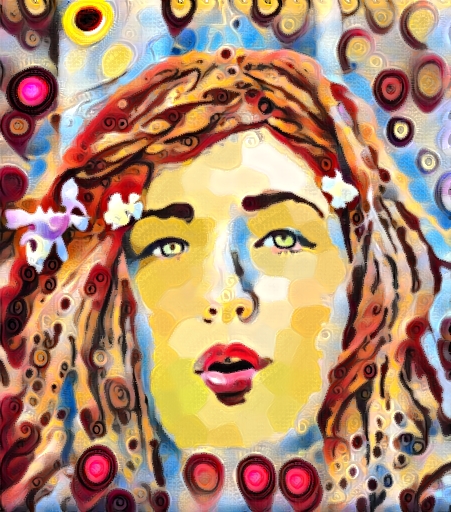}{0.8cm}{-0.6cm}
    \caption{Repredicted Output}
	\end{subfigure}\\
 
$2^{nd}$ Edit & \begin{subfigure}[B]{0.26\textwidth}%
		\editingexamplespy{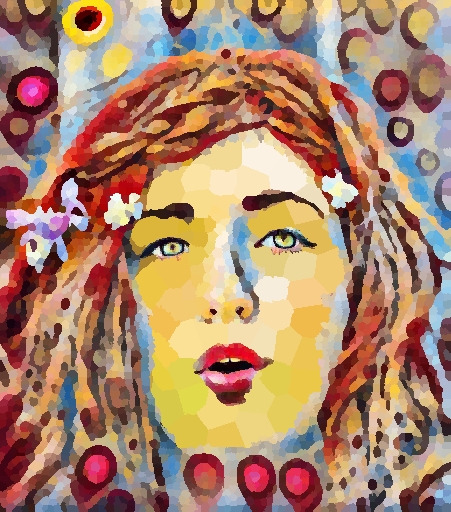}{0.05cm}{-1.1cm}
    \caption{Segmentation}
	\end{subfigure} \begin{subfigure}[B]{0.26\textwidth}%
		\editingexamplespy{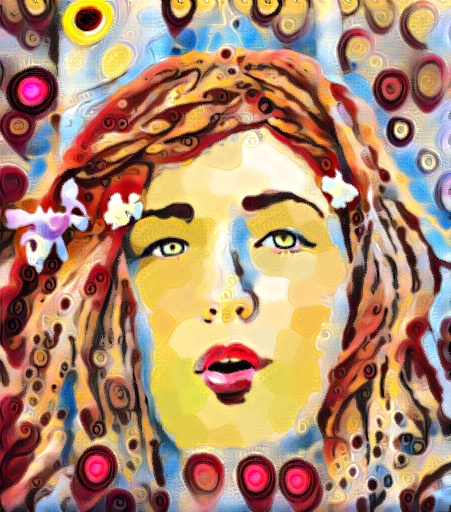}{0.05cm}{-1.1cm}
    \caption{Output}
	\end{subfigure} \begin{subfigure}[B]{0.26\textwidth}%
		\editingexamplespy{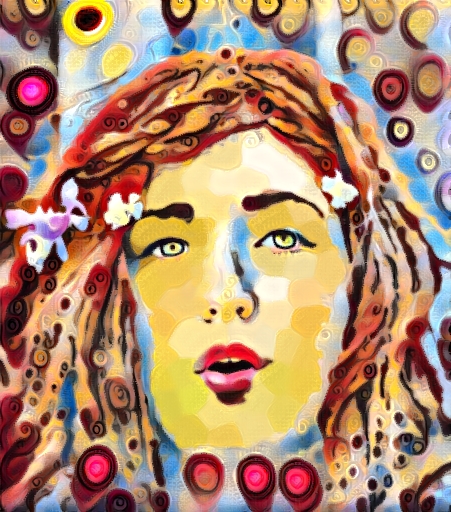}{0.05cm}{-1.1cm}
    \caption{Repredicted Output}
	\end{subfigure}\\
 
 & \begin{subfigure}[B]{0.26\textwidth}%
		\includegraphics[width=\textwidth]{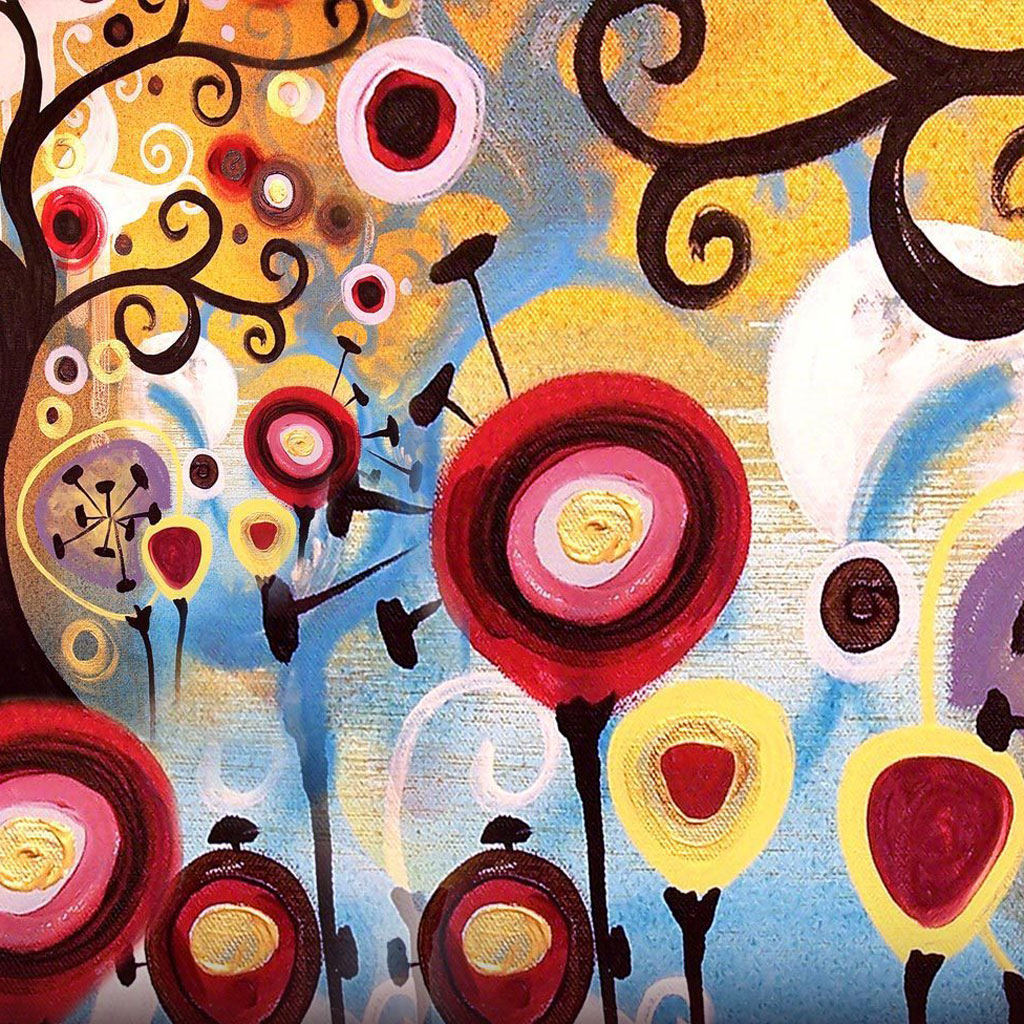}%
        \caption{Style}
        \vspace{-0.2cm}
	\end{subfigure} \begin{subfigure}[B]{0.26\textwidth}%
		\includegraphics[trim=0cm 2cm 0cm 3cm, clip,width=\textwidth]{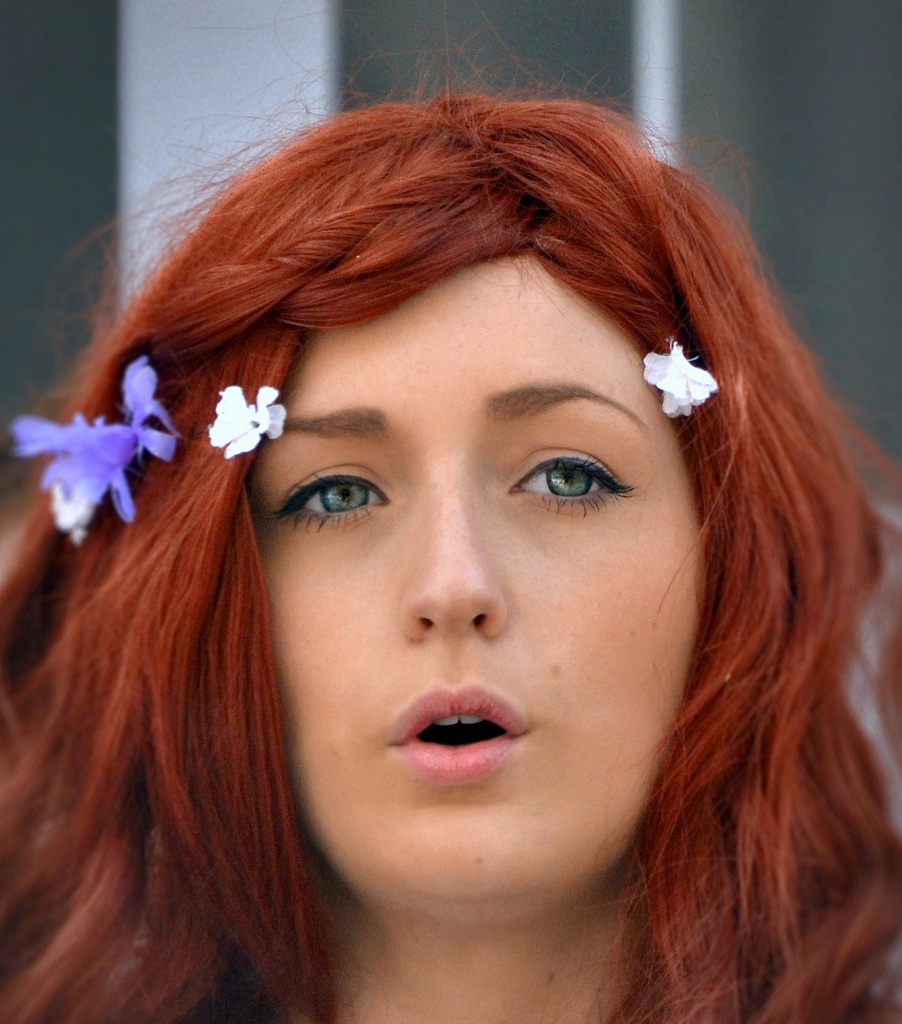}%
  \caption{Content}
  \vspace{-0.2cm}
	\end{subfigure} \begin{subfigure}[B]{0.26\textwidth}%
		\includegraphics[trim=0cm 1cm 0cm 1.5cm, clip,width=\textwidth]{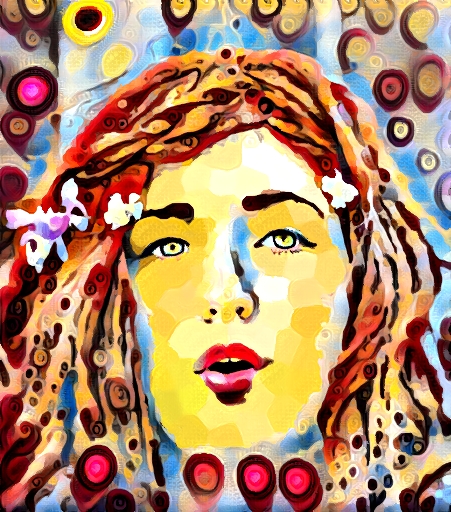}%
        \caption{Output after Mask interpolation}
        \vspace{-0.2cm}
	\end{subfigure}\\
\end{tabular}
\caption{Illustration of the example editing process described in \Cref{par:editing_example}. The top row displays the initial state, while the middle two rows present the intermediate results after the first and second editing step, respectively. The bottom row exhibits the final output after the parameter mask interpolation, alongside the content and style image used in the editing process. Our method allows changing structures of the image and the parameter masks adapt automatically after reprediction.}
\label{fig:sup:example_edit}
\end{figure*}

\subsection{Editing Tools}
The user interface allows selection of one or multiple individual segments to perform various forms of editing, which are detailed in the following:

\begin{description}
\item [Content Image Interpolation] Interpolates $I_a$ with the content image $I_c$, enabling structure reconstruction from  $I_c$ by adjusting the colors of the affected segments.
\item [Color Interpolation] Interpolates a region with a user-defined color, allowing for adjustment of a region's color.
\item [Color Palette Matching] Matches the color palette of a destination region to the color palette of a source region, facilitating the matching of color patterns between regions while maintaining their structures.
\item [Copy Region] Copies a region to another image region, making it possible to move a structure or style element.
\item [Change Level of Detail] Alters the number of segments used to display a region, which can increase the level of detail for important regions like facial features or decrease the level of detail for backgrounds.
\label{list:segment_manipulation_description}
\end{description}


\newcommand{\adamlisterCompareFig}{
    \begin{figure*}
        \centering
        \begin{subfigure}{0.195\linewidth}
          \includegraphics[width=\linewidth]{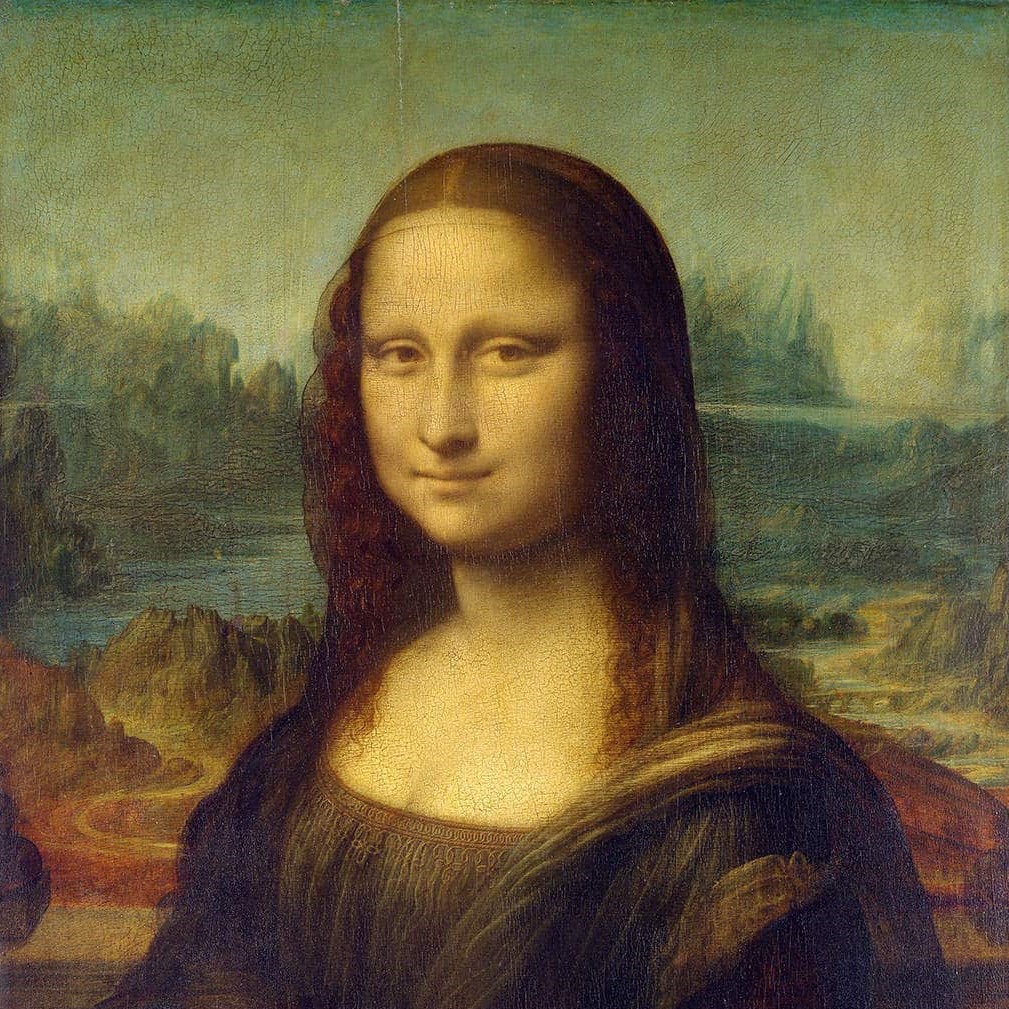}
        \end{subfigure}\hfill
        \begin{subfigure}{0.195\linewidth}
          \includegraphics[width=\linewidth]{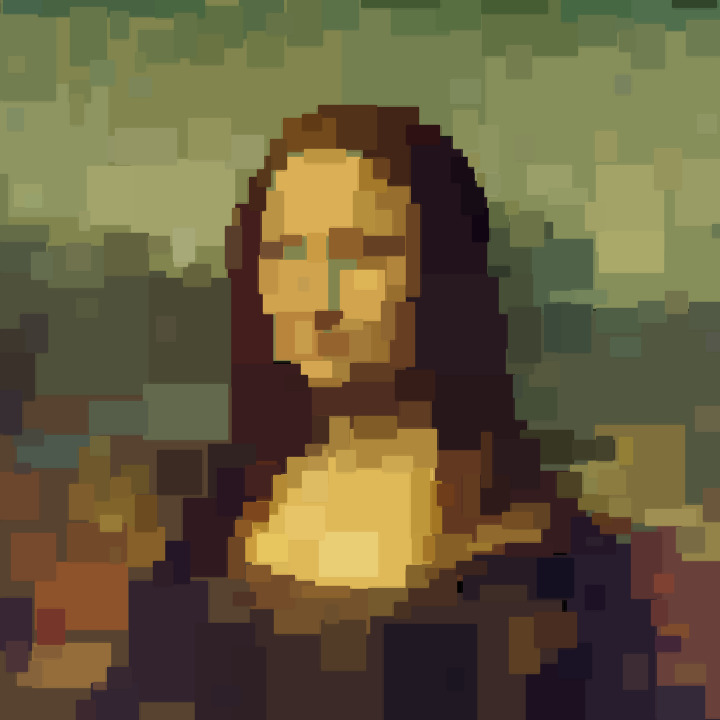}
        \end{subfigure}\hfill
        \begin{subfigure}{0.195\linewidth}
          \includegraphics[width=\linewidth]{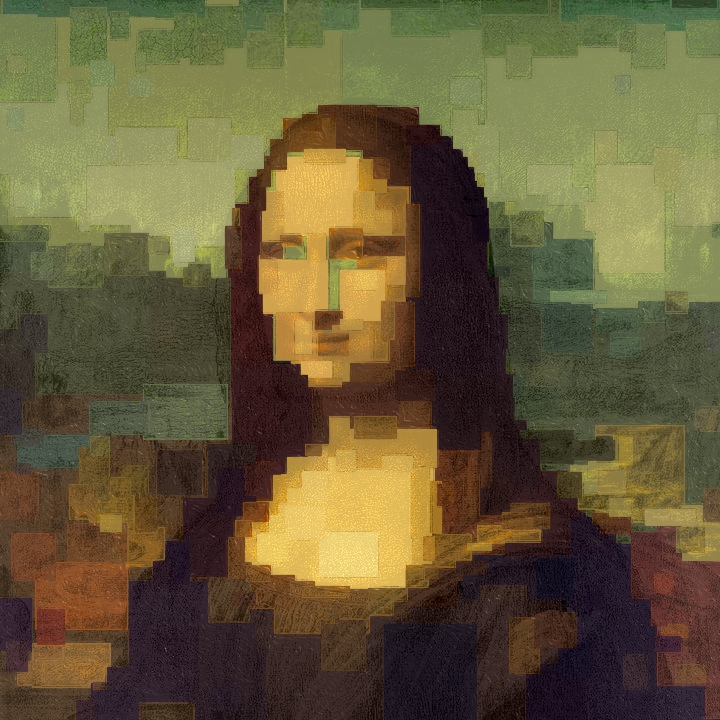}
        \end{subfigure}\hfill
        \begin{subfigure}{0.195\linewidth}
          \includegraphics[width=\linewidth]{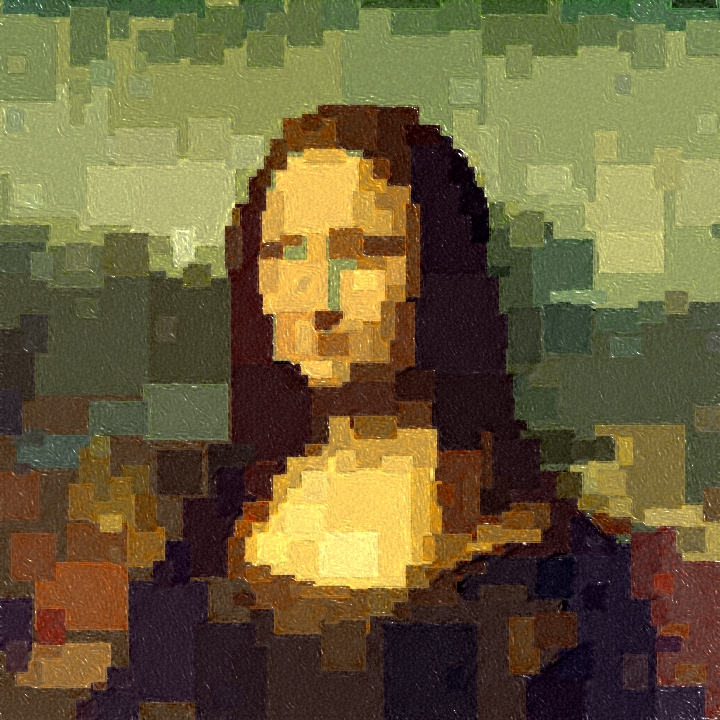}
        \end{subfigure}\hfill
        \begin{subfigure}{0.195\linewidth}
          \includegraphics[width=\linewidth]{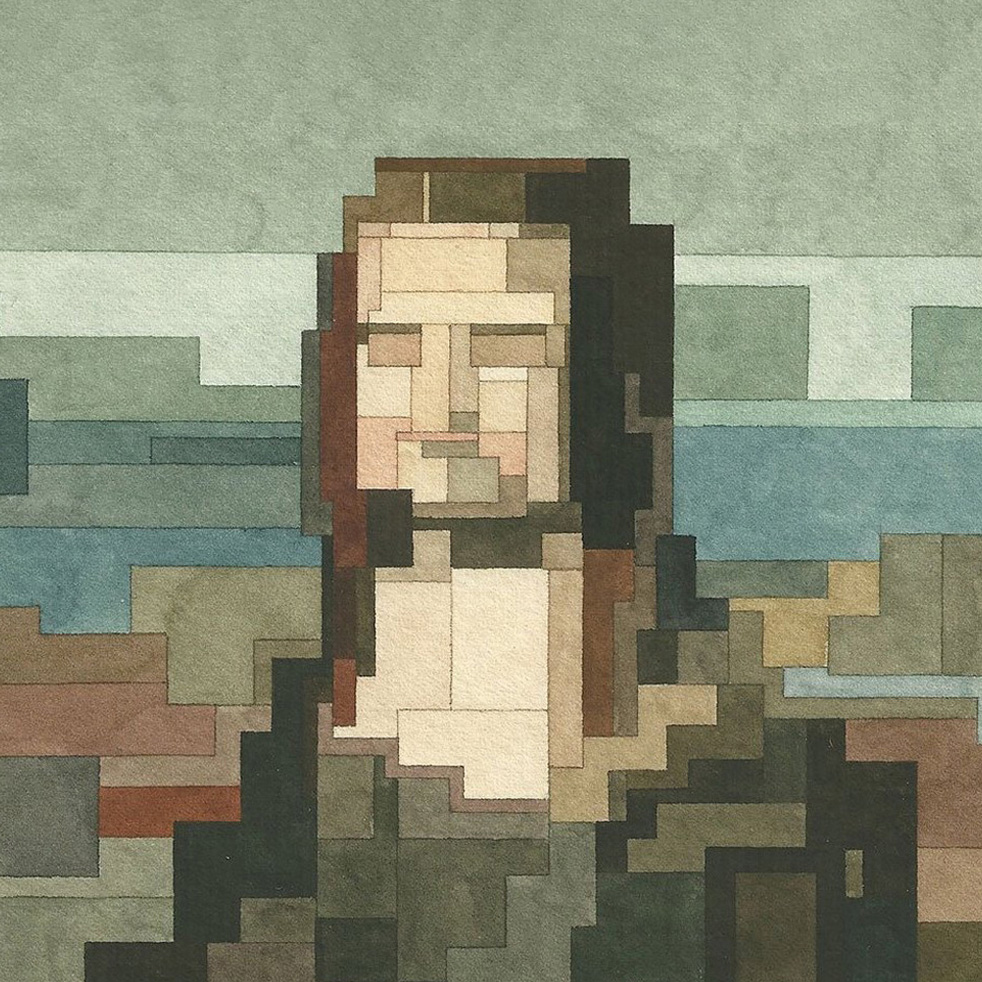}
        \end{subfigure}\\
        \begin{subfigure}{0.195\linewidth}
          \includegraphics[width=\linewidth]{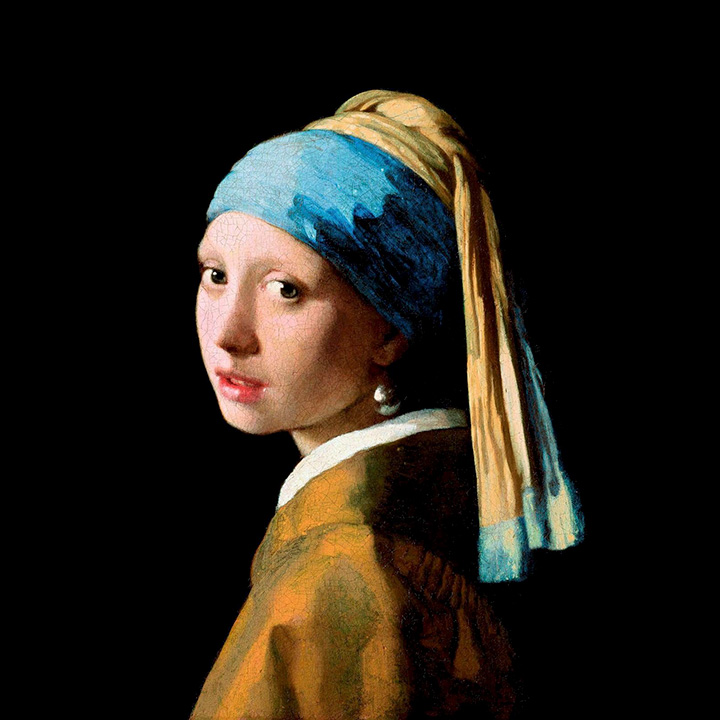}
          \caption{Input}
          \label{fig:sup:8bitart:input}
        \end{subfigure}\hfill
        \begin{subfigure}{0.195\linewidth}
          \includegraphics[width=\linewidth]{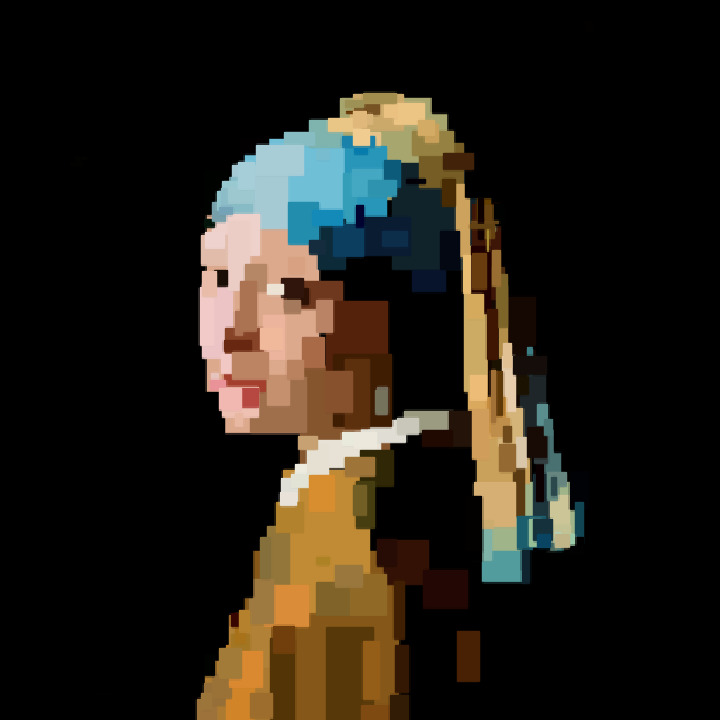}
          \caption{Segmented (NPtr \cite{zou2021stylized})} 
          \label{fig:sup:8bitart:segmented}
        \end{subfigure}\hfill
        \begin{subfigure}{0.195\linewidth}
          \includegraphics[width=\linewidth]{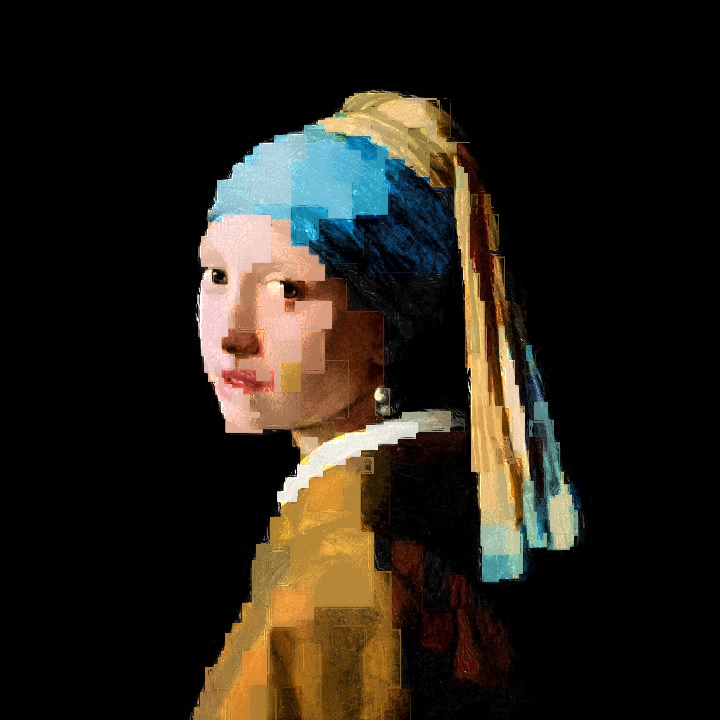}
          \caption{$P_M$ Optimized ($\ell_1$)}
          \label{fig:sup:8bitart:optimized}
        \end{subfigure}\hfill
        \begin{subfigure}{0.195\linewidth}
          \includegraphics[width=\linewidth]{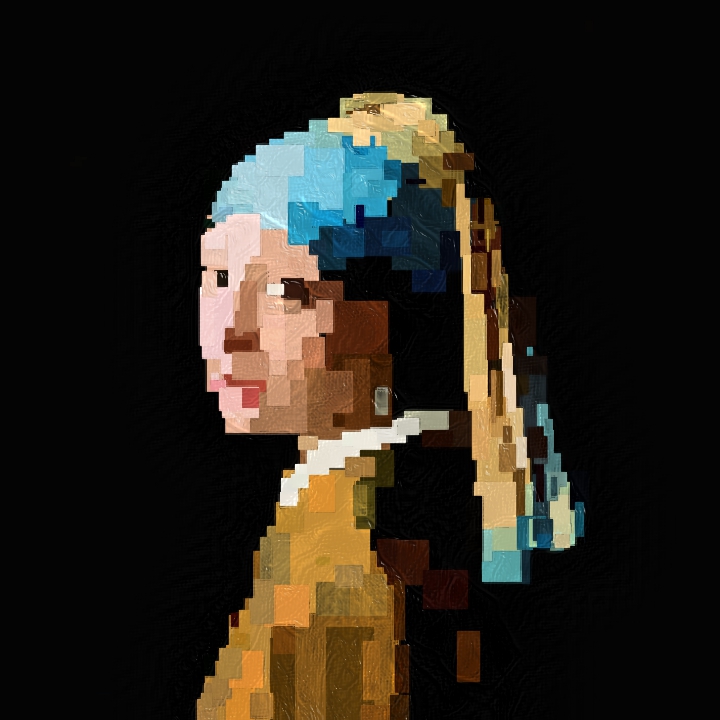}
          \caption{\footnotesize $P_M$ re-predict (AST PPN) }
          \label{fig:sup:8bitart:repredict}
        \end{subfigure}\hfill
        \begin{subfigure}{0.195\linewidth}
          \includegraphics[width=\linewidth]{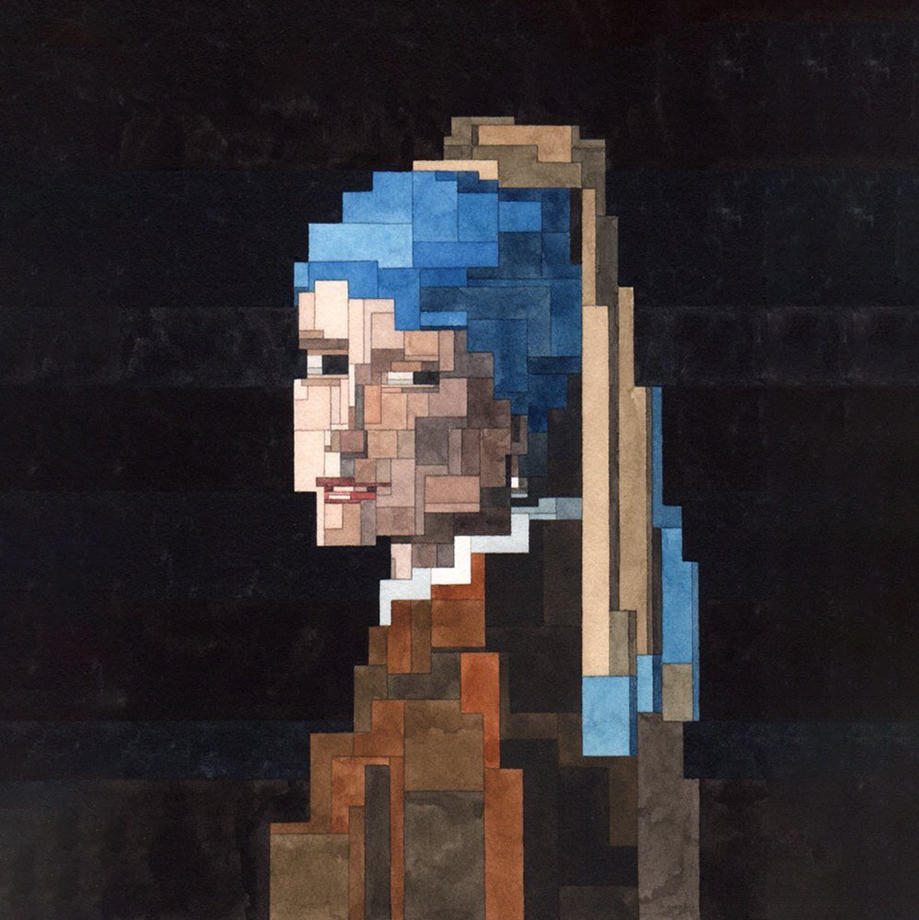}
          \caption{\footnotesize Hand-painted by A. Lister} 
          \label{fig:sup:8bitart:adamlister}
        \end{subfigure}%
        \caption{Creating 8-bit art similar to the style of Adam Lister (\subref{fig:sup:8bitart:adamlister}). The input image (\subref{fig:sup:8bitart:input}) is segmented by the neural painter~\cite{zou2021stylized} (NPtr) method using 300 strokes (\subref{fig:sup:8bitart:segmented}). After optimizing the parameter masks $P_M$ with $\mathcal{L}$ to re-add the details (\subref{fig:sup:8bitart:optimized}), they are further edited, \eg by re-reprediction using the arbitrary style \ac{PPN} (\subref{fig:sup:8bitart:repredict}). }
        \label{fig:sup:8bitart}
    \end{figure*}
}

\ifdefined\isArxiv
    \adamlisterCompareFig
\fi

\subsection{Editing Example}
\label{par:editing_example}
The application's capabilities are demonstrated through an example usage, highlighting both structural and detail edits. To showcase an editing workflow, we depict the editing steps in \Cref{fig:sup:example_edit}. We use the candy style image to train both a Johnson NST and a $\text{PPN}_\mathit{sst}$ and then use our application to stylize a content image (sourced from NPRP~\cite{rosin2020nprportrait}). For each editing step (row two and three of \Cref{fig:sup:example_edit}), we provide the segmentation, the pipeline output, and the pipeline output after repredicting the visual parameters after the described edits have been made.

The first step of the editing process, depicted in the second row of \Cref{fig:sup:example_edit}, entails eliminating all circular style elements. This is achieved by selecting the entire face, excluding the mouth, nose, and eyes, and utilizing the \emph{Content Image Interpolation} tool to remove circular structures. Since the content image lacks these circular structures, they are replaced by the original content. Next, the \emph{Change Level of Detail} tool is applied to reduce the facial region's level of detail, followed by employing the \emph{Color Palette Matching} tool selecting a yellow face region to reinstate the face color, ensuring consistency with the style image. Throughout this procedure, parameter masks encoding fine details remain unaltered. After re-predicting the parameter masks using the $\text{PPN}_\mathit{sst}$, they adapt to the structural modifications, leading to the elimination of the initial circular structures in the parameter masks.

In the second editing step displayed row three of \Cref{fig:sup:example_edit}, the \emph{Change Level of Detail} tool was used to increase the level of detail of the mouth, eyes, and nose.  Upon re-prediction of parameter masks, only minor differences emerge, as the structure remains unchanged.

In the final step, the saliency mask is utilized to adaptively enhance the contrast parameter mask. This leads to an increased contrast within the saliency region, which, in this instance, corresponds to the woman's face.

\section{Further results}

\paragraph{Geometric Abstractions} 
The combination of the geometric abstraction stage with the subsequent filter-based stage allows for reintroduction of details, which can mimic the style of "8-bit" quantized artwork such as the watercolor paintings of Adam Lister, see \Cref{fig:sup:8bitart}.
Direct $\ell_1$ optimization may, however, cause more details to be reintroduced than desired, even though such "glitch-art" effects (\Cref{fig:sup:8bitart:optimized}) have artistic purpose of their own \cite{menkman2011glitch}. Instead, the parameters can, for example, be repredicted using $\text{PPN}_{arb}$ with $I_s=I_t$ to only retain the inputs' textures (\Cref{fig:sup:8bitart:repredict}).

Further, in \Cref{fig:sup:goghabstract} we show the full teaser images for geometric abstraction editing. In \Cref{fig:sup:realsegmented} and \Cref{fig:sup:painttransformer_triangles}, we vary the geometric abstraction while keeping the texture fixed. 

\paragraph{Comparisons with related works}
In \Cref{fig:sup:rw_compare_clip,fig:sup:rw_compare_strotss} we provide further qualitative comparisons to text-based and image-based stylization methods.

\paragraph{Parameter Masks} In \Cref{fig:sup:param_mask_showcase} we showcase a full set of parameter masks after optimization.

\paragraph{CLIPStyler Experiments}
 In \Cref{fig:sup:clipstyler_rainprincess_row,fig:sup:clipsterstyledemos4}
 we experiment with different CLIPStyler-loss text-prompts.

\paragraph{Editing and Blending} In \Cref{fig:sup:iterativeediting} we show the intermediate results of our interactive editing workflow (see supplemental video). In \Cref{fig:sup:interpolation} we show results for blending with a saliency and depth mask.

\FloatBarrier

\begin{figure*}
  \centering
      \begin{subfigure}{0.24\textwidth}
          \centering
          \includegraphics[width=\textwidth]{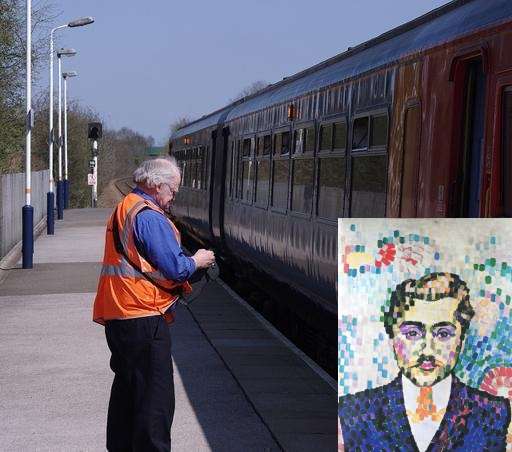}
      \end{subfigure}
      \begin{subfigure}{0.24\textwidth}
          \centering
        \includegraphics[width=\textwidth]{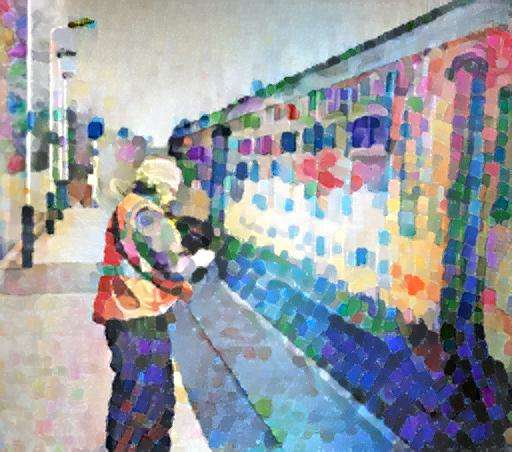}
      \end{subfigure}
      \begin{subfigure}{0.24\textwidth}
          \centering
          \includegraphics[width=\textwidth]{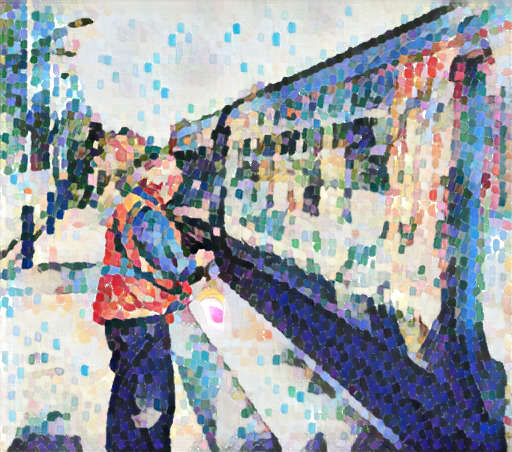}
      \end{subfigure}
      \begin{subfigure}{0.24\textwidth}
          \centering
          \includegraphics[width=\textwidth]{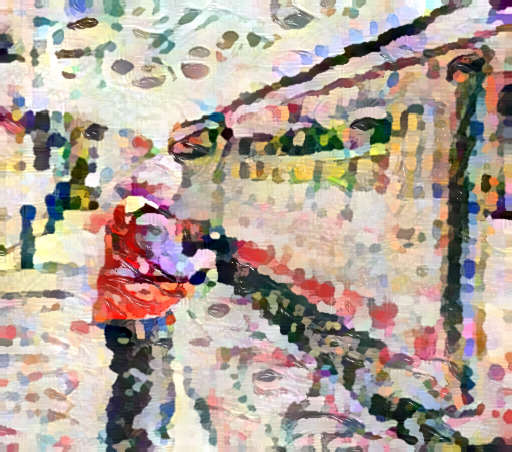}
      \end{subfigure}\\ [0.5ex]
  
      \begin{subfigure}{0.24\textwidth}
          \centering
          \includegraphics[width=\textwidth]{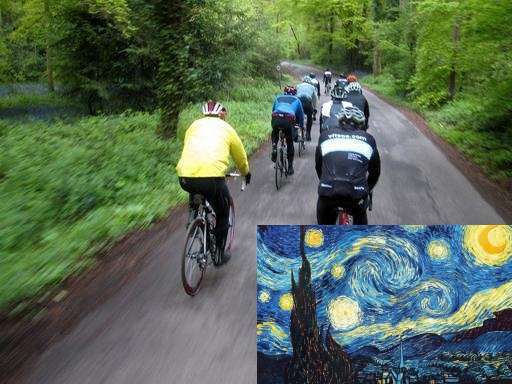}
      \end{subfigure}
      \begin{subfigure}{0.24\textwidth}
          \centering
        \includegraphics[width=\textwidth]{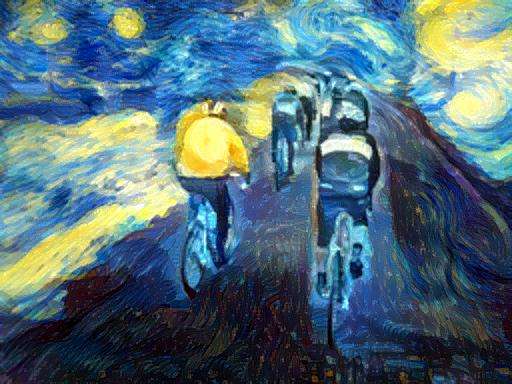}
      \end{subfigure}
      \begin{subfigure}{0.24\textwidth}
          \centering
          \includegraphics[width=\textwidth]{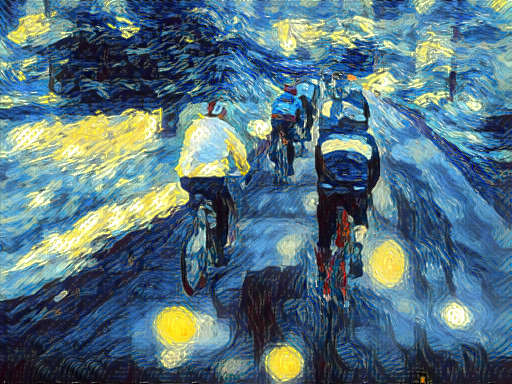}
      \end{subfigure}
      \begin{subfigure}{0.24\textwidth}
          \centering
          \includegraphics[width=\textwidth]{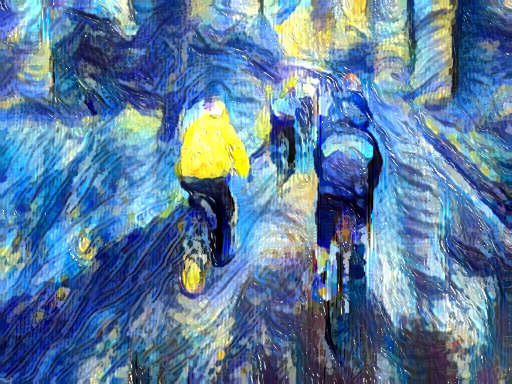}
      \end{subfigure}\\ [0.5ex]

      \begin{subfigure}{0.24\textwidth}
          \centering
          \includegraphics[width=\textwidth]{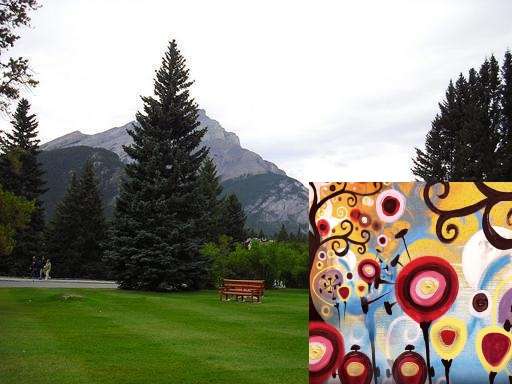}
      \end{subfigure}
      \begin{subfigure}{0.24\textwidth}
          \centering
        \includegraphics[width=\textwidth]{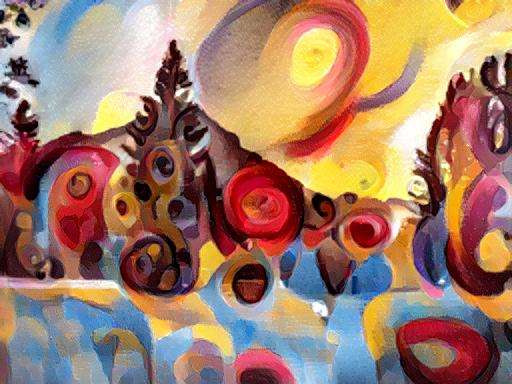}
      \end{subfigure}
      \begin{subfigure}{0.24\textwidth}
          \centering
          \includegraphics[width=\textwidth]{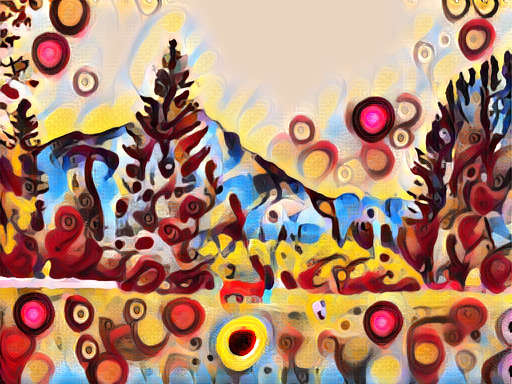}
      \end{subfigure}
      \begin{subfigure}{0.24\textwidth}
          \centering
          \includegraphics[width=\textwidth]{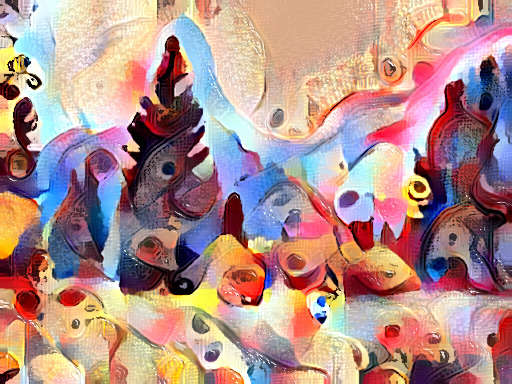}
      \end{subfigure}\\ [0.5ex]
  
      \begin{subfigure}{0.24\textwidth}
          \centering
          \includegraphics[width=\textwidth]{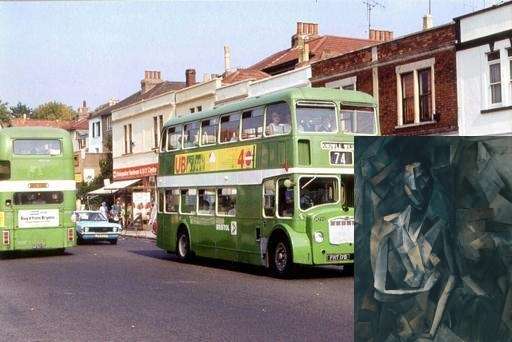}
      \end{subfigure}
      \begin{subfigure}{0.24\textwidth}
          \centering
        \includegraphics[width=\textwidth]{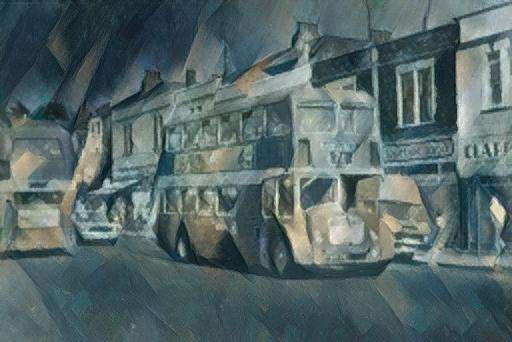}
      \end{subfigure}
      \begin{subfigure}{0.24\textwidth}
          \centering
          \includegraphics[width=\textwidth]{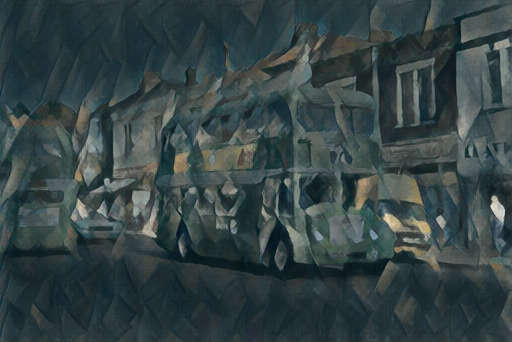}
      \end{subfigure}
      \begin{subfigure}{0.24\textwidth}
          \centering
          \includegraphics[width=\textwidth]{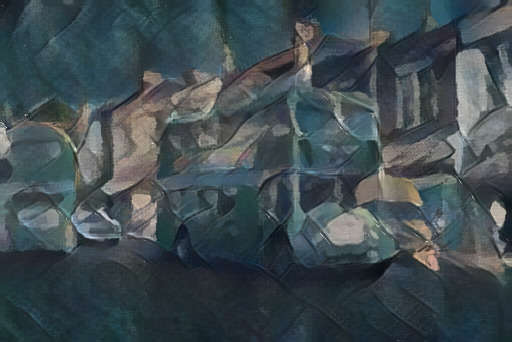}
      \end{subfigure}\\ [0.5ex]
  
      \begin{subfigure}{0.24\textwidth}
          \centering
          \includegraphics[width=\textwidth]{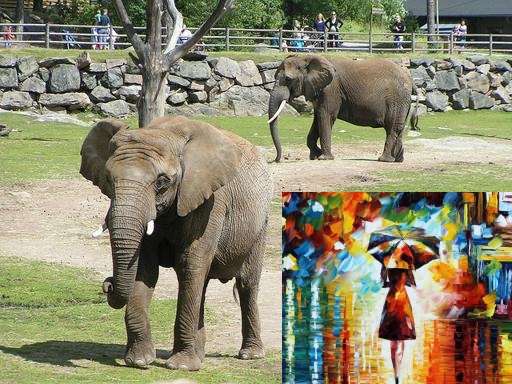}
      \end{subfigure}
      \begin{subfigure}{0.24\textwidth}
          \centering
        \includegraphics[width=\textwidth]{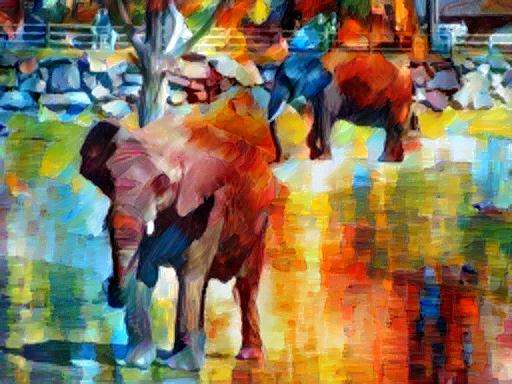}
      \end{subfigure}
      \begin{subfigure}{0.24\textwidth}
          \centering
          \includegraphics[width=\textwidth]{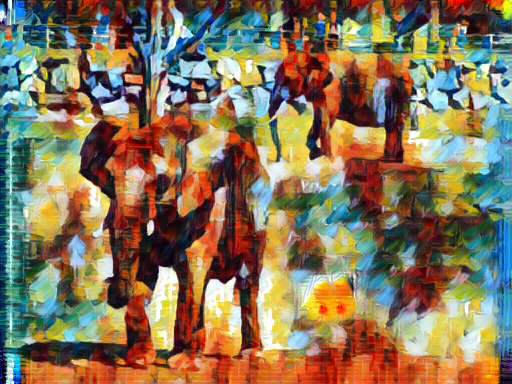}
      \end{subfigure}
      \begin{subfigure}{0.24\textwidth}
          \centering
          \includegraphics[width=\textwidth]{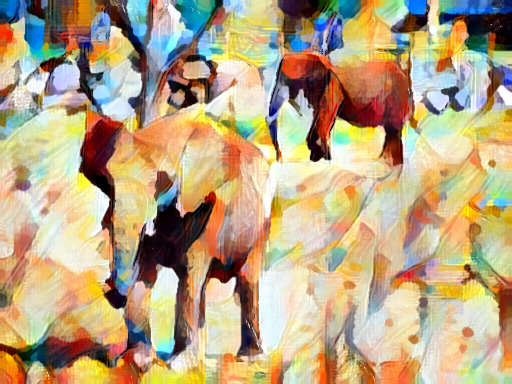}
      \end{subfigure}\\ [0.5ex]
      
      \begin{subfigure}{0.24\textwidth}
          \centering
          \includegraphics[width=\textwidth]{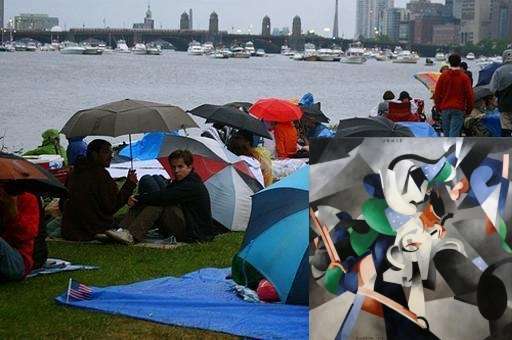}
          \caption{Content \& Style}
      \end{subfigure}
      \begin{subfigure}{0.24\textwidth}
          \centering
        \includegraphics[width=\textwidth]{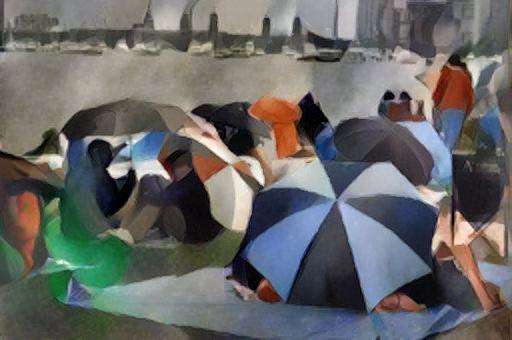}
        \caption{Optimization}
        \label{fig:sup:nst_comparison_grid:optim}
      \end{subfigure}
      \begin{subfigure}{0.24\textwidth}
          \centering
          \includegraphics[width=\textwidth]{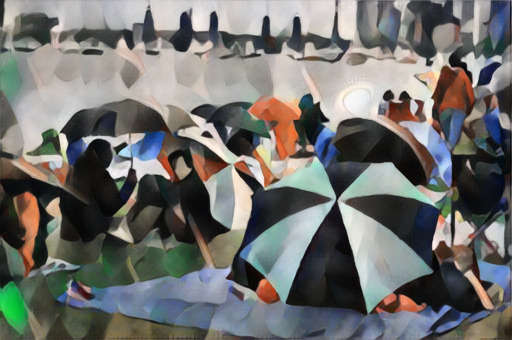}
          \caption{$\text{PPN}_\mathit{sst}$}
      \end{subfigure}
      \begin{subfigure}{0.24\textwidth}
          \centering
          \includegraphics[width=\textwidth]{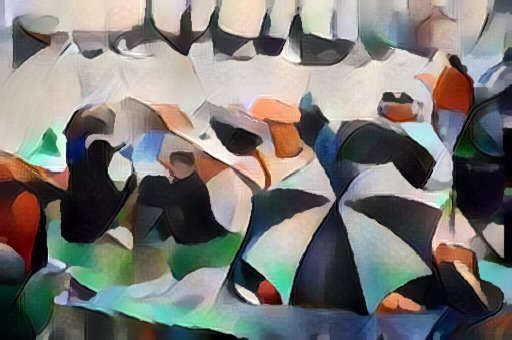}
          \caption{$\text{PPN}_\mathit{arb}$}
      \end{subfigure}\\
      \caption{Comparison of parameter optimization and prediction for the task of style transfer. For each method, we use the style target shown on the left.  We compare optimization, the single style PPN ($\text{PPN}_\mathit{sst}$) and the arbitrary style PPN ($\text{PPN}_\mathit{arb}$). During parameter optimization (\subref{fig:sup:nst_comparison_grid:optim}), we optimize $\mathcal{L}_{\mathit{target}}$ using $\ell_1$ loss to the output of STROTSS~\cite{kolkin2019style}. Each method is capable of predicting or optimizing parameters that produce artifact-free textures and match the visual fidelity of their CNN-based counterparts. We show results predicted on 5000 segments of SLIC~\cite{achanta2012slic}.    
      }
      \label{fig:sup:nst_comparison_grid}
  \end{figure*}

\newcommand{\pipelinecompspy}[3]{
    \begin{tikzpicture}[every node/.style={inner sep=0,outer sep=0}]
        \begin{scope}[
			inner sep = 0pt,outer sep=0,spy using outlines={rectangle, red, magnification=2}
                ]
        \node (n0)  { \includegraphics[width=\textwidth]{#1}};
        \spy [red,size=1.2cm] on (#2,#3) in node[anchor=south west,inner sep=0pt] at (n0.south west);
        \end{scope}
    \end{tikzpicture}
}

\begin{figure*}[ht]
\centering
\captionsetup[subfigure]{labelformat=empty}
    \begin{subfigure}[T]{0.193\textwidth}
        \includegraphics[width=\textwidth]{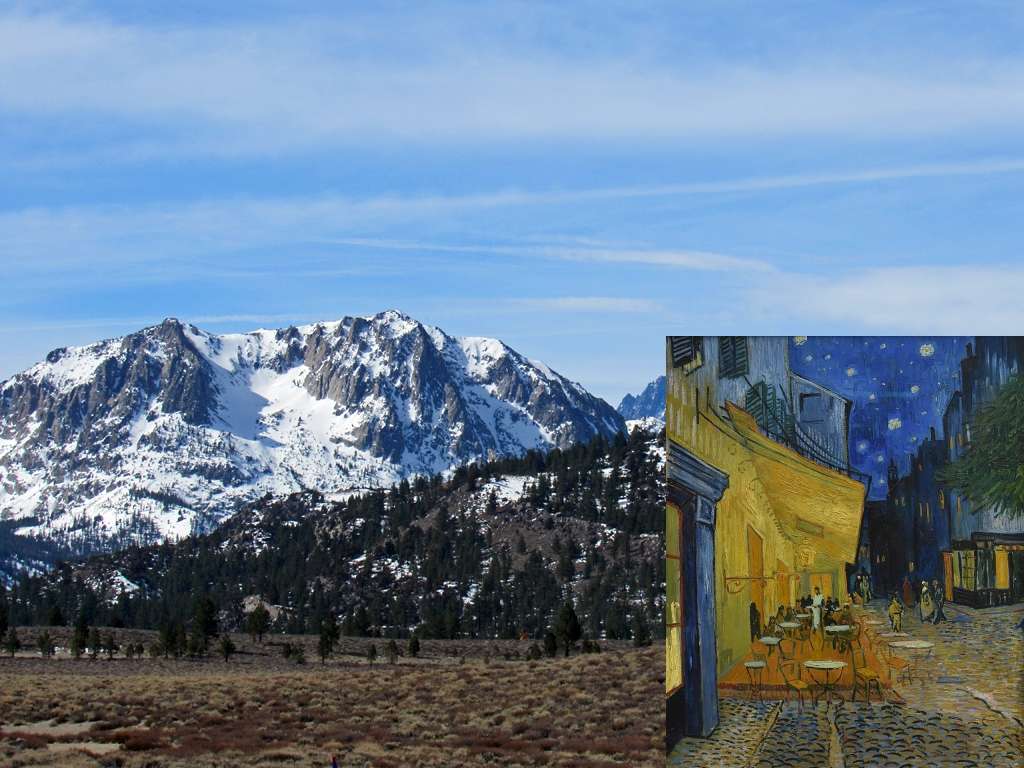}
    \end{subfigure}
    \begin{subfigure}[T]{0.193\textwidth}
        \pipelinecompspy{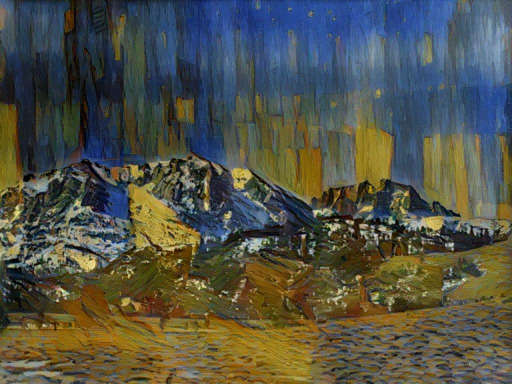}{0.2cm}{0.2cm}
        \vspace{-0.4cm}
    \end{subfigure}
    \begin{subfigure}[T]{0.193\textwidth}
        \pipelinecompspy{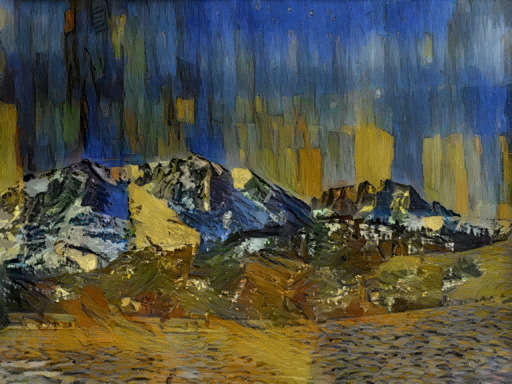}{0.2cm}{0.2cm}
        \vspace{-0.4cm}
    \end{subfigure}
    \begin{subfigure}[T]{0.193\textwidth}
        \pipelinecompspy{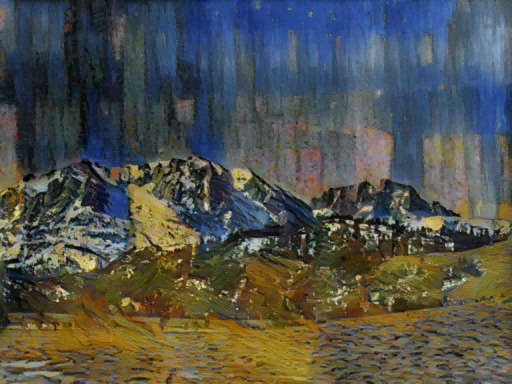}{0.2cm}{0.2cm}
        \vspace{-0.4cm}
    \end{subfigure}
    \begin{subfigure}[T]{0.193\textwidth}
        \pipelinecompspy{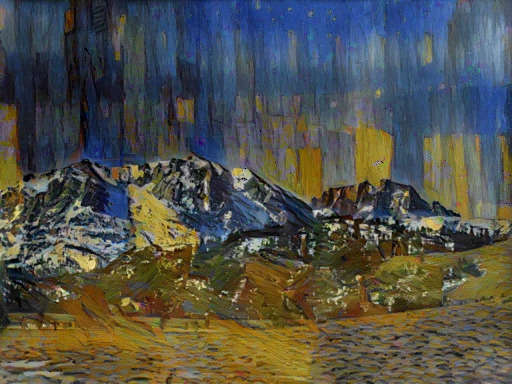}{0.2cm}{0.2cm}
        \vspace{-0.4cm}
    \end{subfigure}\\

    \begin{subfigure}[T]{0.193\textwidth}
        \includegraphics[width=\textwidth]{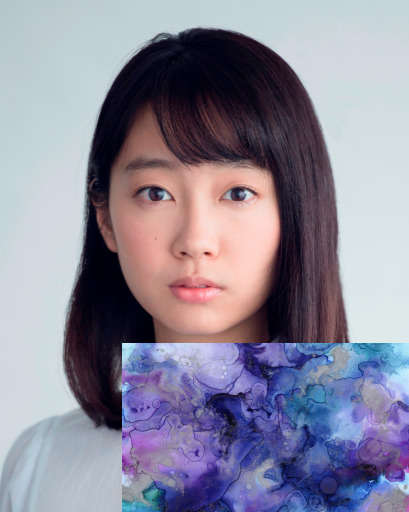}
    \end{subfigure}
    \begin{subfigure}[T]{0.193\textwidth}
        \includegraphics[width=\textwidth]{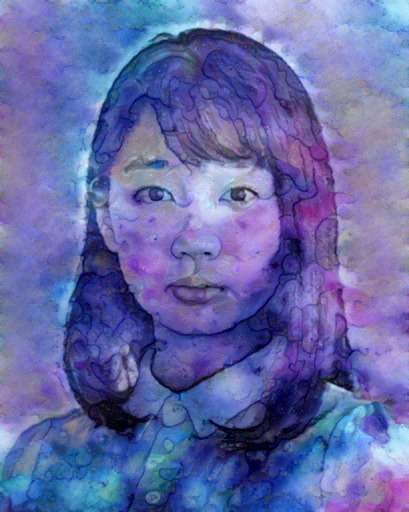}
        \vspace{-0.4cm}
    \end{subfigure}
    \begin{subfigure}[T]{0.193\textwidth}
        \includegraphics[width=\textwidth]{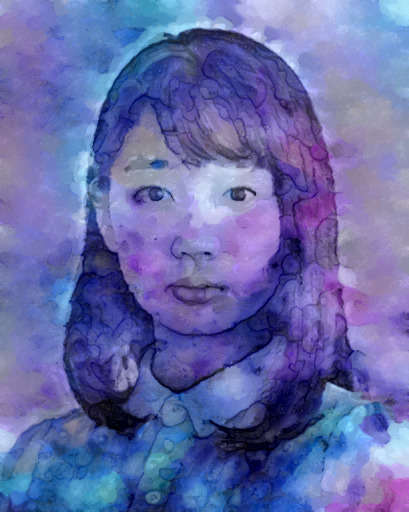}
    \end{subfigure}
    \begin{subfigure}[T]{0.193\textwidth}
        \includegraphics[width=\textwidth]{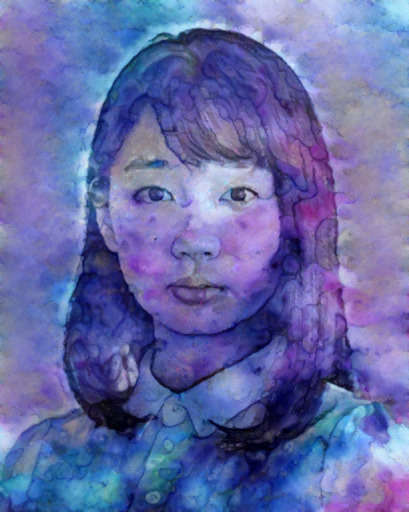}
    \end{subfigure}
    \begin{subfigure}[T]{0.193\textwidth}
        \includegraphics[width=\textwidth]{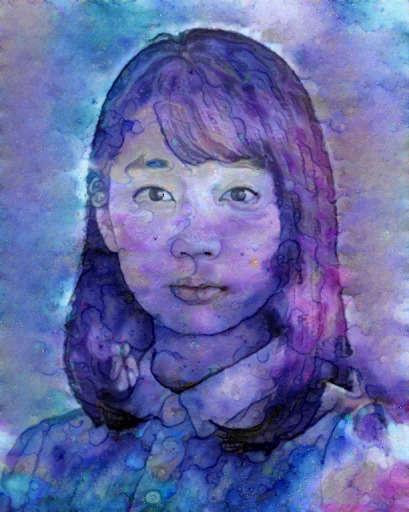}
    \end{subfigure}\\

    \begin{subfigure}[T]{0.193\textwidth}
        \includegraphics[width=\textwidth]{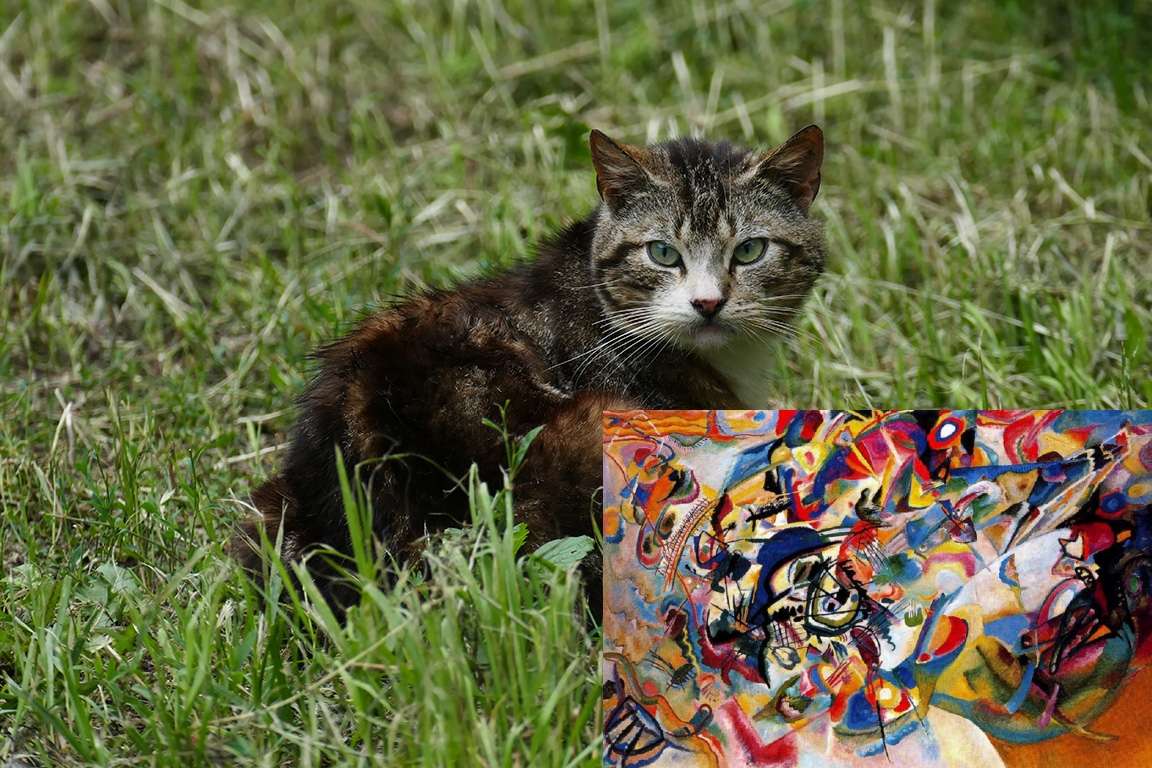}
        \caption{Content \& Style}
    \end{subfigure}
    \begin{subfigure}[T]{0.193\textwidth}
        \pipelinecompspy{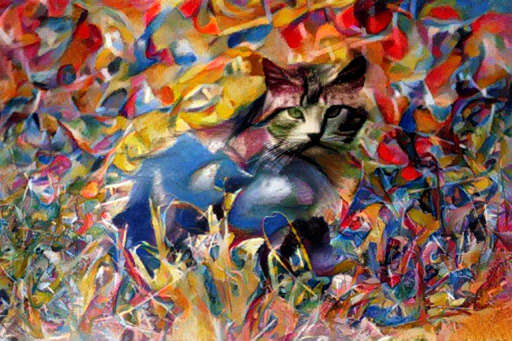}{0.15cm}{-0.2cm}
        \vspace{-0.4cm}
        \caption{STROTSS~\cite{kolkin2019style} Target}
    \end{subfigure}
    \begin{subfigure}[T]{0.193\textwidth}
        \pipelinecompspy{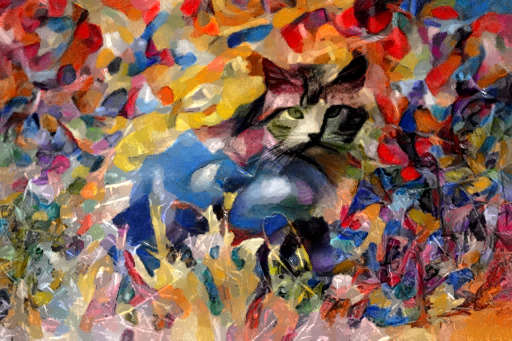}{0.15cm}{-0.2cm}
        \vspace{-0.4cm}
        \caption{Ours}
    \end{subfigure}
    \begin{subfigure}[T]{0.193\textwidth}
        \pipelinecompspy{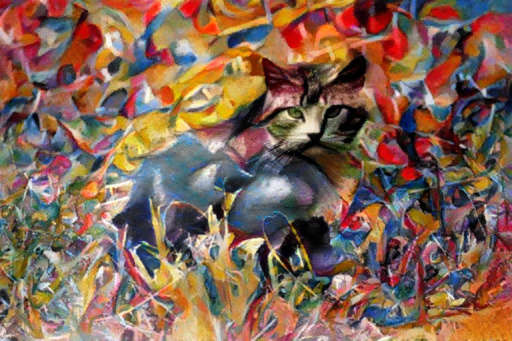}{0.15cm}{-0.2cm}
        \vspace{-0.4cm}
        \caption{Oilpaint~\cite{lotzsch2022wise}}
    \end{subfigure}
    \begin{subfigure}[T]{0.193\textwidth}
        \pipelinecompspy{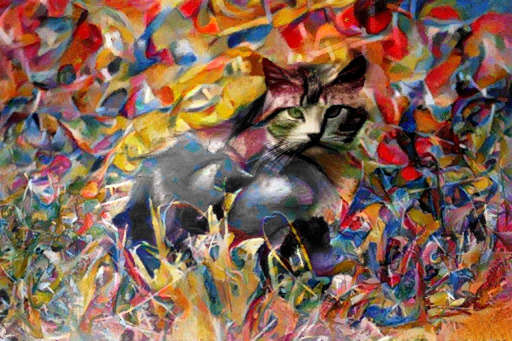}{0.15cm}{-0.2cm}
        \vspace{-0.4cm}
        \caption{Watercolor~\cite{lotzsch2022wise}}
    \end{subfigure}\\
    \caption{Comparison of our proposed Arbitrary Style Pipeline with the differentiable, style-specific Oilpaint~\cite{semmo2016image} and Watercolor~\cite{wang2014towards} effect pipelines by L\"otzsch \etal~\cite{lotzsch2022wise}. The pipelines were optimized to match the stylization result generated using the STROTSS~\cite{kolkin2019style} NST via $\ell_1$ loss.}
    \label{fig:sup:pipeline_comparison}
\end{figure*}

\ifdefined\isArxiv
\else
    \vspace{-0.5cm}
    \adamlisterCompareFig
\fi

\newlength{\figsizerainprincess}
\setlength{\figsizerainprincess}{0.41\textwidth}

\newlength{\hspclip}
\setlength{\hspclip}{0.05cm}

\vspace{-0.5cm}    
\begin{figure*}
    \centering
    \begin{subfigure}[b]{\figsizerainprincess}
        \includegraphics[width=\textwidth]{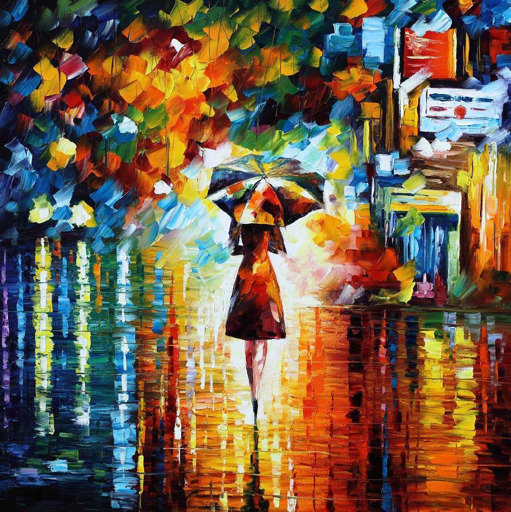}
        \caption{Original}
    \end{subfigure}\hspace{\hspclip}
    \begin{subfigure}[b]{\figsizerainprincess}
        \includegraphics[width=\textwidth]{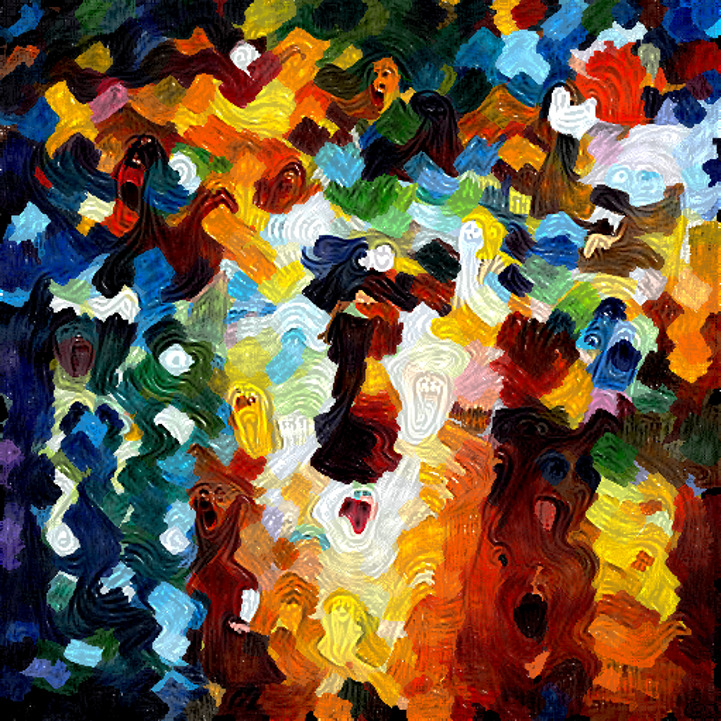}
        \caption{"the scream"}
    \end{subfigure}\hspace{\hspclip}
    \begin{subfigure}[b]{\figsizerainprincess}
        \includegraphics[width=\textwidth]{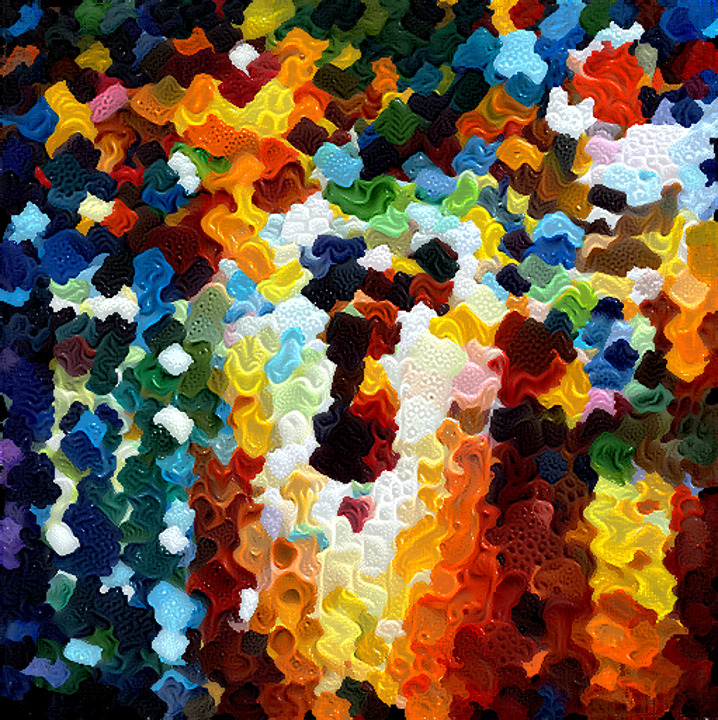}
        \caption{"foam"}
    \end{subfigure}\hspace{\hspclip}
    \begin{subfigure}[b]{\figsizerainprincess}
        \includegraphics[width=\textwidth]{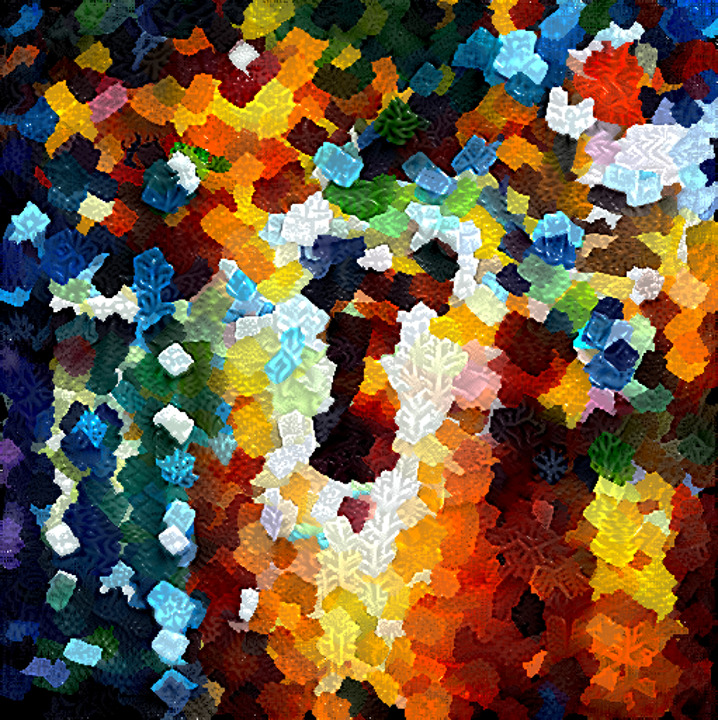}
        \caption{"icy frost"}
    \end{subfigure}\hspace{\hspclip}
    \begin{subfigure}[b]{\figsizerainprincess}
        \includegraphics[width=\textwidth]{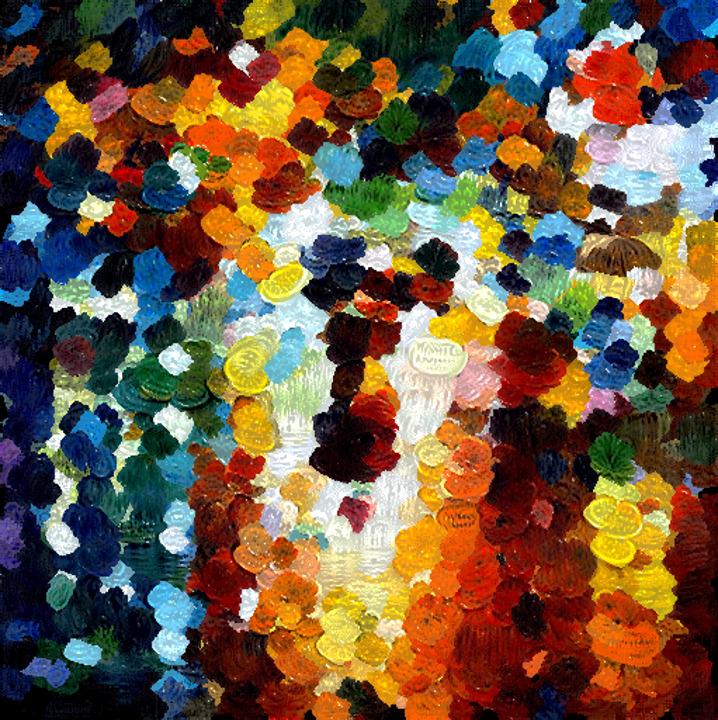}
        \caption{"impressionism by Monet"}
    \end{subfigure}\hspace{\hspclip}
    \begin{subfigure}[b]{\figsizerainprincess}
        \includegraphics[width=\textwidth]{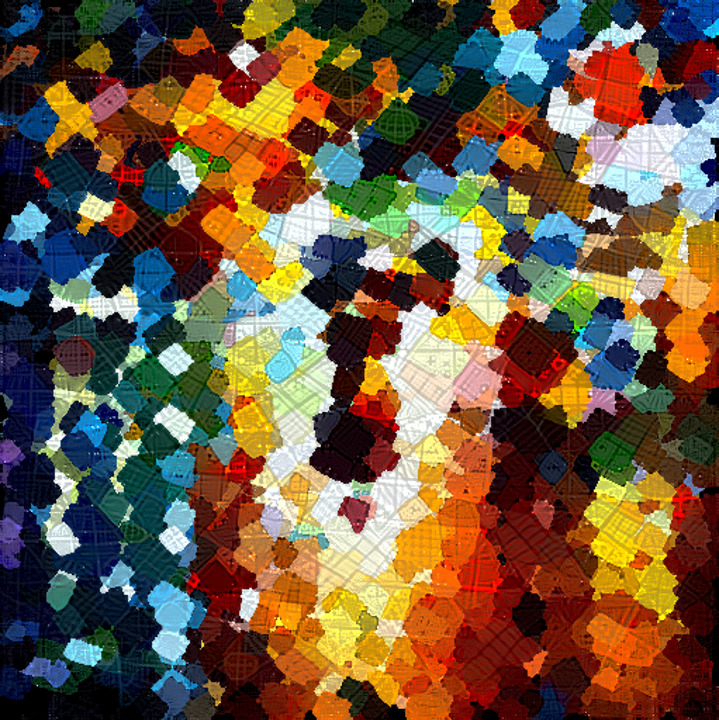}
        \caption{"regular grid"}
    \end{subfigure}
   \caption{Texture style transfer of "rain princess" using CLIPstyler \cite{kwon2022clipstyler}-losses. PtF-Circle is used for segmentation.
   }
    \label{fig:sup:clipstyler_rainprincess_row}
\end{figure*}

\newlength{\figsizecomparisons}
\setlength{\figsizecomparisons}{0.33\textwidth}

\begin{figure*}[tb]
  \begin{subfigure}{\figsizecomparisons}%
		\includegraphics[width=\textwidth]{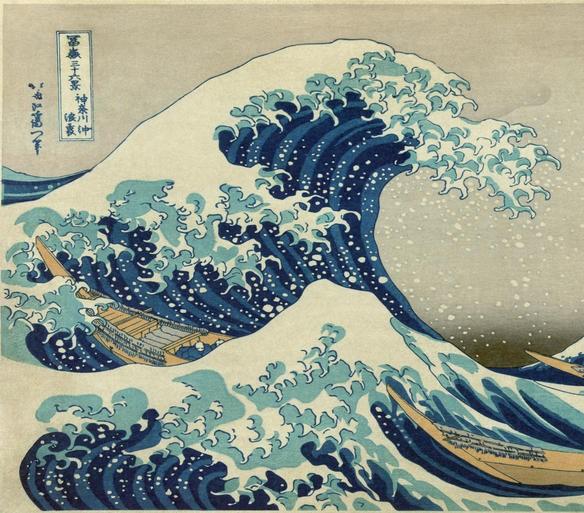}%
             \caption{Original}%
	\end{subfigure}\hfill
	\begin{subfigure}{\figsizecomparisons}%
		\includegraphics[width=\textwidth]{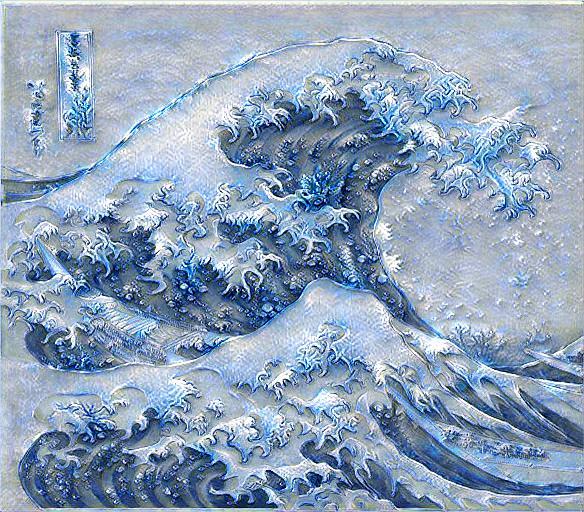}%
      \caption{CLIPstyler \cite{kwon2022clipstyler}}%
	\end{subfigure}\hfill
  \begin{subfigure}{\figsizecomparisons}%
    \includegraphics[width=\textwidth]{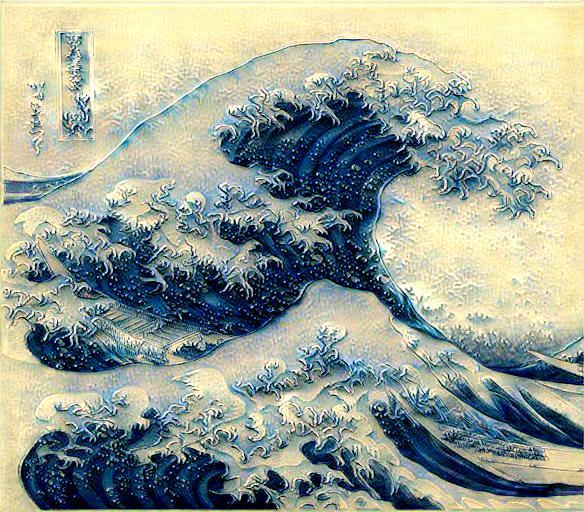}%
    \caption{CLIPstyler-Hist}%
  \end{subfigure}\hfill
  \begin{subfigure}{\figsizecomparisons}%
		\includegraphics[width=\textwidth]{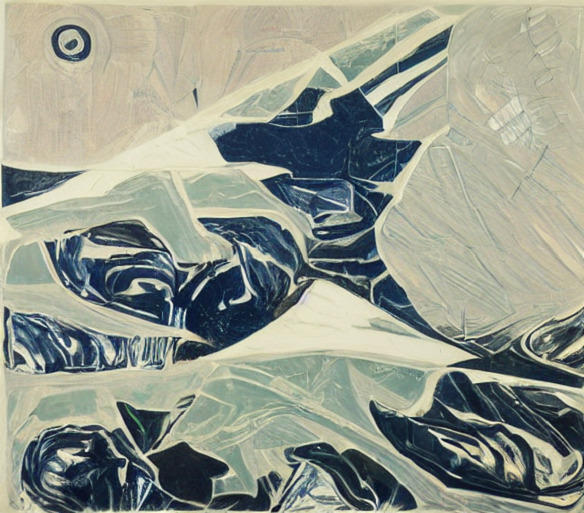}%
        \caption{Stable Diffusion \cite{rombach2022high}}%
	\end{subfigure}\hfill
  \begin{subfigure}{\figsizecomparisons}%
    \includegraphics[width=\textwidth]{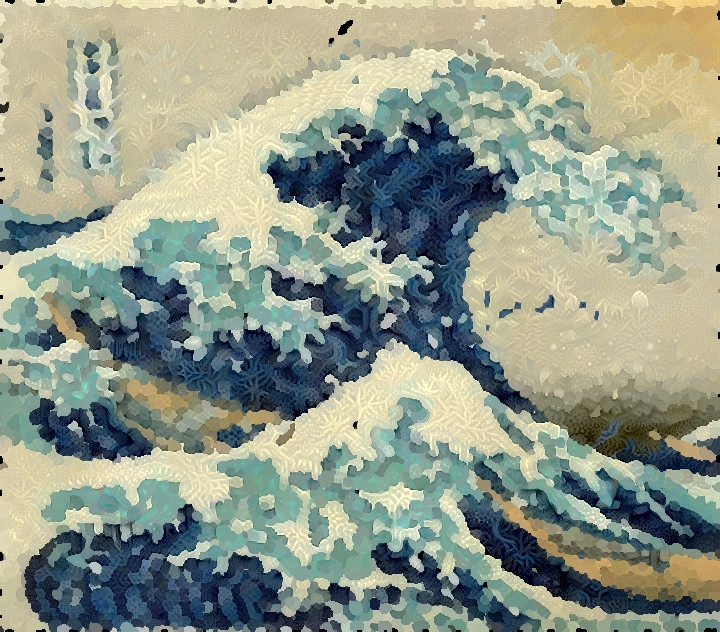}%
    \caption{Ours (fine)}%
  \end{subfigure}\hfill 
  \begin{subfigure}{\figsizecomparisons}%
    \includegraphics[width=\textwidth]{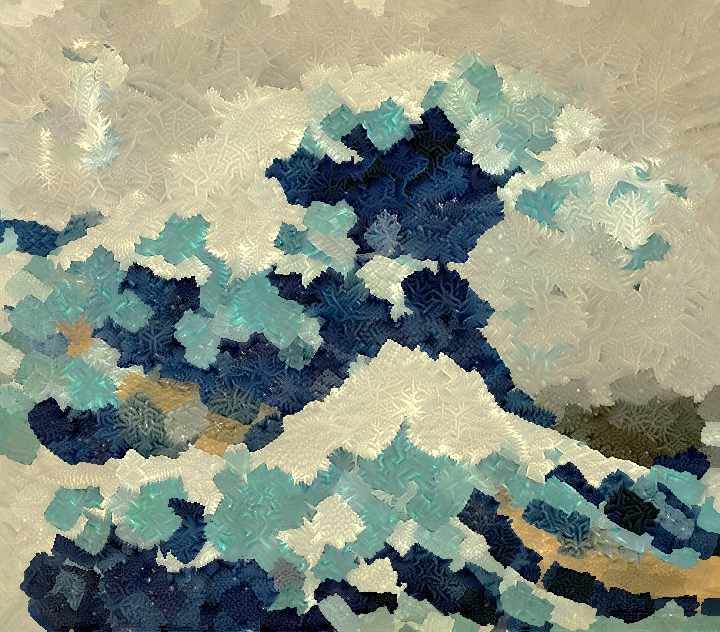}%
    \caption{Ours (coarse)}%
  \end{subfigure}\\
  \begin{subfigure}{\figsizecomparisons}%
		\includegraphics[width=\textwidth]{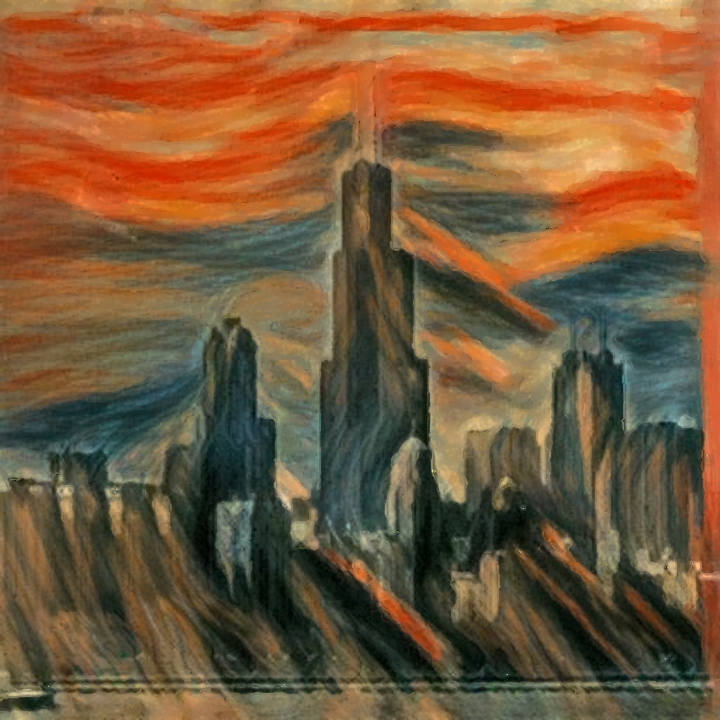}%
            \caption{Original}%
    		\label{fig:sup:rw_compare:orig}%
	\end{subfigure}\hfill
	\begin{subfigure}{\figsizecomparisons}%
		\includegraphics[width=\textwidth]{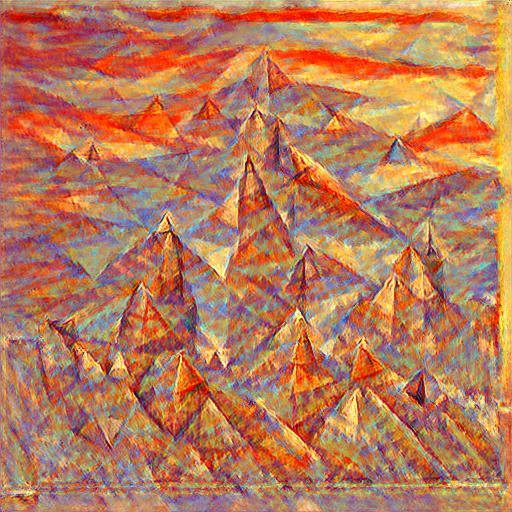}%
    \caption{Clipstyler \cite{kwon2022clipstyler}}%
    		\label{fig:sup:rw_compare:clipstyler}%
	\end{subfigure}\hfill
  \begin{subfigure}{\figsizecomparisons}%
    \includegraphics[width=\textwidth]{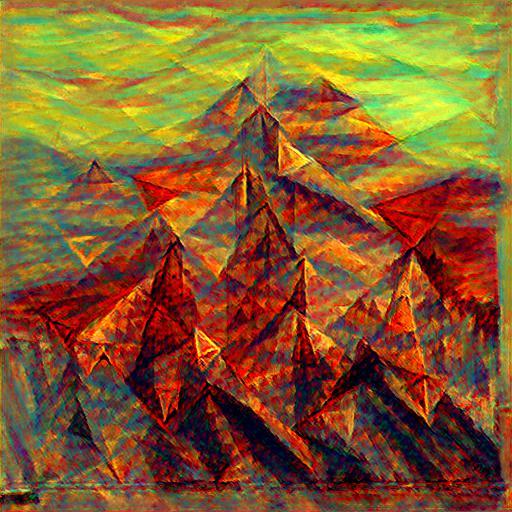}%
    \label{fig:sup:rw_compare:clipstyler-hist}%
    \caption{Clipstyler-Hist}%
  \end{subfigure}\hfill
  \begin{subfigure}{\figsizecomparisons}%
		\includegraphics[width=\textwidth]{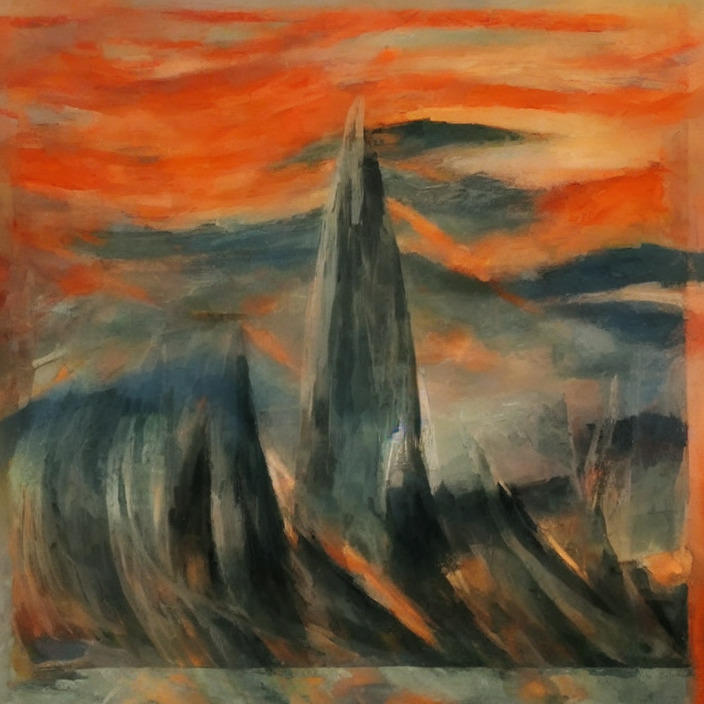}%
		\label{fig:sup:rw_compare:stablediffusion}%
    \caption{\footnotesize Stable Diffusion \cite{rombach2022high}}%
	\end{subfigure}\hfill
  \begin{subfigure}{\figsizecomparisons}%
    \includegraphics[width=\textwidth]{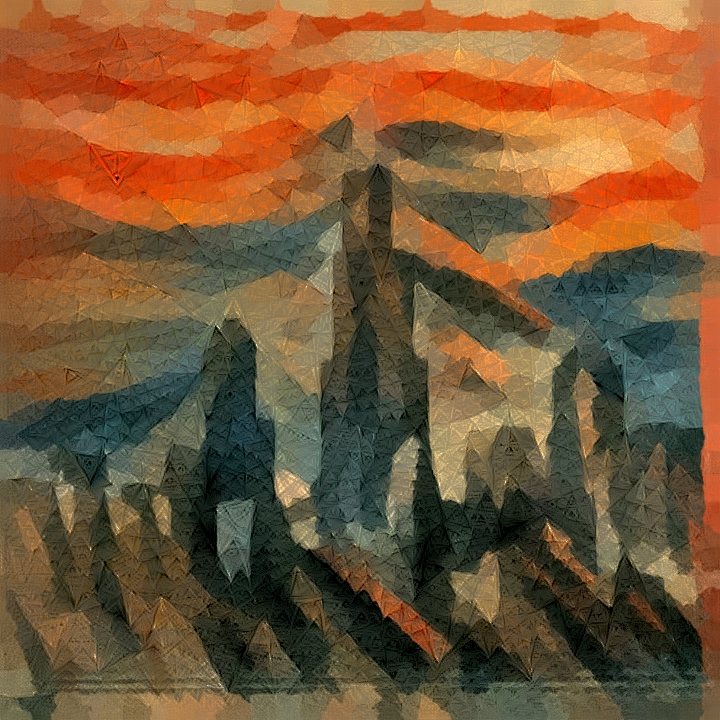}%
    \label{fig:sup:rw_compare:ours_fine}
    \caption{Ours (fine)}%
  \end{subfigure}\hfill 
  \begin{subfigure}{\figsizecomparisons}%
    \includegraphics[width=\textwidth]{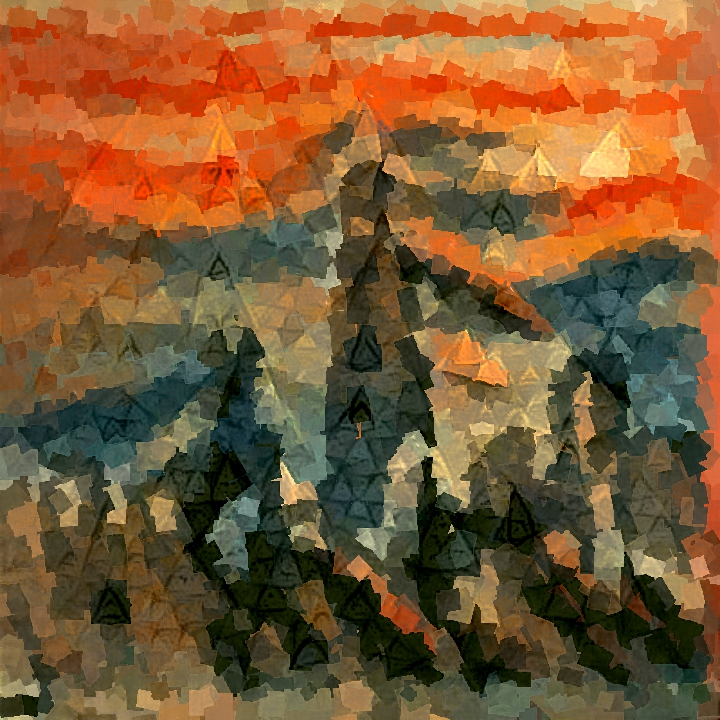}%
    \label{fig:sup:rw_compare:ours_coarse}%
    \caption{Ours (coarse)}%
  \end{subfigure}
  \caption{Further comparisons to related methods for text-based generation. Text prompts, per row, are 1) "icy frost" and 2) "triangles". }
  \label{fig:sup:rw_compare_clip}
\end{figure*}

\newlength{\figsizecomparisonsnst}
\setlength{\figsizecomparisonsnst}{0.315\textwidth}

\newlength{\hspnst}
\setlength{\hspnst}{0.025cm}

\begin{figure*}[tb]
    \centering
    \begin{subfigure}{\figsizecomparisonsnst}%
            \includegraphics[width=\textwidth]{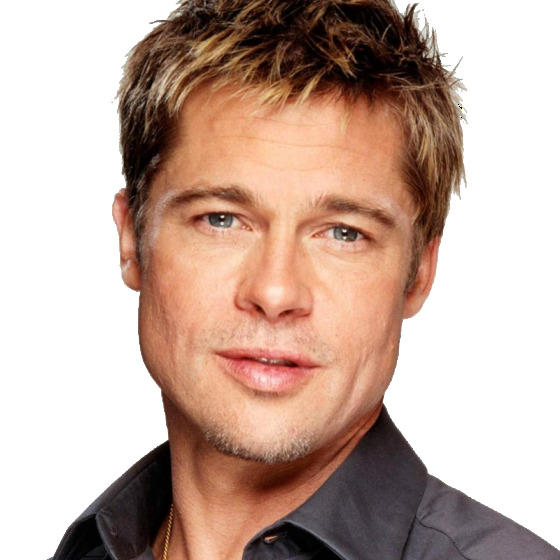}
            \caption{Input}
        \end{subfigure}\hspace{\hspnst}
        \begin{subfigure}{\figsizecomparisonsnst}%
            \includegraphics[width=\textwidth]{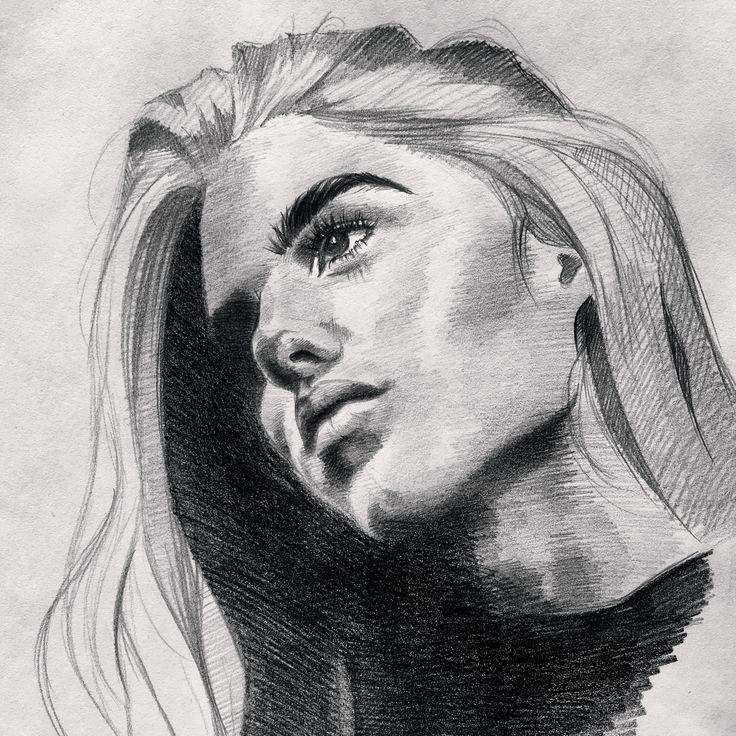}
            \caption{Style}
        \end{subfigure}\hspace{\hspnst}
      \begin{subfigure}{\figsizecomparisonsnst}%
        \includegraphics[width=\textwidth]{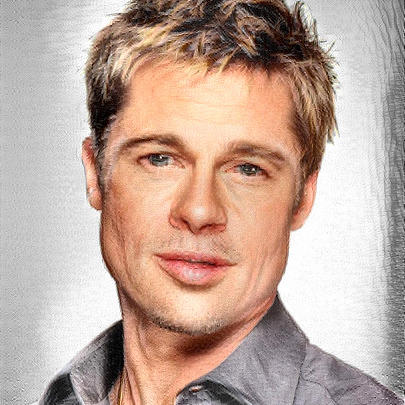}
        \caption{Color preserv. NST \cite{GatysEBHS17}}
      \end{subfigure}\hspace{\hspnst}
      \begin{subfigure}{\figsizecomparisonsnst}%
            \includegraphics[width=\textwidth]{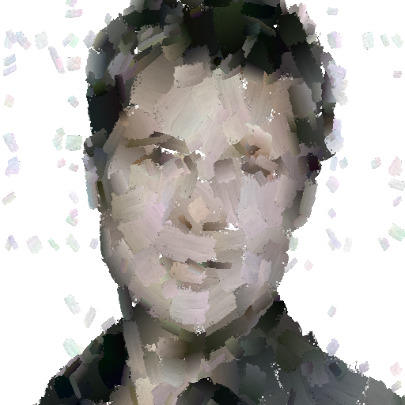}
            \caption{SNP~\cite{zou2021stylized} }
        \end{subfigure}\hspace{\hspnst}
      \begin{subfigure}{\figsizecomparisonsnst}%
        \includegraphics[width=\textwidth]{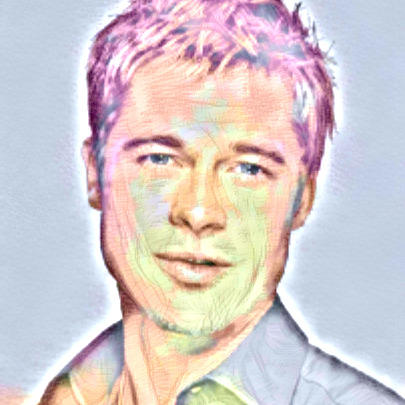}
        \caption{STROTSS~\cite{kolkin2019style}-Hist}
      \end{subfigure}\hspace{\hspnst} 
      \begin{subfigure}{\figsizecomparisonsnst}%
        \includegraphics[width=\textwidth]{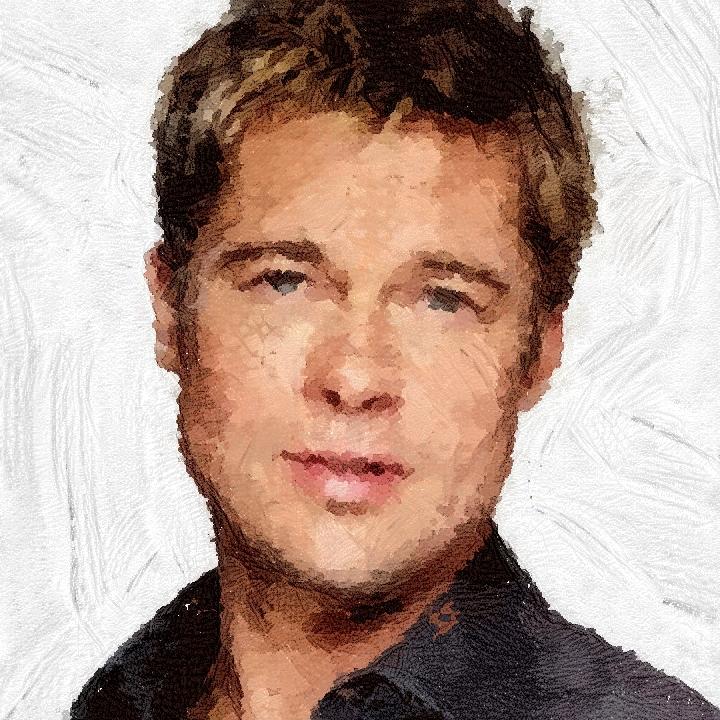}
        \caption{Ours}
      \end{subfigure}\\
      \begin{subfigure}{\figsizecomparisonsnst}%
            \includegraphics[width=\textwidth]{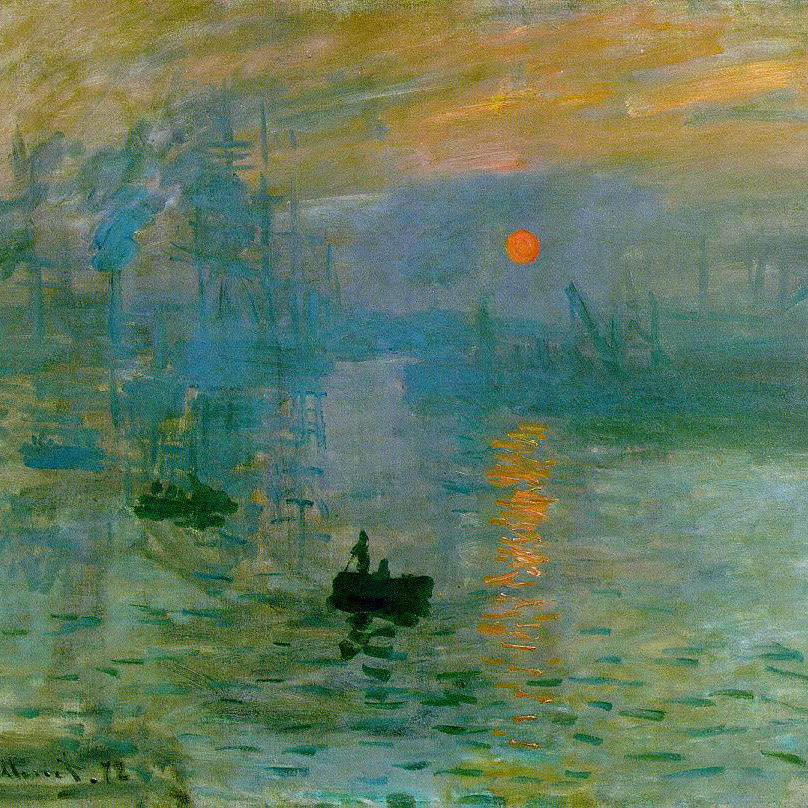}
            \caption{Input}
        \end{subfigure}\hspace{\hspnst}
        \begin{subfigure}{\figsizecomparisonsnst}%
            \includegraphics[width=\textwidth]{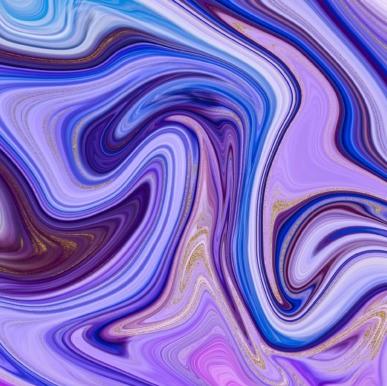}
            \caption{Style}
        \end{subfigure}\hspace{\hspnst}
      \begin{subfigure}{\figsizecomparisonsnst}%
        \includegraphics[width=\textwidth]{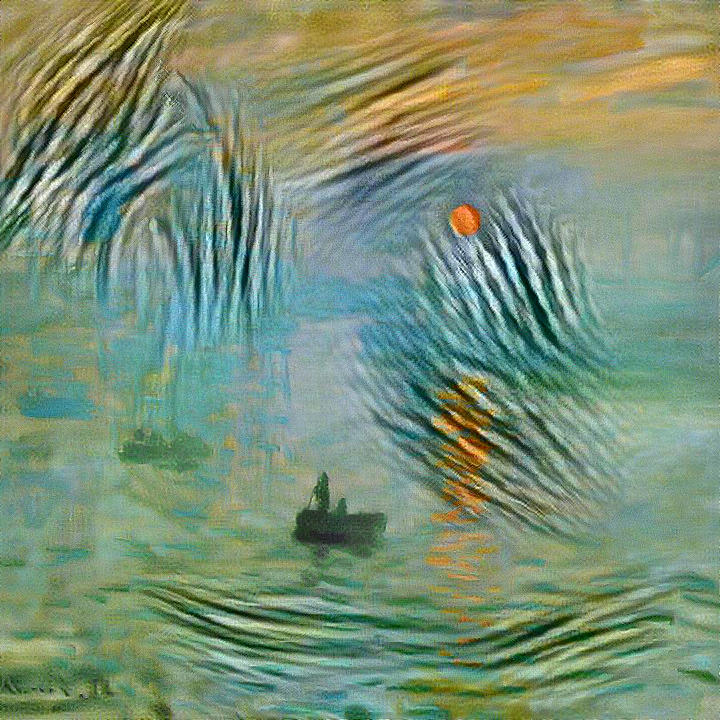}
        \caption{Color preserv. NST \cite{GatysEBHS17}}
      \end{subfigure}\hspace{\hspnst}
      \begin{subfigure}{\figsizecomparisonsnst}%
            \includegraphics[width=\textwidth]{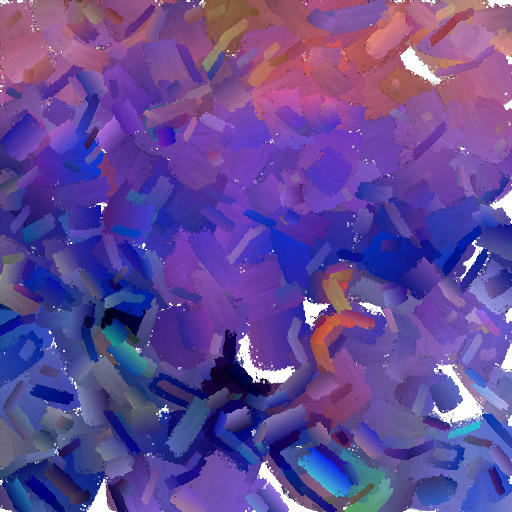}
            \caption{SNP~\cite{zou2021stylized} }
        \end{subfigure}\hspace{\hspnst}
      \begin{subfigure}{\figsizecomparisonsnst}%
        \includegraphics[width=\textwidth]{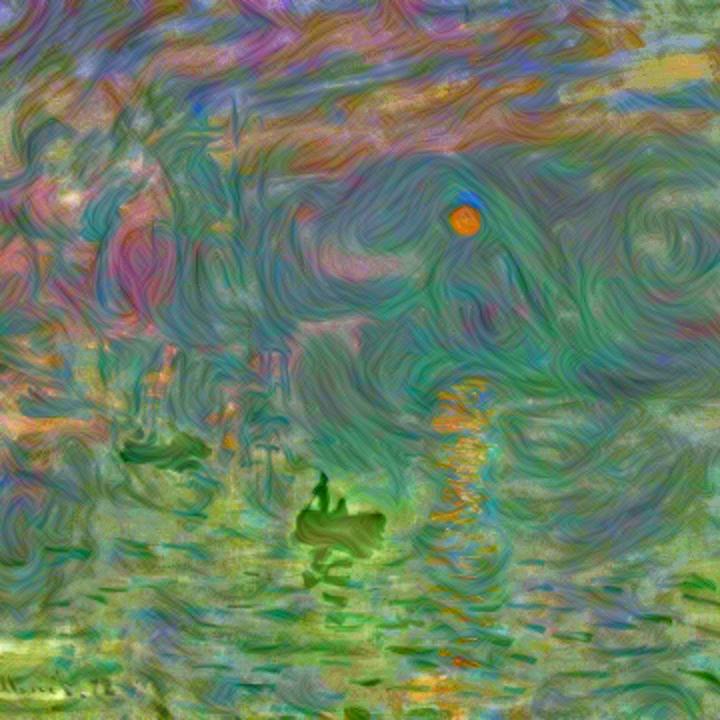}
        \caption{STROTSS~\cite{kolkin2019style}-Hist}
      \end{subfigure}\hspace{\hspnst} 
      \begin{subfigure}{\figsizecomparisonsnst}%
        \includegraphics[width=\textwidth]{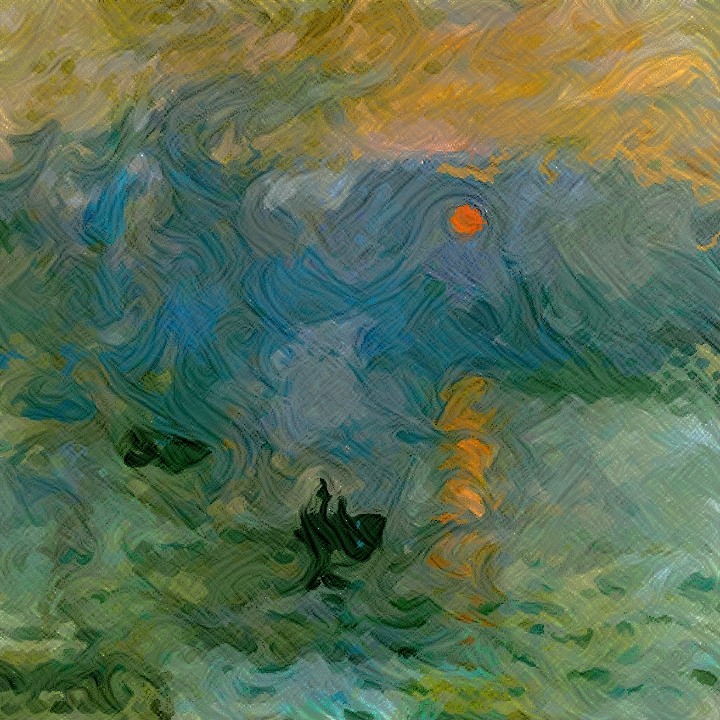}
        \caption{Ours}
      \end{subfigure}
      \caption{Further comparison of image-based style transfer methods with color preservation and our proposed method. We use STROTSS~\cite{kolkin2019style} with Histogram Matching Loss~\cite{jonschkowski2016end}.}
      \label{fig:sup:rw_compare_strotss}
    \end{figure*}

\begin{figure*}
    \centering
    \begin{subfigure}{0.48\linewidth}
        \centering
        \includegraphics[width=\linewidth]{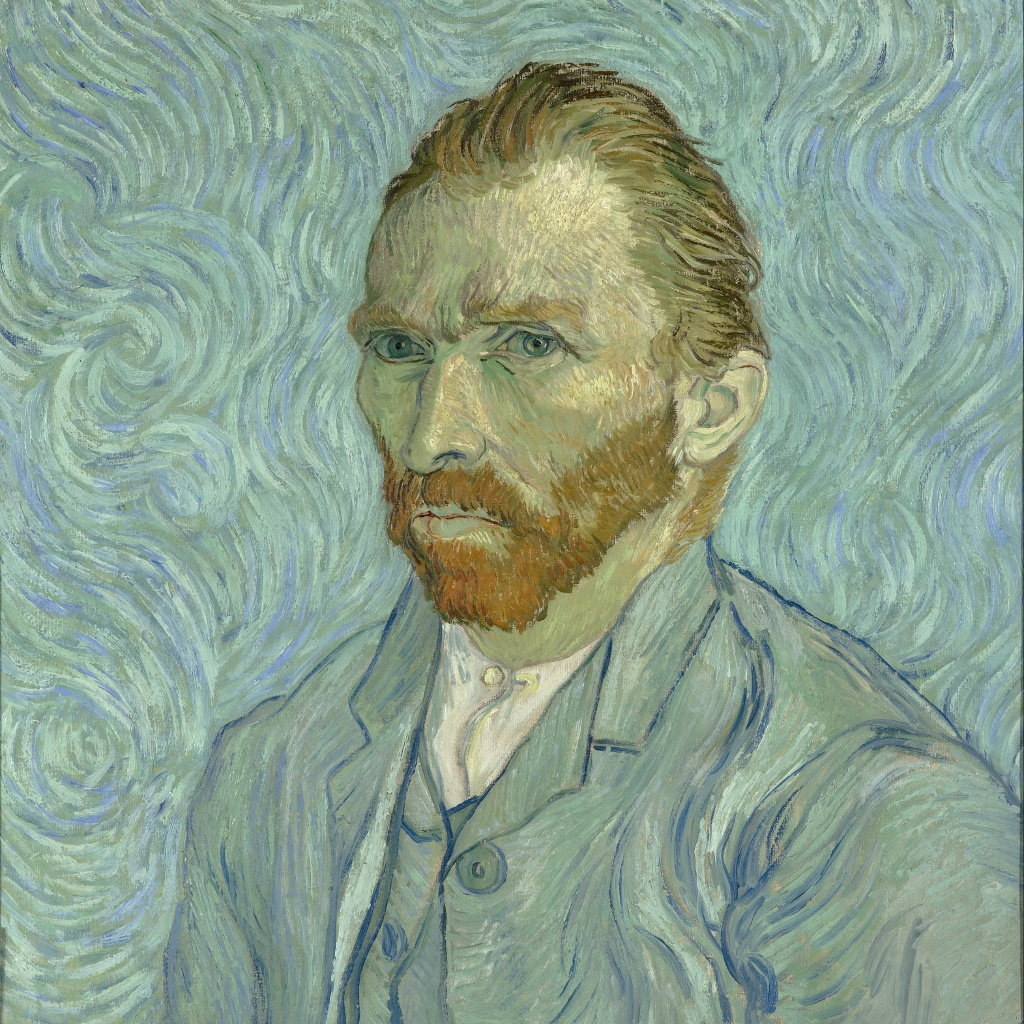}
        \caption{Input}
    \end{subfigure}
    \begin{subfigure}{0.48\linewidth}
        \centering
        \includegraphics[width=\linewidth]{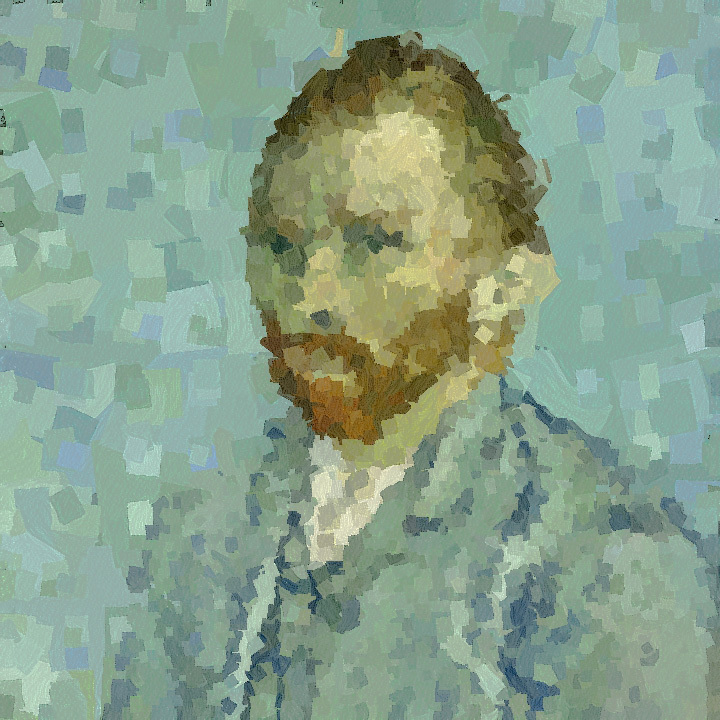}
        \caption{$S$=Ptf-rect. Texture is re-optimized with STROTSS~\cite{kolkin2019style}.}
    \end{subfigure}
    \begin{subfigure}{0.48\linewidth}
        \centering
        \includegraphics[width=\linewidth]{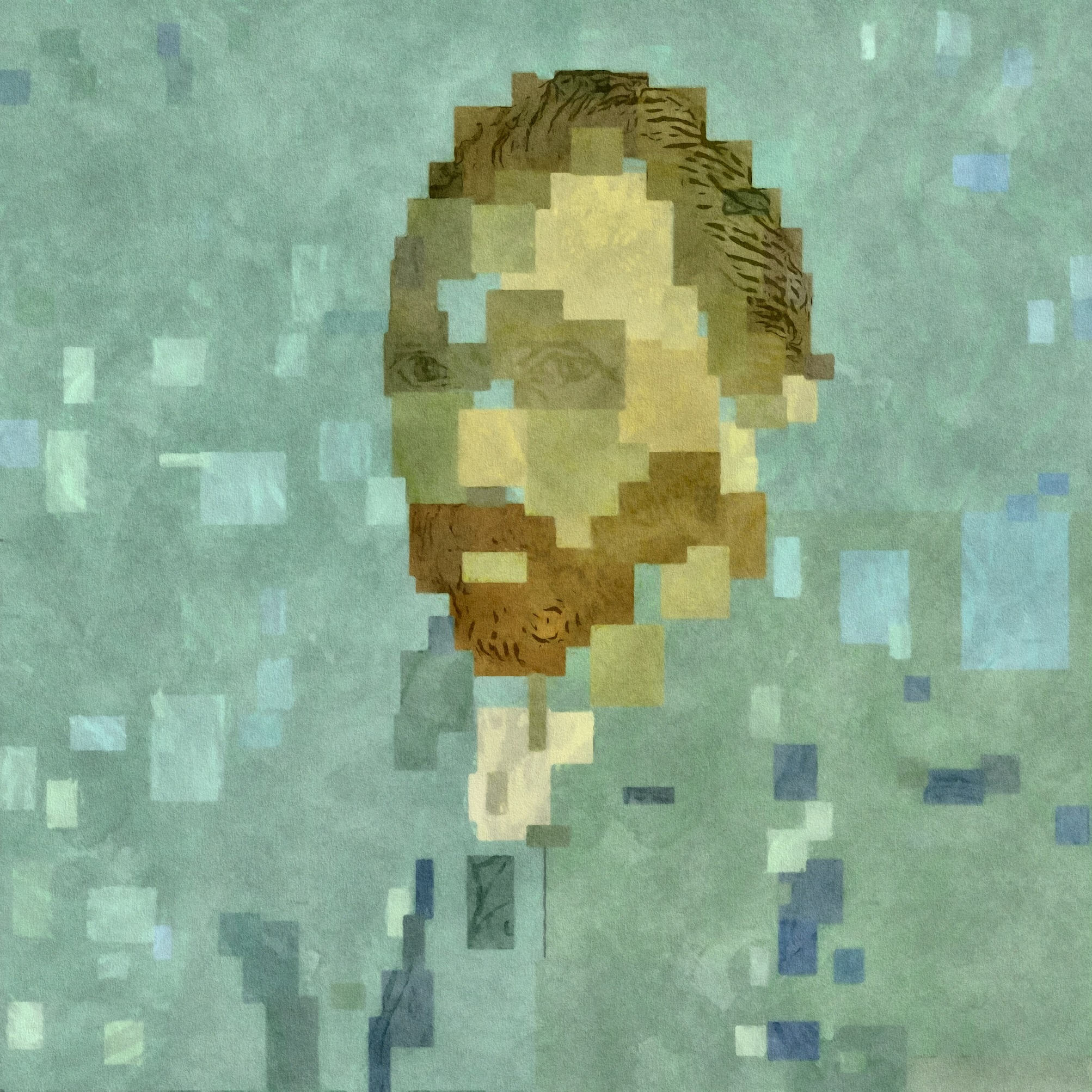}
        \caption{\footnotesize $S$=SNP - 8bit art blocks. Texture is reoptimized with STROTSS~\cite{kolkin2019style}.}
        \label{fig:sup:goghabstract:snp}
    \end{subfigure}
    \begin{subfigure}{0.48\linewidth}
        \centering
        \includegraphics[width=\linewidth]{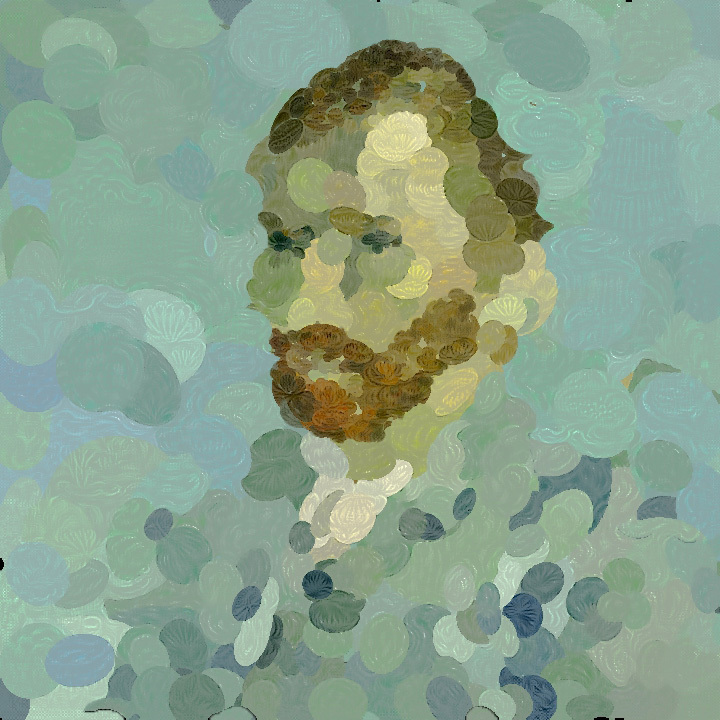}
        \caption{\footnotesize $S$ = PTf-Circles. Texture is reoptimized with CLIPStyler \cite{kwon2022clipstyler}.}
        \label{fig:sup:goghabstract:waterlilies}
    \end{subfigure}
    \caption{Full images of the teaser figure. We compare different geometric abstraction primitives from segmentation using PaintTransformer \cite{liu2021paint} (Ptf) and stylized neural painter \cite{liu2021paint} (SNP) with subsequent reoptimization. See Fig. 3 of the main paper for the outputs of the segmentation method ($I_a$) of the respective images. We reoptimize with a canvas texture as target. In (\subref{fig:sup:goghabstract:snp}), local contour enhancements are made in a final step. In (\subref{fig:sup:goghabstract:waterlilies}) we reoptimize with "round brushstrokes in the style of monet" as target. }
    \label{fig:sup:goghabstract}
\end{figure*}

\begin{figure*}
    \centering
    \begin{subfigure}{0.3\linewidth}
      \includegraphics[trim={0cm 4.24cm 0cm 0cm},clip,width=\linewidth]{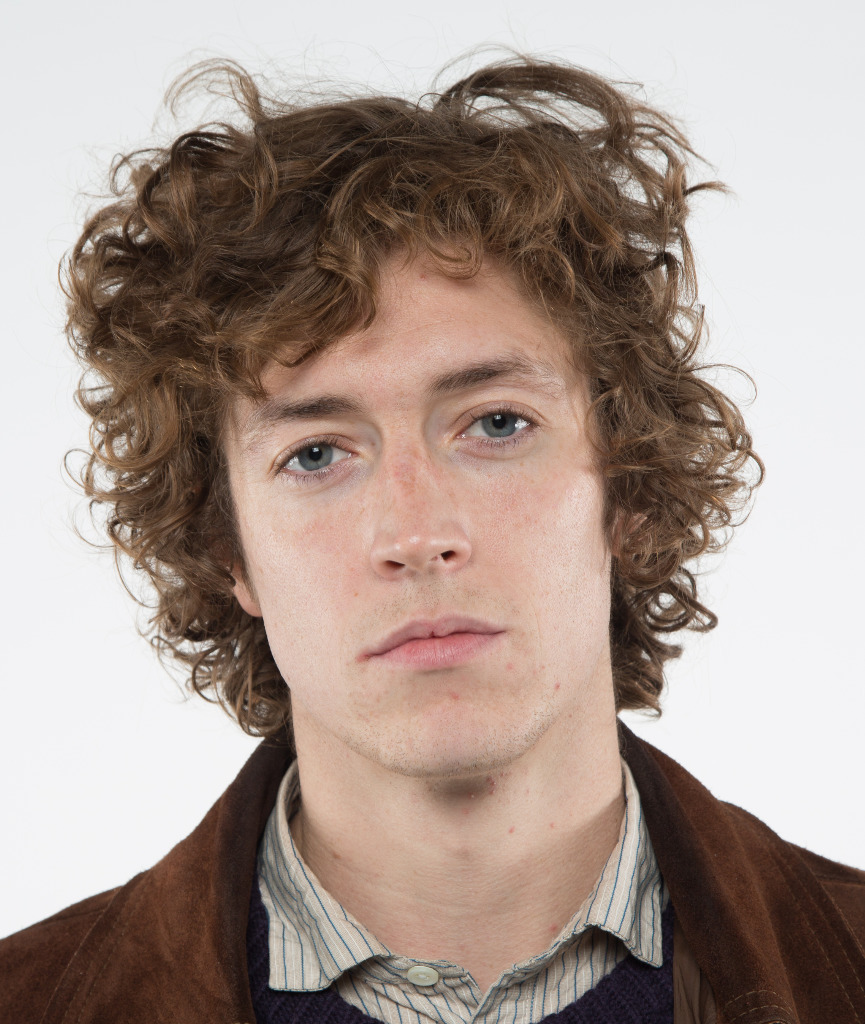}
      \caption{Input}
      \label{realsegmented:input}
    \end{subfigure}
    \begin{subfigure}{0.3\linewidth}
        \includegraphics[trim={0cm 3cm 0cm 0cm},clip,width=\linewidth]{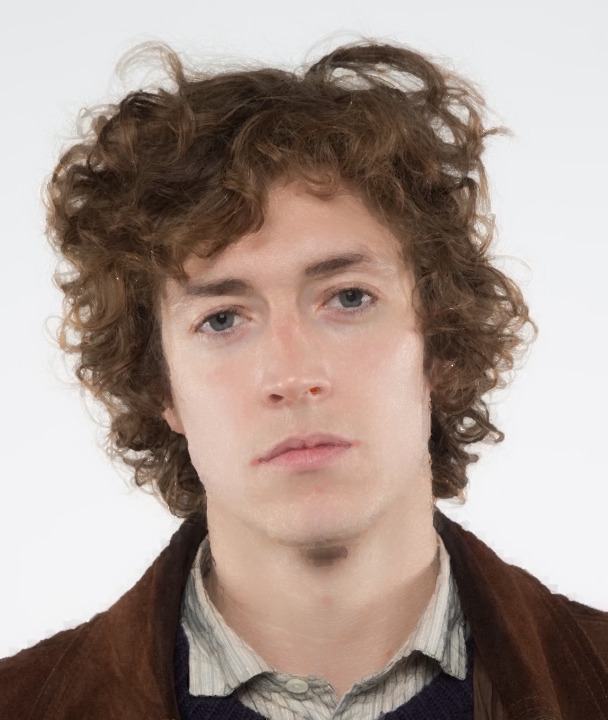}
        \caption{5000 segments}
        \label{realsegmented:5000seg}
    \end{subfigure}
    \begin{subfigure}{0.3\linewidth}
        \includegraphics[trim={0cm 3cm 0cm 0cm},clip,width=\linewidth]{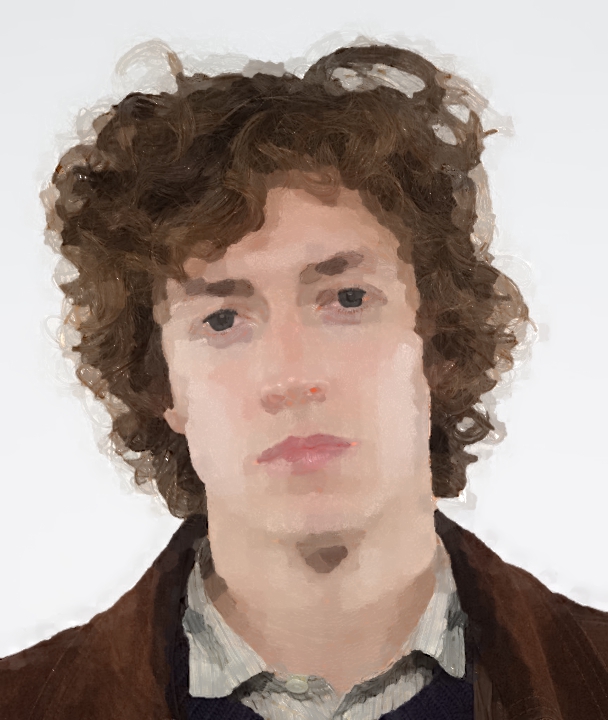}
        \caption{1250 segments}
    \end{subfigure}\\
    \begin{subfigure}{0.3\linewidth}
        \includegraphics[trim={0cm 3cm 0cm 0cm},clip,width=\linewidth]{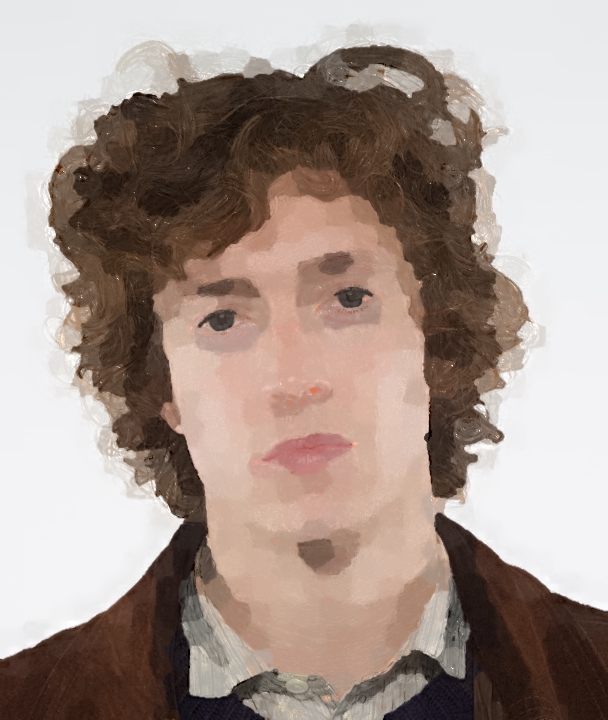}
        \caption{700 segments}
    \end{subfigure}
    \begin{subfigure}{0.3\linewidth}
        \includegraphics[trim={0cm 3cm 0cm 0cm},clip,width=\linewidth]{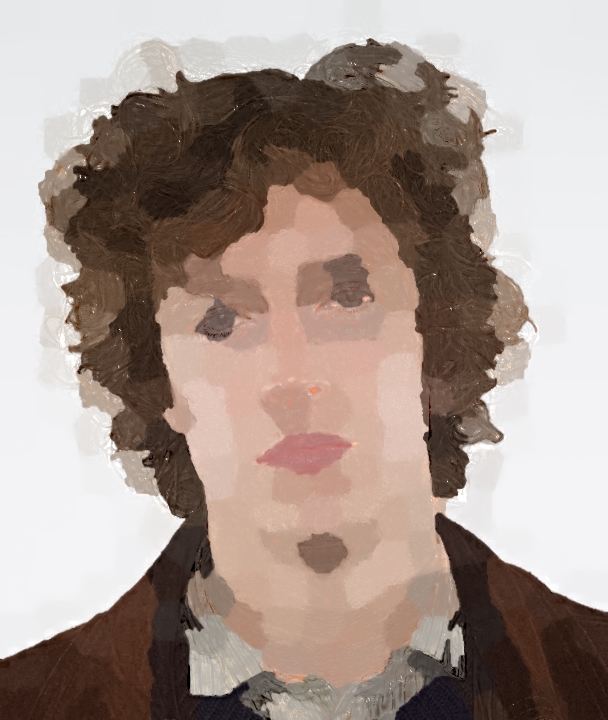}
        \caption{300 segments}
    \end{subfigure}
    \begin{subfigure}{0.3\linewidth}
        \includegraphics[trim={0cm 3cm 0cm 0cm},clip,width=\linewidth]{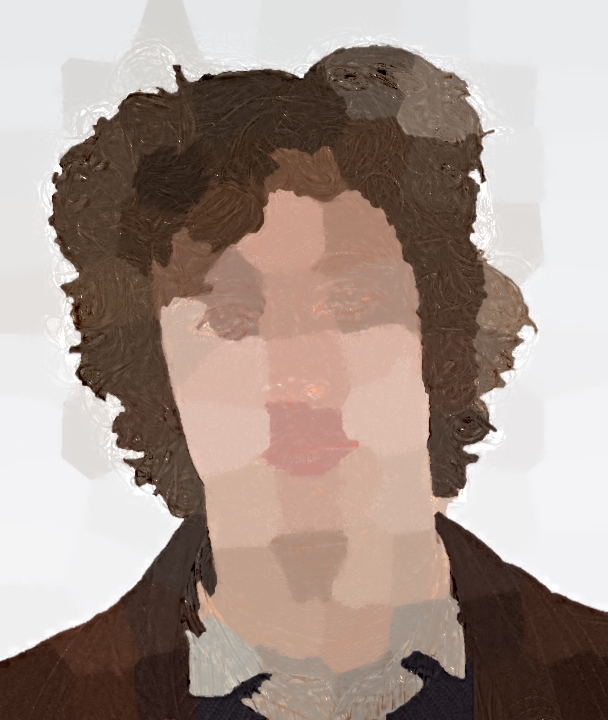}
        \caption{100 segments}
    \end{subfigure}
    \vspace{-0.15cm}\caption{Adjusting the number of segments. The parameters are optimized to match input (\subref{realsegmented:input}) using 5000 segments (\subref{realsegmented:5000seg}). When reducing the number of segments (without re-optimization), the underlying coarse-structure is spatially quantized while the high-frequency details (\ie texture) remain intact.}
    \label{fig:sup:realsegmented}
\end{figure*}

\begin{figure*}
\centering
    \begin{subfigure}{0.45\linewidth}
        \centering
      \includegraphics[width=0.9\linewidth]{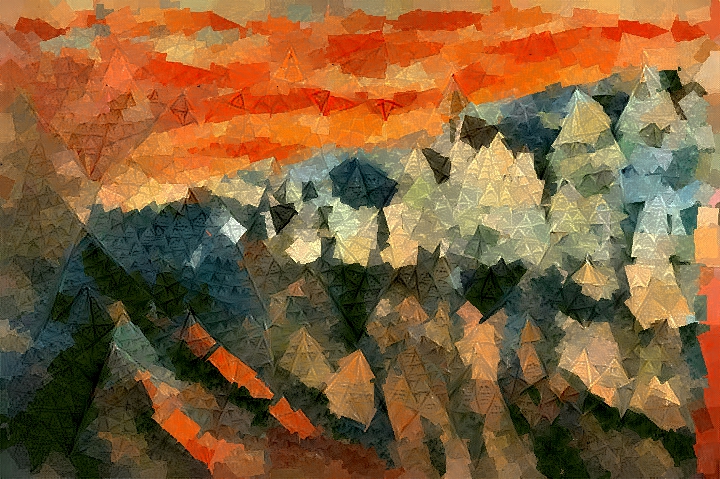}
      \caption{Paint Transformer + CLIPstyler Prompt: Triangles}
    \end{subfigure}
    \begin{subfigure}{0.45\linewidth}
        \centering
        \includegraphics[width=0.9\linewidth]{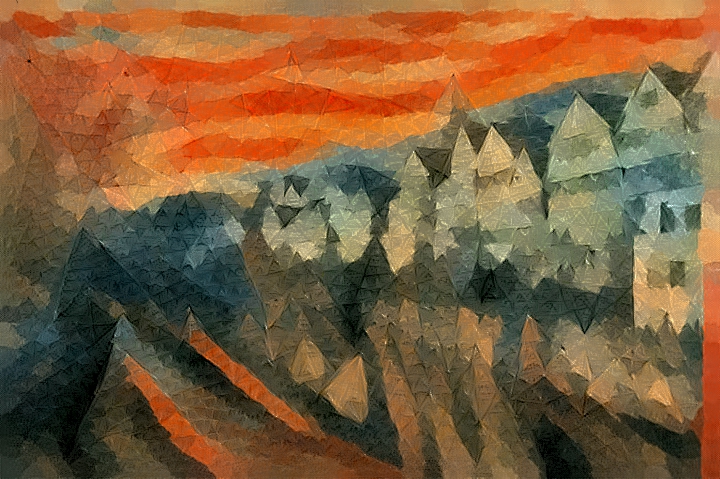}
        \caption{SLIC + CLIPstyler Prompt: Triangles}
    \end{subfigure}\\
    \caption{Comparing the influence of the segmentation stage after re-optimization by a text prompt.}
    \label{fig:sup:painttransformer_triangles}
\end{figure*}


\begin{figure*}
  \centering
  \begin{subfigure}{0.3\textwidth}
          \centering
        \includegraphics[width=\textwidth]{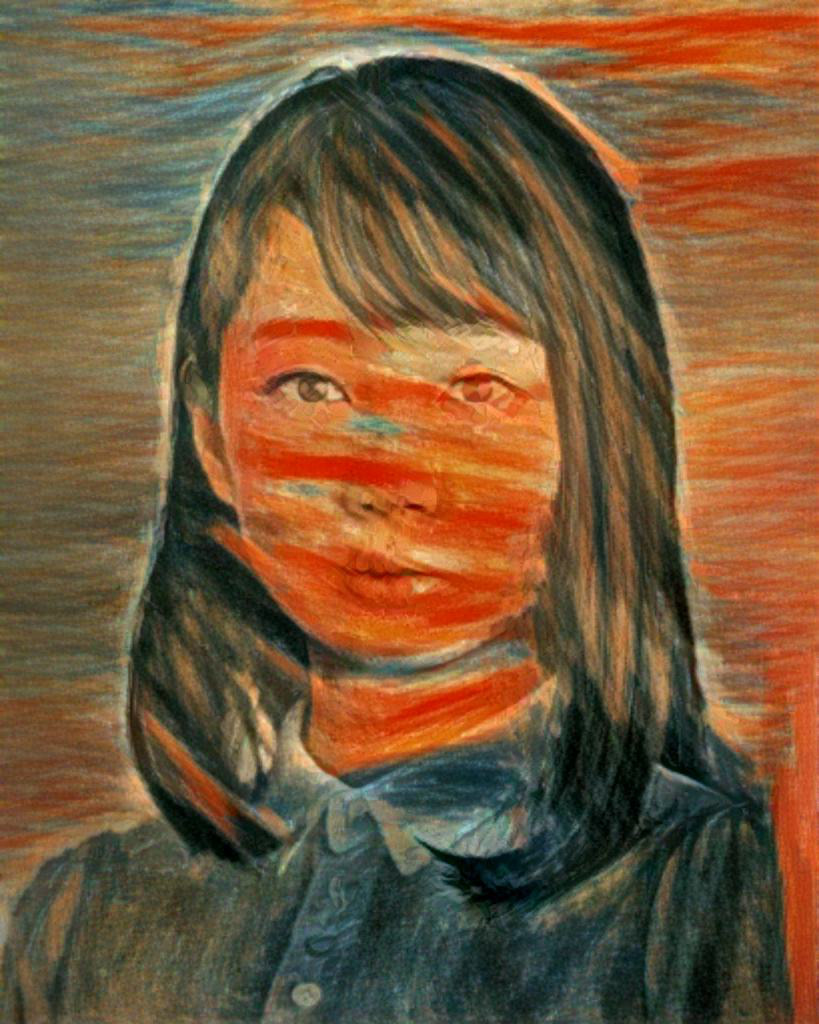}
        \caption{STROTTS~\cite{kolkin2019style} ground truth}
        \label{fig:sup:param_mask_showcase:strottsout}
      \end{subfigure}
      \begin{subfigure}{0.3\textwidth}
          \centering
          \includegraphics[width=\textwidth]{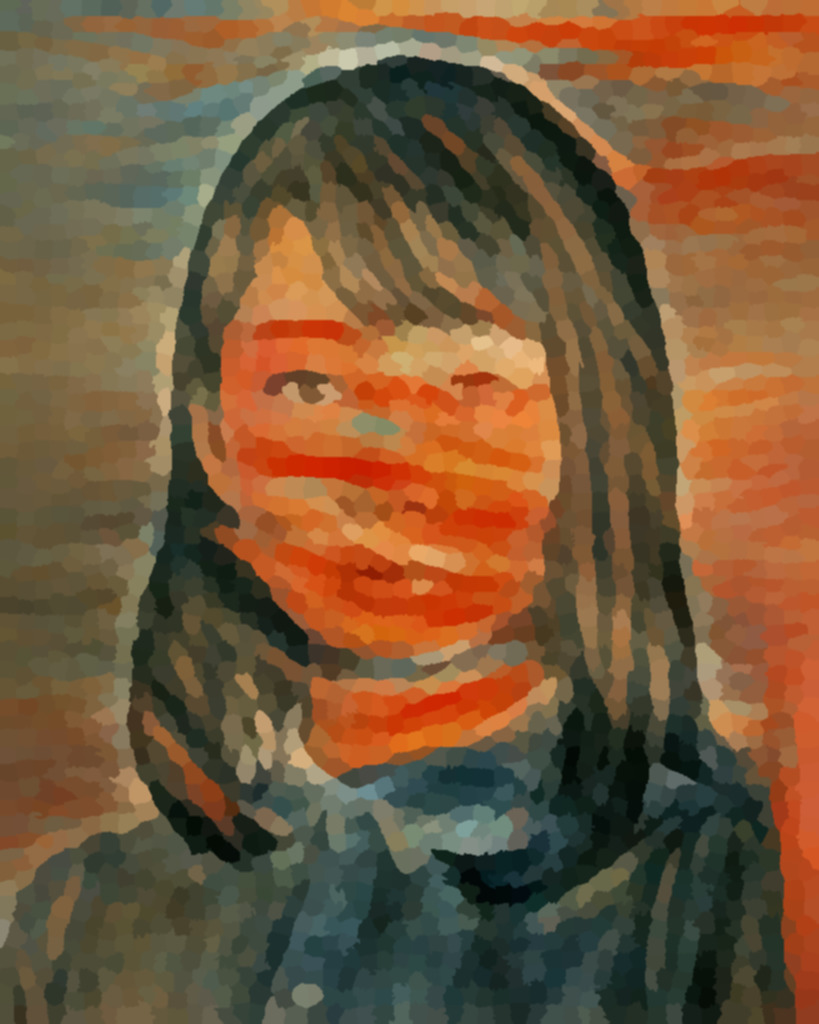}
          \caption{Segmentation Stage (SLIC)}
      \end{subfigure}
      \begin{subfigure}{0.3\textwidth}
          \centering
        \includegraphics[width=\textwidth]{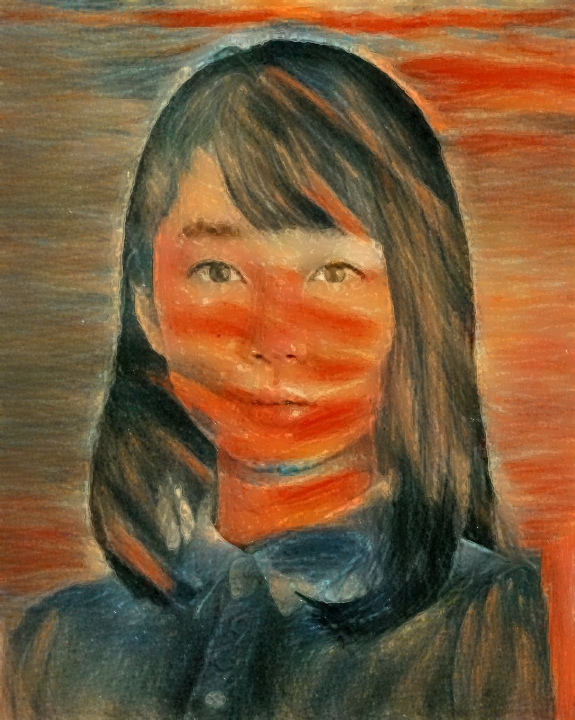}
        \caption{Final Result}
      \end{subfigure}\\ [0.5ex]

      \begin{subfigure}{0.225\textwidth}
          \centering
        \includegraphics[width=\textwidth]{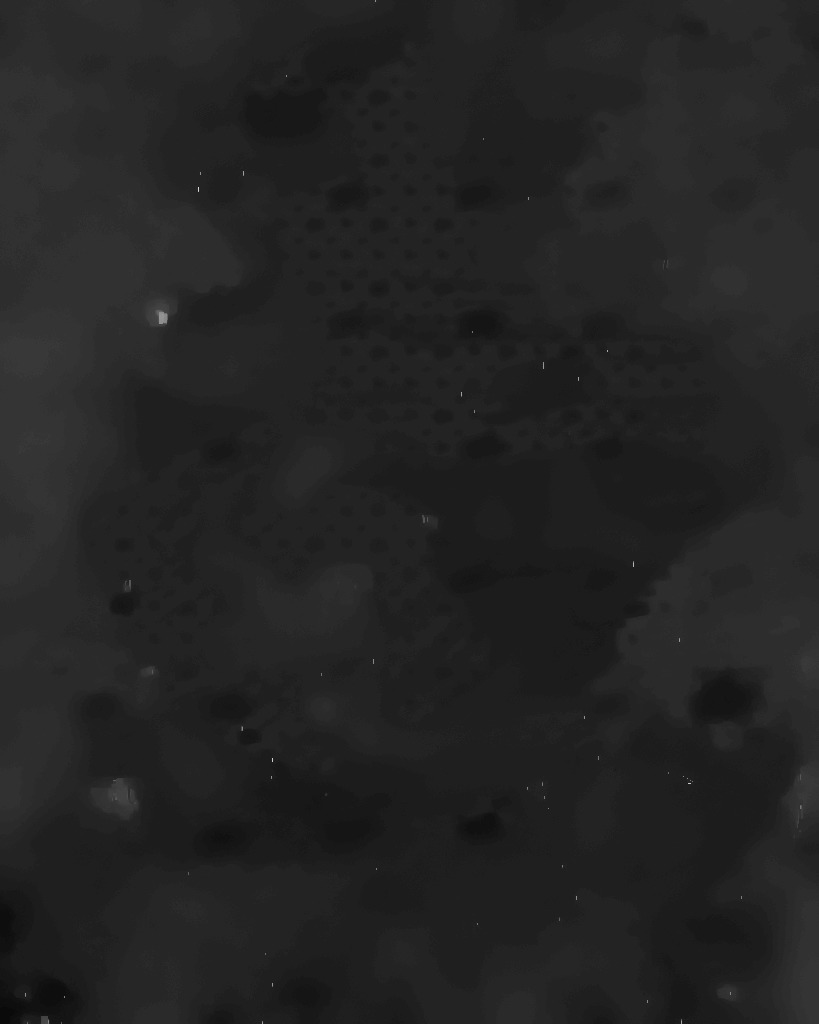}
        \caption{Bilateral D*}
      \end{subfigure}
      \begin{subfigure}{0.225\textwidth}
          \centering
          \includegraphics[width=\textwidth]{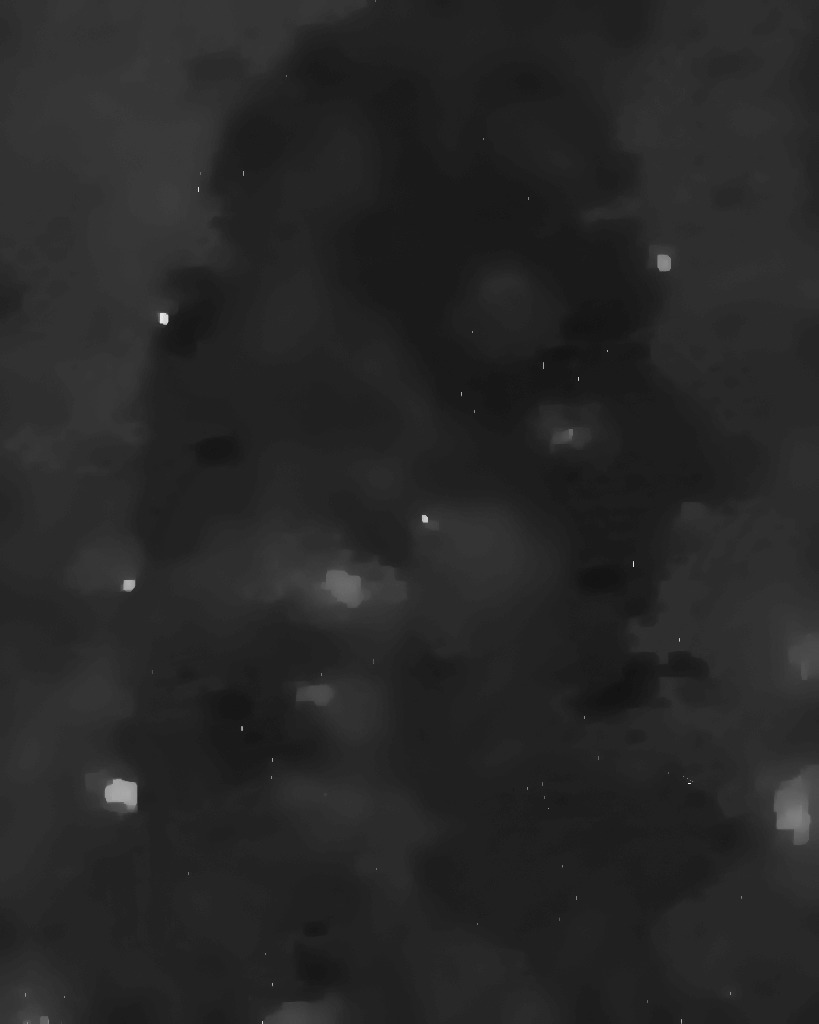}
          \caption{Bilateral R*}
      \end{subfigure}
      \begin{subfigure}{0.225\textwidth}
          \centering
        \includegraphics[width=\textwidth]{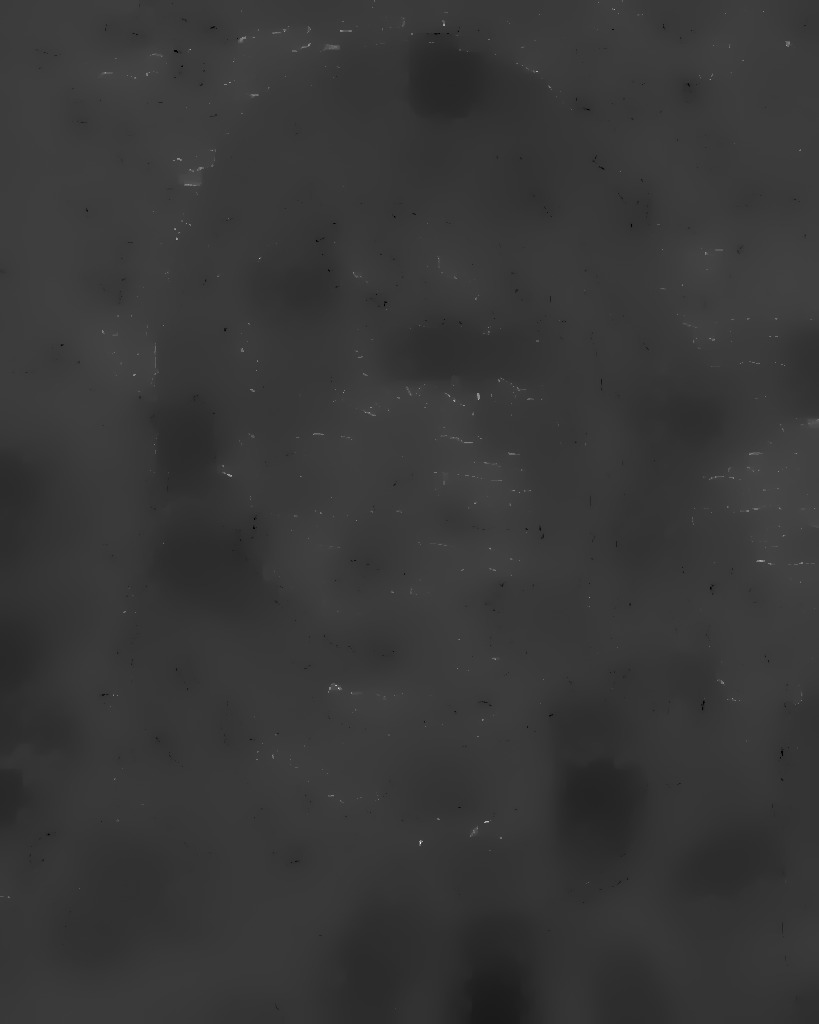}
        \caption{Contour}
      \end{subfigure}
      \begin{subfigure}{0.225\textwidth}
          \centering
        \includegraphics[width=\textwidth]{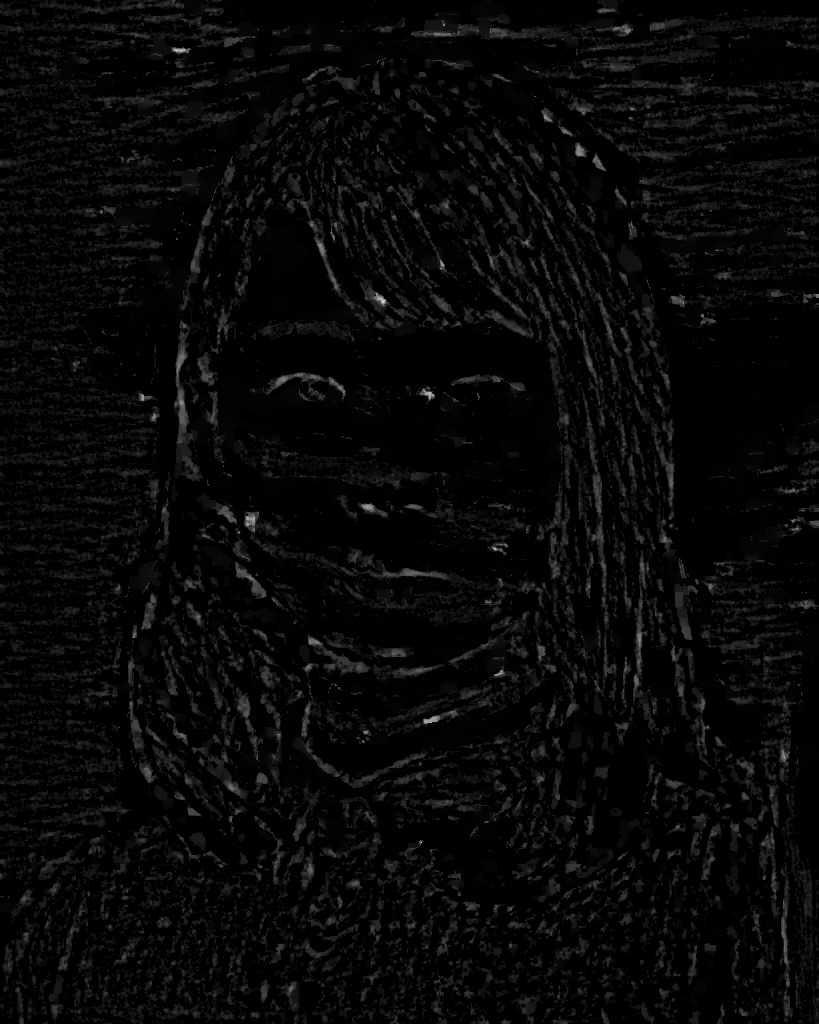}
        \caption{Contour Opacity}
      \end{subfigure}\\ [0.5ex]
      
      \begin{subfigure}{0.225\textwidth}
          \centering
        \includegraphics[width=\textwidth]{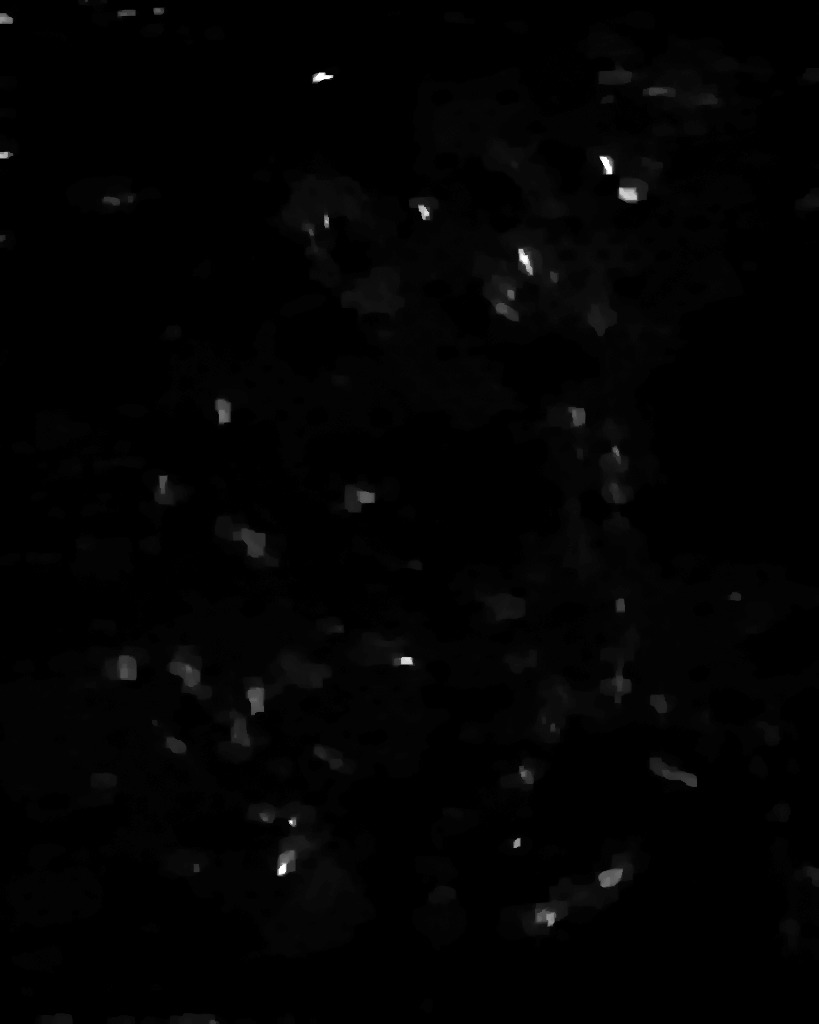}
        \caption{Phong Specular*}
      \end{subfigure}
      \begin{subfigure}{0.225\textwidth}
          \centering
          \includegraphics[width=\textwidth]{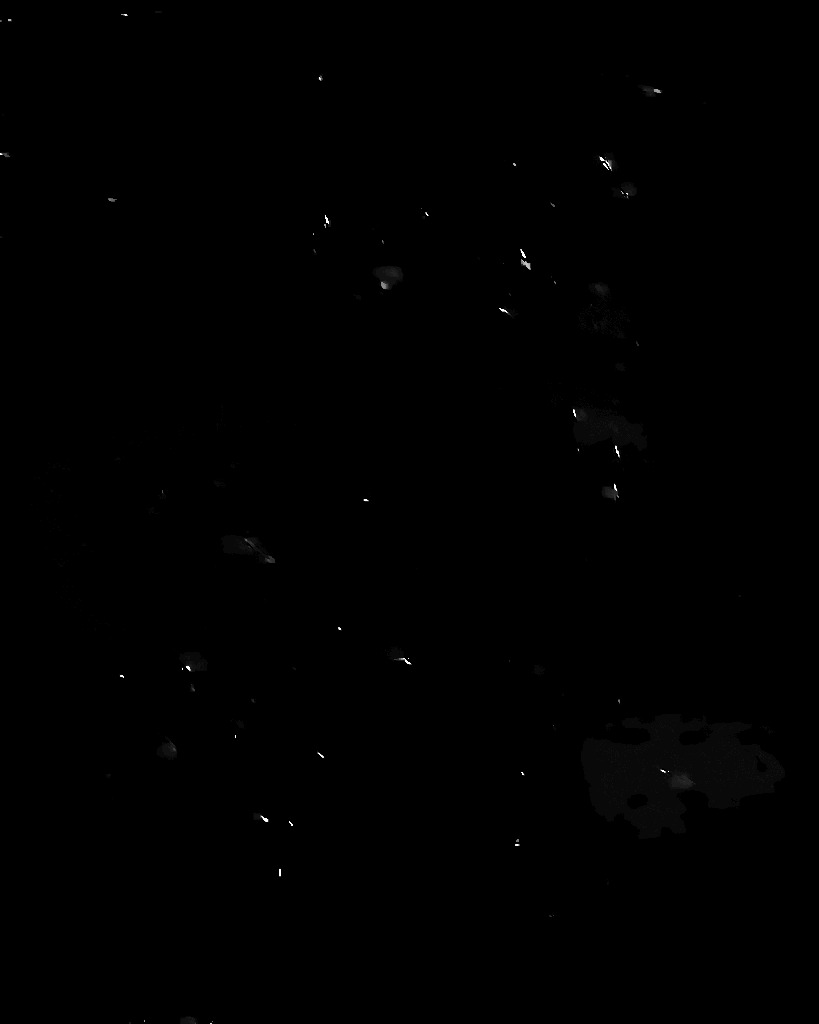}
          \caption{Bump Scale*}
      \end{subfigure}
      \begin{subfigure}{0.225\textwidth}
          \centering
        \includegraphics[width=\textwidth]{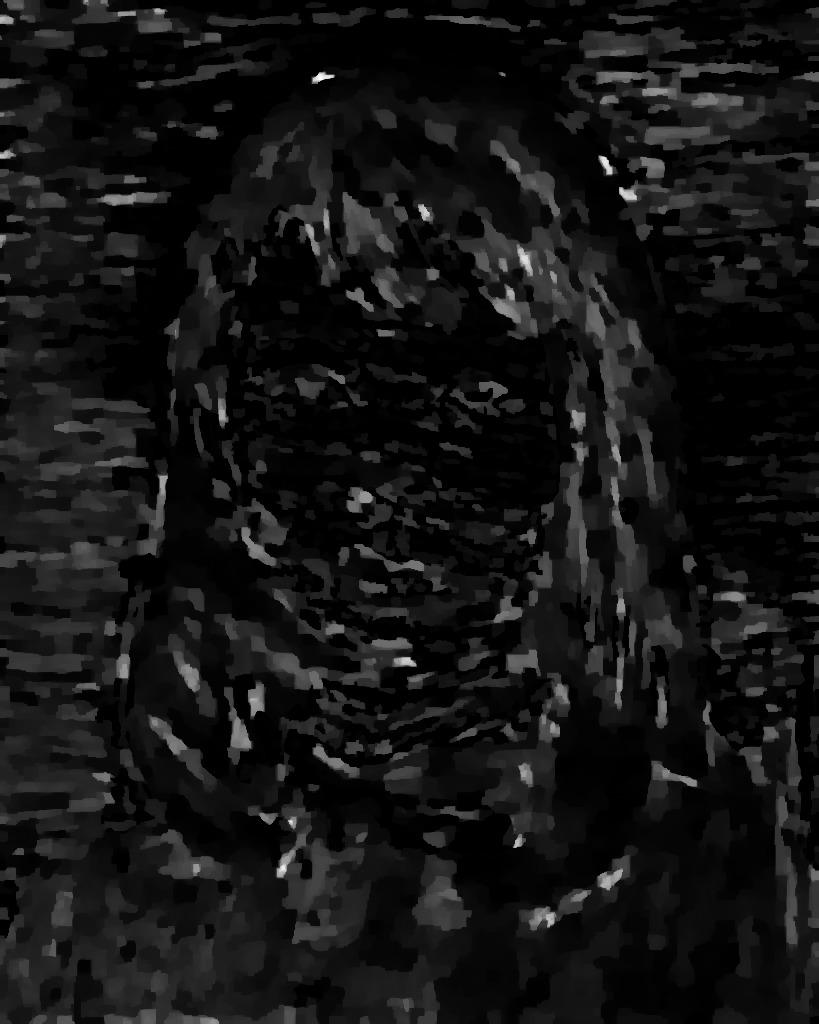}
        \caption{Bump Opacity}
      \end{subfigure}
      \begin{subfigure}{0.225\textwidth}
          \centering
        \includegraphics[width=\textwidth]{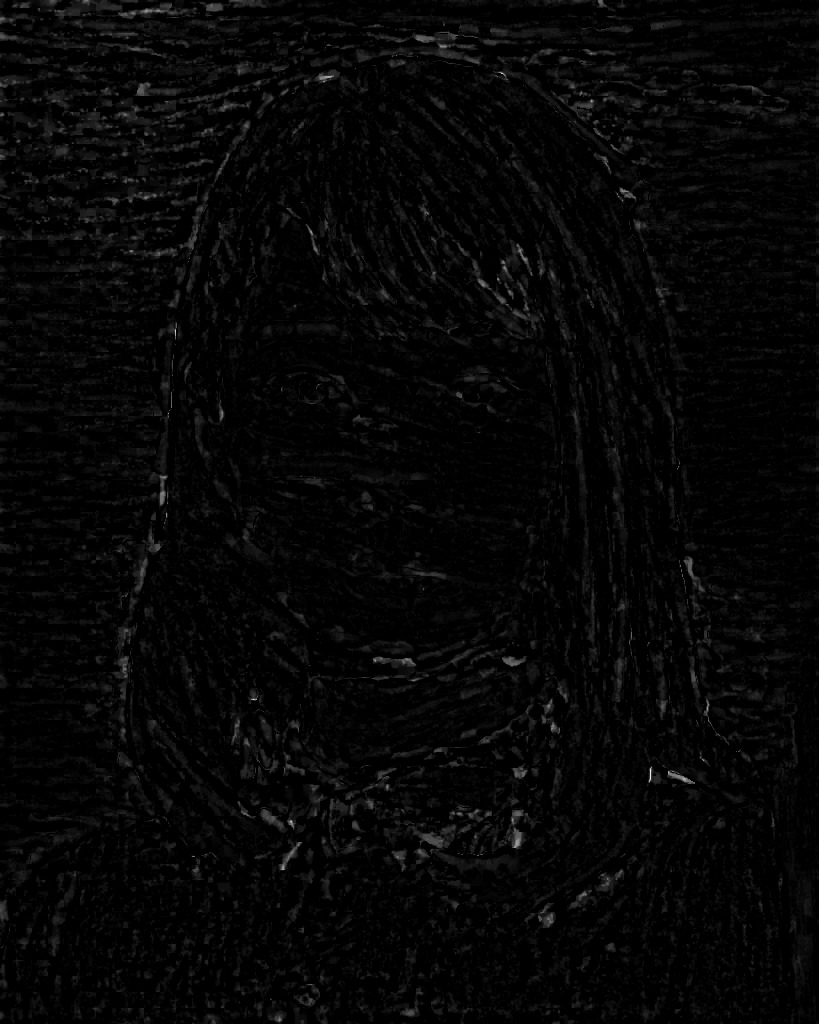}
        \caption{Contrast}
      \end{subfigure}\\
      \caption{A complete set of parameter masks ($P_M$) after $\ell_1$ optimization. We match the target (\subref{fig:sup:param_mask_showcase:strottsout}) and use a SLIC \cite{achanta2012slic} segmentation with 5000 segments. The Total Variation loss was utilized during the optimization process. To enhance visibility, some of the parameter masks have been adjusted to increase luminosity. We highlight these with an asterisk (*).}
      \label{fig:sup:param_mask_showcase}
\end{figure*}

\begin{figure*}
\centering
    \begin{subfigure}{0.48\linewidth}
        \centering
      \includegraphics[width=0.95\linewidth]{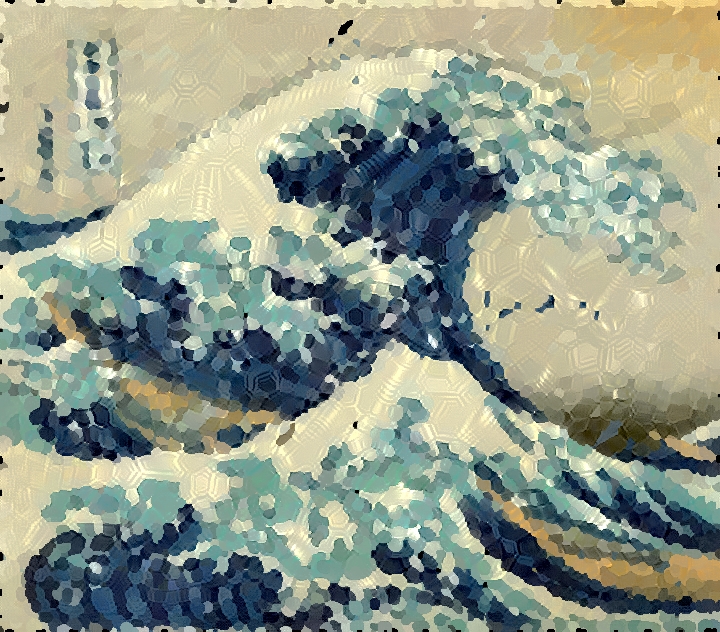}
      \caption{CLIPstyler Prompt: Shiny Chrome}
    \end{subfigure}
    \begin{subfigure}{0.48\linewidth}
        \centering
      \includegraphics[width=0.95\linewidth]{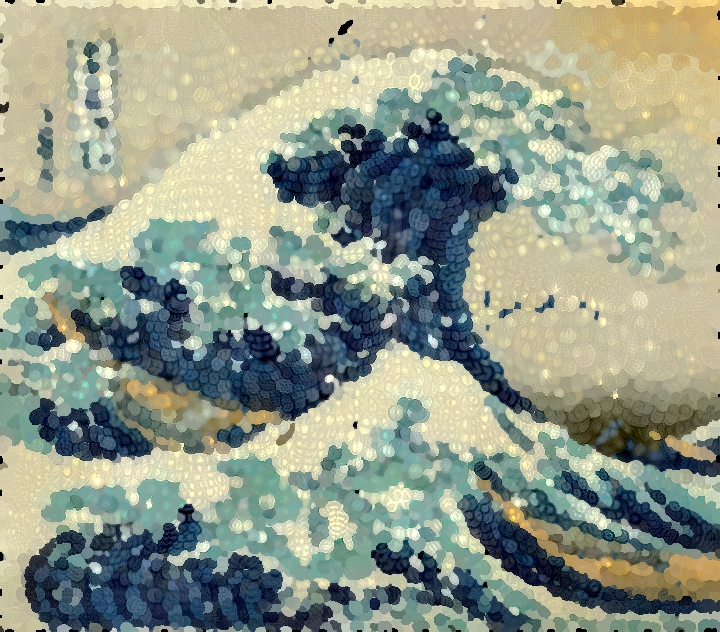}
      \caption{CLIPstyler Prompt: Camera Bokeh}
    \end{subfigure}\\
    \begin{subfigure}{0.48\linewidth}
        \centering
        \includegraphics[width=0.95\linewidth]{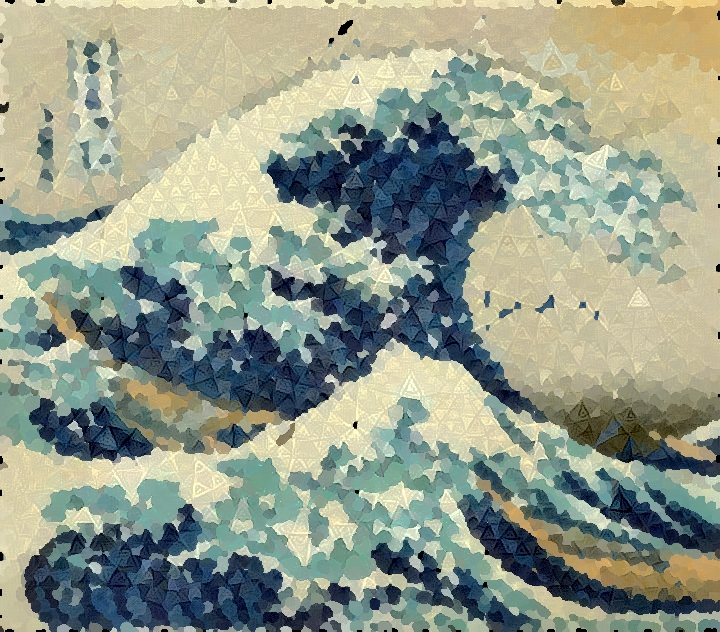}
        \caption{CLIPstyler Prompt: Triangels}
    \end{subfigure}
    \begin{subfigure}{0.48\linewidth}
        \centering
        \includegraphics[width=0.95\linewidth]{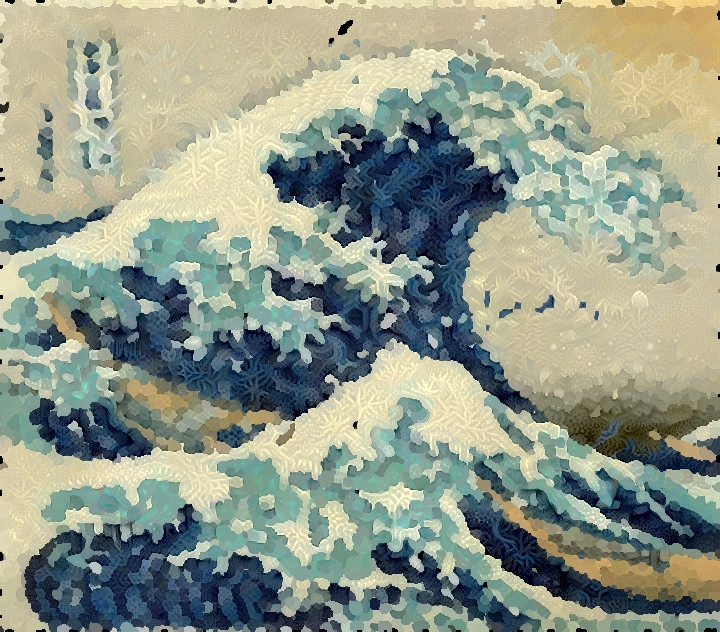}
        \caption{CLIPstyler Prompt: Icy Frost}
    \end{subfigure}\\
    \caption{Re-Optimization experiments using different text prompts. The input painting was segmented with PaintTransformer\cite{liu2021paint} and circle primitives.}
    \label{fig:sup:clipsterstyledemos4}
\end{figure*}

\begin{figure*}
\centering
    \begin{subfigure}{0.45\linewidth}
        \centering
      \includegraphics[width=0.9\linewidth]{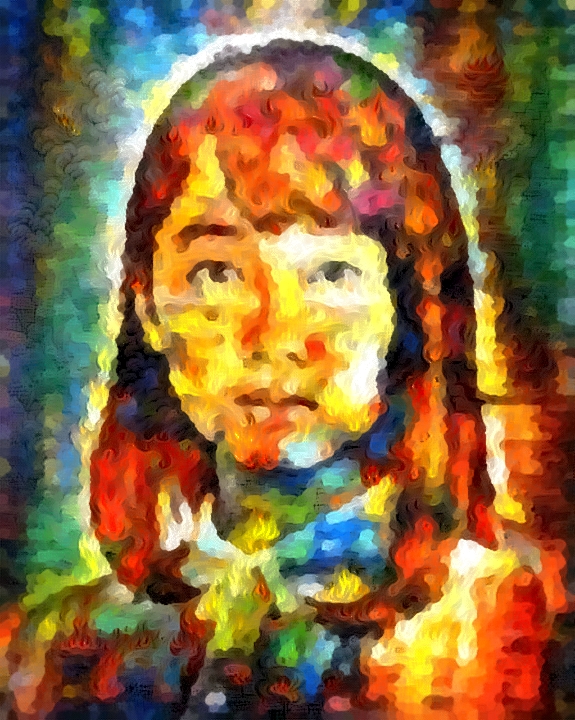}
      \caption{Global Re-Optimization using text prompt: Fire}
    \end{subfigure}
    \begin{subfigure}{0.45\linewidth}
        \centering
        \includegraphics[width=0.9\linewidth]{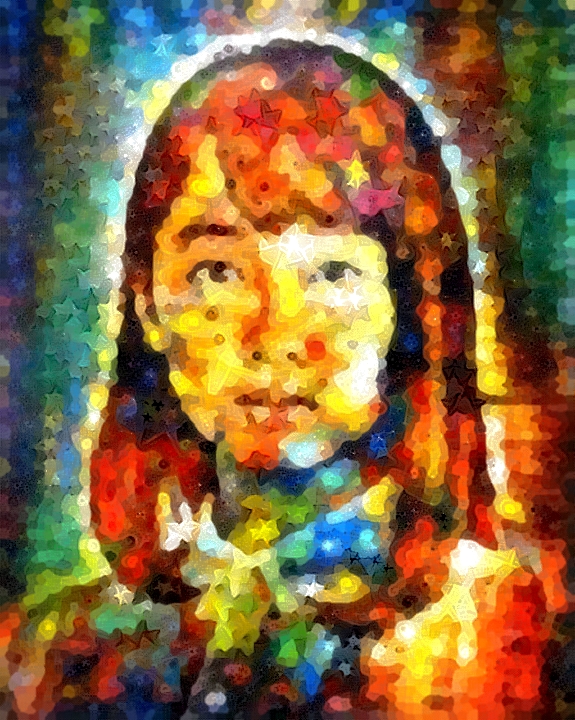}
        \caption{Global Re-Optimization using text prompt: Star}
    \end{subfigure}\\
    \begin{subfigure}{0.45\linewidth}
        \centering
      \includegraphics[width=0.9\linewidth]{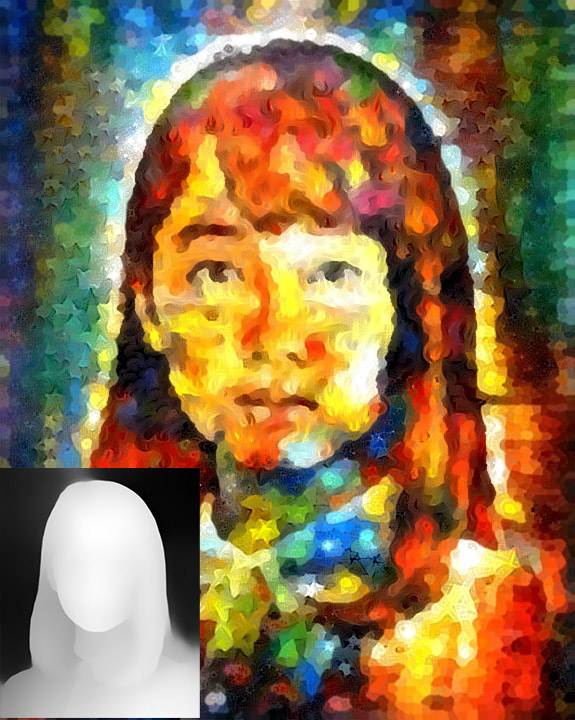}
      \caption{Blending parameters for "Fire" in the face region with "Star" in the background using depth mask.}
    \end{subfigure}
    \begin{subfigure}{0.45\linewidth}
        \centering
        \includegraphics[width=0.9\linewidth]{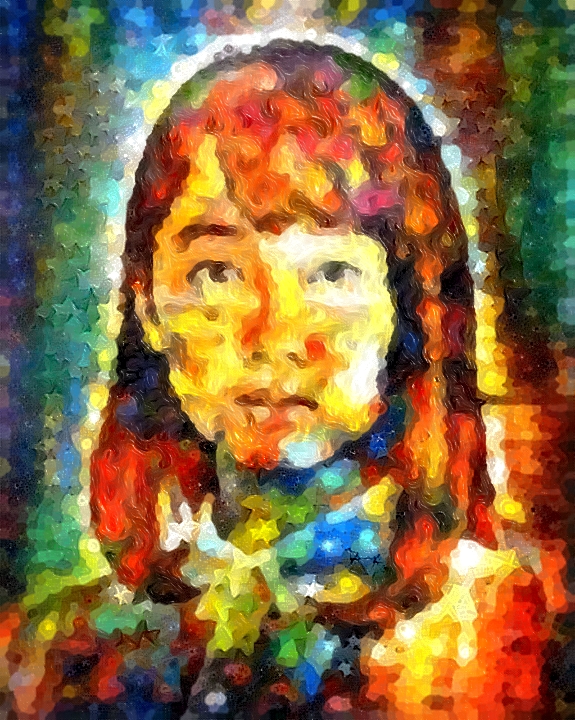}
        \caption{Fine-tune global bump and specularity parameters to increase oiliness.}
    \end{subfigure}\\
    \caption{An exemplary four-step iterative editing process demonstrating the usage of text prompts, parameter interpolation and global editing. Our method offers the advantage of allowing for adjustment of global and local parameter values at any point in the editing workflow.  Our supplementary video illustrates the interactive creation of such images.}
    \label{fig:sup:iterativeediting}
\end{figure*}

\begin{figure*}
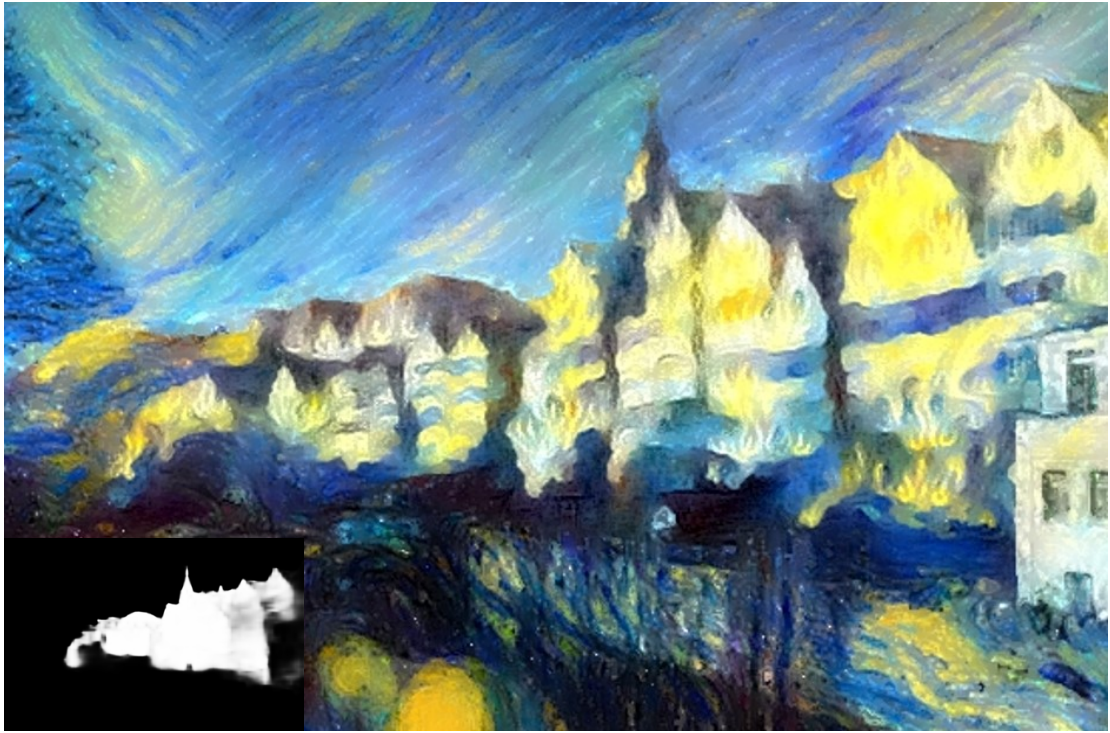
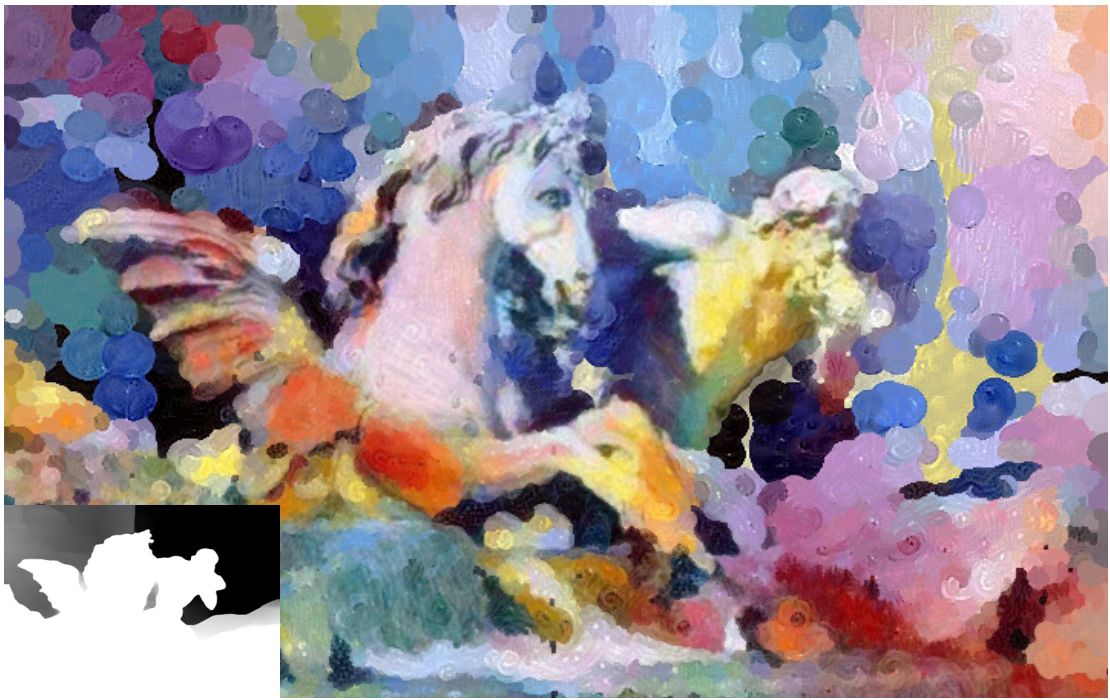

    \centering
    \begin{subfigure}{\textwidth}
    \centering
    \setbox1=\hbox{\includegraphics[width=0.8\textwidth]{graphics/saliency_interpolation/tub_figre_saliency_cropped.jpg}}
      \includegraphics[width=0.8\textwidth]{graphics/saliency_interpolation/tub_figre_saliency_cropped.jpg}\llap{\makebox[\wd1][l]{\includegraphics[trim={1cm 1cm 0.2cm 0cm},clip,height=2.55cm]{graphics/interpolation/tub_starry_fire_inverted_saliency_mask.jpeg}}}
    \end{subfigure}\\[4ex]
    \begin{subfigure}{\textwidth}
    \centering
    \setbox2=\hbox{\includegraphics[width=0.8\textwidth]{graphics/interpolation/trevi_drops_depth_womanwithhat.jpg}}
          \includegraphics[trim={0cm 2cm 0cm 0.6cm},clip,width=0.8\textwidth]{graphics/interpolation/trevi_drops_depth_womanwithhat.jpg}\llap{\makebox[\wd2][l]{\includegraphics[trim={0cm 0.2cm 0cm 0.2cm},clip,height=2.55cm]{graphics/interpolation/trevi_depth_adjusted.jpeg}}}
    \end{subfigure}
    \caption{High-res figure 7 of the main paper. First row shows saliency parameter blending, second row shows depth blending. In the first row, we use "Starry Night" by Van Gogh as our style image and interpolate with the prompt "fire". In the second row, we use "Woman with a Hat" by Matisse as our style image and interpolate with the prompt "drops" and furthermore use the depth mask for adaptive circle sizes in the PaintTransformer. }
    \label{fig:sup:interpolation}
\end{figure*}

\fi

\end{document}